\documentclass{article}

 \usepackage[preprint]{neurips_2026}
\PassOptionsToPackage{dvipsnames,table}{xcolor}
\PassOptionsToPackage{hyphens}{url}

\usepackage[utf8]{inputenc}
\usepackage[T1]{fontenc}
\usepackage{microtype}

\usepackage{amsmath, amssymb, amsthm, amsfonts}
\usepackage{mathtools}
\usepackage{bm}

\makeatletter
\@ifpackageloaded{xcolor}{}{\usepackage[dvipsnames,table]{xcolor}}
\makeatother
\usepackage{graphicx}
\usepackage{booktabs}
\usepackage{multirow}
\usepackage{tabularx}
\usepackage{array}
\usepackage{makecell}
\usepackage{siunitx}
\usepackage{subcaption}       

\usepackage{algorithm}
\usepackage{algpseudocode}
\usepackage{listings}

\usepackage{url}
\makeatletter
\@ifpackageloaded{hyperref}{}{\usepackage{hyperref}}
\@ifpackageloaded{cleveref}{}{\usepackage[capitalize,noabbrev]{cleveref}}
\makeatother

\crefformat{section}{\S#2#1#3}
\Crefformat{section}{\S#2#1#3}
\crefrangeformat{section}{\S#3#1#4--#5#2#6}
\Crefrangeformat{section}{\S#3#1#4--#5#2#6}
\crefmultiformat{section}{\S\S#2#1#3}{ and~#2#1#3}{, #2#1#3}{, and~#2#1#3}
\Crefmultiformat{section}{\S\S#2#1#3}{ and~#2#1#3}{, #2#1#3}{, and~#2#1#3}
\crefformat{subsection}{\S#2#1#3}
\Crefformat{subsection}{\S#2#1#3}
\crefrangeformat{subsection}{\S#3#1#4--#5#2#6}
\Crefrangeformat{subsection}{\S#3#1#4--#5#2#6}
\crefmultiformat{subsection}{\S\S#2#1#3}{ and~#2#1#3}{, #2#1#3}{, and~#2#1#3}
\Crefmultiformat{subsection}{\S\S#2#1#3}{ and~#2#1#3}{, #2#1#3}{, and~#2#1#3}
\hypersetup{
  colorlinks=true,
  linkcolor=black,
  citecolor=blue!60!black,
  urlcolor=blue!60!black,
}

\usepackage{ifthen}
\usepackage{xspace}
\usepackage{enumitem}

\usepackage{appendix}

\definecolor{dropMin}{HTML}{F4F1DE}  
\definecolor{dropLow}{HTML}{F5DDB5}  
\definecolor{dropMed}{HTML}{F4C19F}   
\definecolor{dropHigh}{HTML}{EAA088}  
\newcommand{\legbox}[2]{{\setlength{\fboxsep}{0.1pt}\colorbox{#1}{#2}}}

\newboolean{showcomments}
\setboolean{showcomments}{true}  

\DeclareRobustCommand{\makecomment}[3]{%
  \ifthenelse{\boolean{showcomments}}%
    {\textcolor{#1}{\textbf{[#2: #3]}}}{}%
}

\DeclareRobustCommand{\cut}[1]{\ifthenelse{\boolean{showcomments}}%
  {\textcolor{gray}{\sout{#1}}}{}}

\newcommand{\benchmark}{\textsc{BreakingWeb}}

\theoremstyle{plain}

\theoremstyle{definition}

\theoremstyle{remark}

\usepackage[utf8]{inputenc} 
\usepackage[T1]{fontenc}    
\usepackage{hyperref}       
\usepackage{url}            
\usepackage{booktabs}       
\usepackage{amsfonts}       
\usepackage{nicefrac}       
\usepackage{microtype}      
\usepackage{xcolor}         
\usepackage[most]{tcolorbox} 
\usepackage{wrapfig}        

\title{Constructing Challenging Browser-Use Tasks by \\ Controlled Environment Interventions}

\author{%
  \textbf{Xunjian Yin\textsuperscript{1}\thanks{Equal contribution.} \quad Tianchen Guan\textsuperscript{1}\footnotemark[1] \quad
  Jinao Wang\textsuperscript{1} \quad Weili Cao\textsuperscript{1}} \\
  \textbf{Daisy Xinlei Lin\textsuperscript{2} \quad Royce Cheng-Yue\textsuperscript{2} \quad 
  Keagan Long\textsuperscript{2} \quad Kyle Wong\textsuperscript{2}} \\
  \textbf{Bhuwan Dhingra\textsuperscript{1} \quad Xiangjun Wang\textsuperscript{2} \quad
  Shuyan Zhou\textsuperscript{1}} \\
  \textsuperscript{1}Duke University \qquad \textsuperscript{2}Amazon \\
  \texttt{\{xunjian.yin, shuyan.zhou\}@duke.edu}
}

\begin{document}

\maketitle

\begin{abstract}
As browser-use agents improve, benchmarks keep pace by adding new tasks, websites, and applications. Doing so is costly, and because many factors change at once, it is unclear what makes the new tasks harder. In this work, we instead make existing tasks harder by changing the environment in which they are performed, so that difficulty becomes a programmable property of the environment.
We introduce \benchmark{}, in which each task is run twice: once on an unmodified website and once under an intervention, a controlled change to the website's content, server state, network responses, or rendered page. Each intervention is reproducible, keeps the user instruction and success criterion fixed, can be detected and worked around by a competent user, and is labeled with the behavior that recovery from it mainly requires, such as verification. \benchmark{} contains $519$ task pairs on seven self-hosted websites, and each run is scored on the website's final backend state.
We evaluate six strong browser-use agents, three agents that observe only screenshots, and human participants. Interventions lower the pass rates of the six browser-use agents by $22.9$ percentage points on average, and each agent fails on nearly half of the tasks it solves on the unmodified website. Humans are affected much less: their pass rate drops by $10.0$ points on the first attempt and by $5.7$ points when they repeat the task. Most failures of the six agents ($75\%$) are \emph{belief failures}, in which the agent declares the run finished although a required change was never made.
Our code, data, and environments are available at \href{https://www.breakingweb.app}{\nolinkurl{www.breakingweb.app}}.
\end{abstract}

\section{Introduction}

\begin{figure}[t]
\centering
\includegraphics[width=\linewidth]{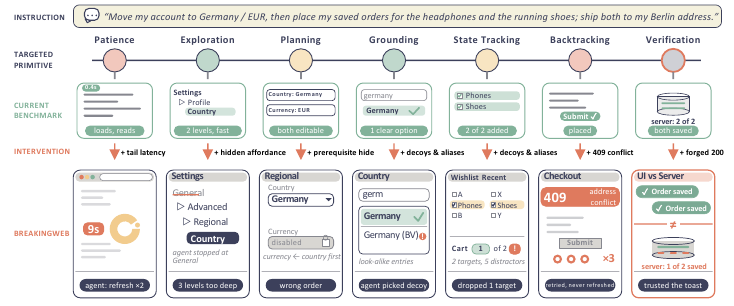}
\caption{\textbf{\benchmark{} constructs challenging browser-use tasks by intervening on the environment.} A shopping task is run in the clean condition (top) and under matched interventions that keep the instruction and the success criterion fixed (bottom). Stage labels name the primary primitive of each intervention.}
\label{fig:teaser}
\vspace{-16pt}
\end{figure}

Benchmarks such as WebArena~\citep{DBLP:conf/iclr/ZhouX0ZLSCOBF0N24}, Mind2Web~\citep{DBLP:conf/nips/DengGZCSWSS23}, and the computer-use benchmark OSWorld~\citep{DBLP:conf/nips/XieZCLZCHCSLLXZ24} evaluate agents across many websites, applications, and multi-step workflows. As agents improve, benchmarks keep pace by collecting new tasks. New tasks extend coverage, but they are costly to build and make difficulty hard to control: when the goal of a task and its environment change together, it is unclear which change made the task harder. We therefore ask how an existing task can be made systematically more challenging without changing what successful completion means.

Related evaluations address other aspects of robustness. AgentDojo~\citep{DBLP:conf/nips/DebenedettiZBB024} measures task utility and security under prompt injection, ST-WebAgentBench~\citep{levy2025stwebagentbench} evaluates task completion under enterprise policies, and ReliabilityBench~\citep{gupta2026reliabilitybench} studies repeated-run consistency, instruction perturbations, and tool-level faults. In contrast, we construct recoverable changes to the website itself, keep the instruction and the backend success criterion fixed, and compare each changed task with the same task on the unmodified website.

In this work, we introduce \benchmark{}, a framework for constructing harder versions of existing browser-use tasks. Given a task in its \emph{clean} condition, on the unmodified website, we construct an \emph{intervention} condition by applying a realistic change to the website that keeps the user instruction and the backend success criterion fixed and from which a competent user can recover. Because the change blocks the usual way of completing the task but not the intended outcome, the intervention condition tests whether an agent can adapt when the environment behaves unexpectedly. The performance gap between the two conditions measures the agent's robustness to that change, which makes difficulty a programmable property of the environment.

\Cref{fig:teaser} illustrates this construction. We view a web application as four layers at which an intervention can act: the content seeded into the website, the server state, the network traffic, and the page rendered in the browser. Interventions at these layers introduce complications such as decoy records, hidden prerequisites, transient errors, and misleading success messages. We also label each intervention with the behavior that recovery from it mainly requires. We call these behaviors \emph{cognitive primitives} and use seven of them, following earlier analyses of agent failures~\citep{xi2023agentsurvey,DBLP:conf/nips/XieZCLZCHCSLLXZ24,DBLP:conf/iclr/0036YZXLL0DMYZ024}: grounding, planning, state tracking, backtracking, patience, exploration, and verification. The layer specifies \emph{where} the environment changes, and the primitive describes \emph{how} the agent must adapt. Because recovery can involve more than one primitive, the labels serve only to organize the catalog and the analysis; the intervention remains the unit of construction and measurement.

We implement the framework on seven self-hosted websites and construct $519$ task pairs, each with a clean and an intervention condition. Every intervention is reproducible from a seed, detectable by the agent, and recoverable (\cref{sec:suite}). We score each run on the final backend state: a task passes only if all required changes were made and no protected data was modified, regardless of the agent's actions or its claim of success.

We evaluate six browser-use agents built on the \texttt{browser-use} library~\citep{browser_use2024} with Opus-4.7, Sonnet-4.6, GPT-5.4, GPT-5.4-mini, Gemini-3.1-Pro, and Gemini-3-Flash~\citep{anthropic2026opus47,anthropic2026sonnet46,openai2026gpt54,openai2026gpt54minino,google2025gemini3,google2025gemini3flash,deepmind2026gemini31pro}. We also evaluate three \emph{GUI-only} agents (Gemini-3.1-Pro, GPT-5.4, and Opus-4.7) that observe only screenshots through BrowserGym~\citep{DBLP:journals/tmlr/ChezellesGLCDBT25}, and two open-weight models, Kimi-K2.5 and Qwen3-VL-235B~\citep{kimiteam2026k25,qwenteam2025qwen3vl}, with \texttt{browser-use}.
Under intervention, the pass rates of the six browser-use agents drop by $17.9$--$27.6$ percentage points, and each agent fails on $39.6$--$68.0\%$ of the tasks that it solves in the clean condition. Most of their failures ($75\%$) are \emph{belief failures}, in which the agent declares the run finished although a required change is missing. Of the intervention runs that one of these agents declares finished, $54$--$85\%$ have in fact failed, depending on the agent. The GUI-only agents also lose performance, but most of their failures ($57\%$ in aggregate) are action failures, such as repeating an action without progress.

Humans are affected much less. In a study on $140$ tasks, the pass rate of human participants drops by $10.0$ percentage points on the first attempt and by $5.7$ points when they repeat the same task immediately after a reset. Small, recoverable changes to a website are therefore enough to make agents fail on many tasks that they complete on the unmodified website.


\section{Benchmark Design}
\label{sec:framework}

\subsection{Tasks and Paired Interventions}
\label{sec:eval-method}

A task is defined by the backend state it should produce rather than by a reference sequence of UI actions, so any sequence of actions that produces this state passes. Scoring compares the actual change in backend state with a per-task specification that has two parts: \emph{positive obligations} (entries the agent must produce) and \emph{invariants} (state the agent must not modify). For example, the Gmail task ``mark all five unread invoice emails as read'' requires five matching state updates and forbids changes to any other message. Marking four of the five emails earns partial credit but fails the task, because one obligation is unmet; marking all five plus an unrelated message satisfies the obligations but violates an invariant. Because we score the backend state, a save that the page reports as successful but that never reaches the server (for example, one confirmed by a forged ``Saved'' message) counts as a failure. \Cref{app:bench} formalizes the scoring rule and the per-environment evaluator.

Each task is run twice, with the same instruction, initial state, and success criterion. The \emph{clean} run uses the unmodified environment; the \emph{intervention} run repeats the task with one \emph{variant}, a concrete intervention written for this task and labeled with the primitive that recovery from it mainly requires (\cref{sec:primitives}). Because everything else is held fixed, the score difference between the two runs, which we call the \emph{paired drop}, measures the effect of the intervention itself, and grouping variants by primary primitive shows where the drop is largest.

\subsection{The Intervention Catalog}
\label{sec:suite}

A variant is admitted to the catalog only if it satisfies five \emph{design rules}. It must be \emph{deterministic} (the same seed always produces exactly the same intervention), \emph{detectable} (the changed state remains observable through the DOM, the HTTP status, or form readback), \emph{recoverable} (a competent agent can work around the change with a limited number of extra actions, so the intervention makes the task harder but not impossible), \emph{primitive-pure} (each variant has a single primary primitive, defined in \cref{sec:primitives}, and the base task and its variant together involve at most two primitives; this makes the grouping by primitive well defined), and \emph{realistic} (every variant corresponds to a real-world failure class such as a slow network, a phishing email, a broken layout, or a transient server error). \Cref{app:invariant-justification} explains the reason for each rule.

\begin{wrapfigure}{r}{0.39\linewidth}
\vspace{-1.0em}
\centering
\includegraphics[width=\linewidth]{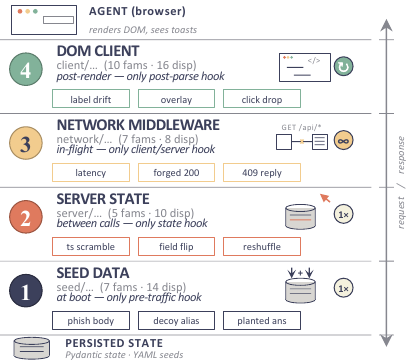}
\caption{The four injection layers and where each acts in the request path.}
\label{fig:inject-stack}
\vspace{-1.5em}
\end{wrapfigure}
We place each intervention at the layer of the web application where the corresponding real failure occurs, which keeps the intervention realistic and leaves the meaning of the task unchanged.
Web failures arise at different points of the request path. A phishing email is already in the inbox when the page loads, a $503$ error arrives from the network during the task, and a click that the page silently ignores happens in the browser. No single injection mechanism reproduces all of these: an error simulated inside the page lacks real network timing, and a network component cannot hide elements of the page. \benchmark{} therefore divides interventions into four layers according to where each takes effect (\cref{fig:inject-stack}). The \emph{seed} layer changes the content once at session creation, before the page loads (phishing bodies, decoys, planted wrong answers).\footnote{Seed- and server-layer content was drafted with LLM assistance and reviewed by humans against the five design rules.} The \emph{server} layer changes the structure of the state once after seeding (scrambled timestamps, hidden labels, single-field corruption). The \emph{network} layer intercepts matching API calls at runtime (transient $5xx$ and $401$ errors, $409$ conflicts, stale data, forged $200$ responses). The \emph{client} layer changes the DOM and the interaction in the browser (misaligned labels, swallowed clicks, overlays).

\begin{figure}[t]
\centering
\includegraphics[width=\linewidth]{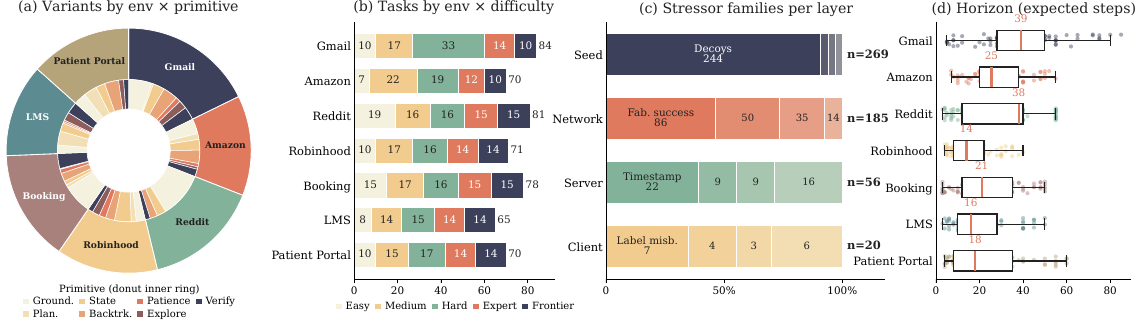}
\caption{Composition of \benchmark{}. (a)~Intervention variants by environment and target primitive. (b)~Base tasks by environment and difficulty tier. (c)~Injection records; multi-layer variants contribute one record per layer they touch. (d)~Expected step counts per task by environment.}
\label{fig:composition}
\vspace{-10.5pt}
\end{figure}

We group the $519$ variants into $29$ \emph{intervention families} (``stressor families'' in some figure panels). The variants in one family use the same mechanism, for example added network latency. \Cref{fig:composition} summarizes the catalog by environment, primitive, layer, family, and number of steps. All but four combinations of environment and primitive contain at least one variant (\cref{tab:coverage-prim-env}).

\subsection{Cognitive Primitives as Intervention Labels}
\label{sec:primitives}

The seven primitives are general abilities that browser-use tasks require on any website, as opposed to site-specific knowledge such as how the filters on Booking work. We label every intervention with the primitive that recovery from it mainly requires, so that paired results can be grouped by the kind of recovery rather than by website. We chose the seven primitives under two constraints. First, they must cover the failure modes reported in recent agent studies (grounding errors and operational gaps in OSWorld~\citep{DBLP:conf/nips/XieZCLZCHCSLLXZ24}, long-horizon reasoning and instruction-following failures in AgentBench~\citep{DBLP:conf/iclr/0036YZXLL0DMYZ024}, and failures of multi-step plans on real websites in Mind2Web~\citep{DBLP:conf/nips/DengGZCSWSS23}), so that every reported failure class falls under some primitive. Second, they must be distinct enough that each intervention has one clear primary primitive, even though recovery from a single intervention can involve more than one primitive. \Cref{tab:primitives} lists each primitive with examples of the interventions that target it; \cref{app:primitive-choice} discusses alternative sets of primitives and whether the set transfers to other domains.

\begin{table}[t]
  \caption{The seven cognitive primitives, with representative interventions. The examples are illustrative; \cref{app:catalog} describes the full catalog of $519$ variants in $29$ intervention families.}
  \label{tab:primitives}
  \centering
  \footnotesize
  \begin{tabular}{lp{0.40\linewidth}p{0.40\linewidth}}
    \toprule
    Primitive & Recovery behavior it requires & Representative interventions \\
    \midrule
    Grounding       & Identify the correct UI target among noise, decoys, and adversarial content.                                  & phishing emails, decoys \& aliases, label-input misalignment \\ \addlinespace[0.5pt]
    Planning        & Decompose a goal into ordered sub-goals and revise the plan as new information arrives.                       & scrambled timestamps, missing prerequisites \\ \addlinespace[0.5pt]
    State tracking  & Track which steps are done and which are pending across a long trajectory, including out-of-order updates.           & shuffled lists, scrambled timestamps, stale responses \\ \addlinespace[0.5pt]
    Backtracking    & Detect a blocked path, revert to a prior decision point, and try an alternative.                              & transient errors, 409 conflict, planted wrong answer \\ \addlinespace[0.5pt]
    Patience        & Wait or retry at a suitable rate when responses are slow, unreliable, or rate-limited.       & tail latency, progressive delay, transient errors \\ \addlinespace[0.5pt]
    Exploration     & Find another way to act when the obvious path is blocked.                                                 & hidden target, hidden affordance, restricted affordance set \\ \addlinespace[0.5pt]
    Verification    & Check after an action that the resulting state matches the intended outcome.                                                    & silent fail, misleading ``Saved'' toast, save drift \\
    \bottomrule
  \end{tabular}
\vspace{-12pt}
\end{table}

\vspace{-4pt}
\section{Benchmarking Browser-Use Agents}
\label{sec:benchmarking}

This section asks three questions: how much the interventions reduce agent performance relative to the clean runs, which primitives show the largest paired drops, and whether the same primitives are weak for all text-based agents. We compare agents with humans in \cref{sec:human}.

\noindent\textbf{Experimental setup.}\label{sec:setup}
Each base task is paired with one variant, and we run both conditions with every model. Unless otherwise stated, runs use the public seed $s=42$ and the text-based harness (the \texttt{browser-use} library, which serializes the DOM as text). A full evaluation of one model therefore consists of $1{,}038$ runs ($519$ tasks in two conditions), and paired drops are computed over the base tasks for which both runs are available. All runs are limited to $40$ actions and a model-specific time limit of $600$--$1{,}200$~s in both conditions; the step and time estimates stored with each task only assign difficulty levels (\cref{app:difficulty}). \Cref{app:baselines} lists the model snapshots, decoding settings, budgets, and harness versions.

\noindent\textbf{Run lifecycle and scoring.}
Each run starts from a fresh backend state, browser context, and model conversation; agents carry no memory across tasks or between the two conditions of a base task. A run ends when the agent declares it finished, exhausts its action or time budget, or the harness raises an unrecoverable error; in all cases, we score the final backend state. We report the \emph{pass rate} (the fraction of runs with score $1.0$ and no penalty) and the \emph{mean score} (the average graded score with partial credit). The paired drop of model $m$ on base task $t$ is $\Delta_{m,t}=\mathrm{score}_{m,t,\textsc{clean}}-\mathrm{score}_{m,t,\textsc{intervention}}$. Ranking the seven primitives by their average drop over the six text-based agents gives the same order for pass rate and mean score (Spearman $\rho=1.0$). With a pass threshold of $0.5$ instead of $1.0$, the average drop is still $16.6$ points (versus $22.9$), so it is not caused only by runs that narrowly miss the threshold.

\begin{table}[t]
\centering
\caption{Per-primitive intervention pass rate (\%). Cell background encodes the relative drop $|\Delta|/\text{pass}_\textsc{clean}$:
\legbox{dropMin}{$<\!15\%$}, \legbox{dropLow}{$15\text{--}30\%$},
\legbox{dropMed}{$30\text{--}45\%$}, \legbox{dropHigh}{$\geq\!45\%$}. Each cell shows the pass rate under intervention, followed by the change from the clean condition in points ($\downarrow$ decrease, $\uparrow$ increase). \Cref{tab:headline-full} adds the clean pass rates and the drops in mean score, and \cref{fig:env-primitive-grid} breaks the results down by environment. Rows prefixed \emph{v-} are GUI-only runs of the same models. The last two rows report the cold and warm human attempts on Human-140 (\cref{sec:human}).
}
\label{tab:headline}
\footnotesize
\setlength{\tabcolsep}{4pt}
\begin{tabular}{lrrrrrrrr}
\toprule
 & & \multicolumn{7}{c}{Per primitive} \\
\cmidrule(lr){3-9}
Agent
& Total & Grounding & Plan & State. & Back. & Patience & Expl. & Verif. \\
\midrule
Gemini-3.1-Pro     & \cellcolor{dropMed}44.5\,{\scriptsize $\downarrow$27.6} & \cellcolor{dropMed}42.3\,{\scriptsize $\downarrow$34.5} & \cellcolor{dropMin}59.6\,{\scriptsize $\downarrow$9.6}  & \cellcolor{dropLow}48.8\,{\scriptsize $\downarrow$13.1} & \cellcolor{dropHigh}40.2\,{\scriptsize $\downarrow$37.8} & \cellcolor{dropMed}48.0\,{\scriptsize $\downarrow$24.0} & \cellcolor{dropLow}37.8\,{\scriptsize $\downarrow$10.8} & \cellcolor{dropHigh}40.8\,{\scriptsize $\downarrow$39.4} \\
Gemini-3-Flash     & \cellcolor{dropMed}42.6\,{\scriptsize $\downarrow$20.6} & \cellcolor{dropMed}34.5\,{\scriptsize $\downarrow$22.0} & \cellcolor{dropMin}61.5\,{\scriptsize $\downarrow$3.8}  & \cellcolor{dropLow}47.6\,{\scriptsize $\downarrow$13.1} & \cellcolor{dropHigh}42.7\,{\scriptsize $\downarrow$35.4} & \cellcolor{dropMin}48.0\,{\scriptsize $\downarrow$4.0}  & \cellcolor{dropLow}43.2\,{\scriptsize $\downarrow$8.1}  & \cellcolor{dropHigh}39.4\,{\scriptsize $\downarrow$33.8} \\
GPT-5.4            & \cellcolor{dropMed}35.1\,{\scriptsize $\downarrow$26.2} & \cellcolor{dropHigh}26.8\,{\scriptsize $\downarrow$28.6} & \cellcolor{dropMed}44.2\,{\scriptsize $\downarrow$28.8} & \cellcolor{dropLow}36.9\,{\scriptsize $\downarrow$14.3} & \cellcolor{dropHigh}41.5\,{\scriptsize $\downarrow$35.4} & \cellcolor{dropMin}52.0\,{\scriptsize $\downarrow$4.0}  & \cellcolor{dropMed}29.7\,{\scriptsize $\downarrow$13.5} & \cellcolor{dropHigh}35.2\,{\scriptsize $\downarrow$36.6} \\
GPT-5.4-mini       & \cellcolor{dropHigh}15.2\,{\scriptsize $\downarrow$17.9} & \cellcolor{dropHigh}15.5\,{\scriptsize $\downarrow$14.3} & \cellcolor{dropMed}15.4\,{\scriptsize $\downarrow$11.5} & \cellcolor{dropMed}17.9\,{\scriptsize $\downarrow$8.3}  & \cellcolor{dropHigh} 8.5\,{\scriptsize $\downarrow$36.6} & \cellcolor{dropHigh} 8.0\,{\scriptsize $\downarrow$20.0} & \cellcolor{dropMin}18.9\,{\scriptsize $\downarrow$2.7}  & \cellcolor{dropHigh}19.7\,{\scriptsize $\downarrow$28.2} \\
Opus-4.7           & \cellcolor{dropMed}31.6\,{\scriptsize $\downarrow$22.9} & \cellcolor{dropHigh}29.2\,{\scriptsize $\downarrow$24.4} & \cellcolor{dropHigh}23.1\,{\scriptsize $\downarrow$21.2} & \cellcolor{dropMed}31.0\,{\scriptsize $\downarrow$19.0} & \cellcolor{dropMed}37.8\,{\scriptsize $\downarrow$29.3} & \cellcolor{dropLow}32.0\,{\scriptsize $\downarrow$8.0}  & \cellcolor{dropLow}27.0\,{\scriptsize $\downarrow$5.4}  & \cellcolor{dropHigh}39.4\,{\scriptsize $\downarrow$32.4} \\
Sonnet-4.6         & \cellcolor{dropMed}28.1\,{\scriptsize $\downarrow$22.2} & \cellcolor{dropHigh}24.4\,{\scriptsize $\downarrow$29.8} & \cellcolor{dropHigh}21.2\,{\scriptsize $\downarrow$19.2} & \cellcolor{dropMed}26.2\,{\scriptsize $\downarrow$17.9} & \cellcolor{dropMed}34.1\,{\scriptsize $\downarrow$25.6} & \cellcolor{dropMin}28.0\,{\scriptsize $\uparrow$0.0}  & \cellcolor{dropMin}27.0\,{\scriptsize $\downarrow$2.7}  & \cellcolor{dropMed}38.0\,{\scriptsize $\downarrow$25.4} \\
\midrule
v-Gemini-3.1-Pro  & \cellcolor{dropHigh}13.1\,{\scriptsize $\downarrow$15.0} & \cellcolor{dropHigh}11.9\,{\scriptsize $\downarrow$14.3} & \cellcolor{dropMed}11.5\,{\scriptsize $\downarrow$7.7}  & \cellcolor{dropHigh}15.5\,{\scriptsize $\downarrow$15.5} & \cellcolor{dropHigh}12.2\,{\scriptsize $\downarrow$18.3} & \cellcolor{dropMin} 8.0\,{\scriptsize $\uparrow$0.0}   & \cellcolor{dropLow}16.2\,{\scriptsize $\downarrow$5.4}  & \cellcolor{dropHigh}15.5\,{\scriptsize $\downarrow$28.2} \\
v-GPT-5.4         & \cellcolor{dropHigh} 5.6\,{\scriptsize $\downarrow$4.6}  & \cellcolor{dropLow} 7.1\,{\scriptsize $\downarrow$2.4}  & \cellcolor{dropHigh} 3.8\,{\scriptsize $\downarrow$3.8}  & \cellcolor{dropHigh} 4.8\,{\scriptsize $\downarrow$6.0}  & \cellcolor{dropHigh} 4.9\,{\scriptsize $\downarrow$6.1}  & \cellcolor{dropMin} 4.0\,{\scriptsize $\uparrow$4.0}   & \cellcolor{dropMin} 5.4\,{\scriptsize $\uparrow$2.7}   & \cellcolor{dropHigh} 5.6\,{\scriptsize $\downarrow$14.1} \\
v-Opus-4.7        & \cellcolor{dropHigh}15.8\,{\scriptsize $\downarrow$17.9} & \cellcolor{dropHigh}16.1\,{\scriptsize $\downarrow$17.3} & \cellcolor{dropMed}17.3\,{\scriptsize $\downarrow$7.7}  & \cellcolor{dropMed}16.7\,{\scriptsize $\downarrow$11.9} & \cellcolor{dropHigh}14.6\,{\scriptsize $\downarrow$25.6} & \cellcolor{dropLow}16.0\,{\scriptsize $\downarrow$4.0}  & \cellcolor{dropMed}13.5\,{\scriptsize $\downarrow$10.8} & \cellcolor{dropHigh}15.5\,{\scriptsize $\downarrow$33.8} \\
\midrule
\textit{Human cold} & \cellcolor{dropMin}66.4\,{\scriptsize $\downarrow$10.0} & \cellcolor{dropLow}66.7\,{\scriptsize $\downarrow$13.3} & \cellcolor{dropMin}50.0\,{\scriptsize $\downarrow$7.1} & \cellcolor{dropLow}50.0\,{\scriptsize $\downarrow$18.2} & \cellcolor{dropMin}86.4\,{\scriptsize $\downarrow$9.1} & \cellcolor{dropMin}71.4\,{\scriptsize $\uparrow$28.6} & \cellcolor{dropLow}40.0\,{\scriptsize $\downarrow$10.0} & \cellcolor{dropMin}85.0\,{\scriptsize $\downarrow$10.0} \\
\textit{Human warm} & \cellcolor{dropMin}75.0\,{\scriptsize $\downarrow$5.7} & \cellcolor{dropLow}68.9\,{\scriptsize $\downarrow$15.6} & \cellcolor{dropMin}64.3\,{\scriptsize $\downarrow$7.1} & \cellcolor{dropMin}68.2\,{\scriptsize $\downarrow$4.5} & \cellcolor{dropMin}100.0\,{\scriptsize $\uparrow$4.5} & \cellcolor{dropMin}71.4\,{\scriptsize $\uparrow$42.9} & \cellcolor{dropMin}60.0\,{\scriptsize $\downarrow$10.0} & \cellcolor{dropMin}85.0\,{\scriptsize $\downarrow$10.0} \\
\bottomrule
\end{tabular}
\vspace{-14pt}
\end{table}

\begin{figure}[t]
\centering
\includegraphics[width=\linewidth]{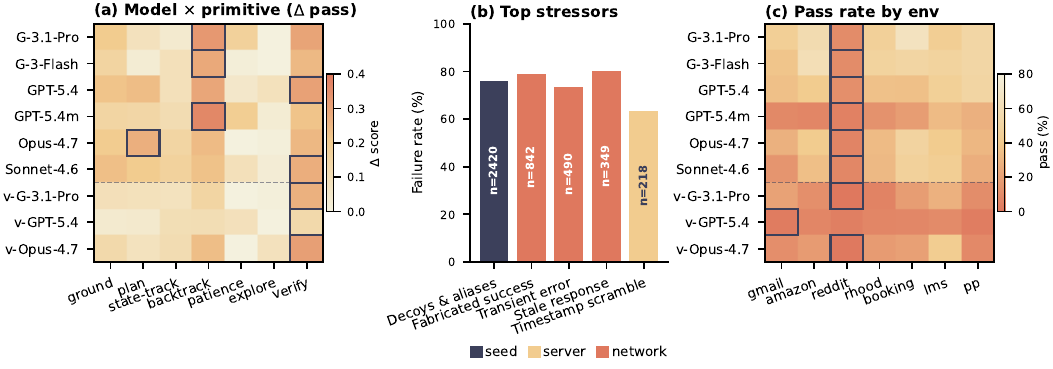}
\caption{Distribution of intervention effects. \textbf{(a)} Primitive vulnerability matrix: each cell is the mean paired drop $\Delta = \mathrm{score}_{\mathrm{clean}} - \mathrm{score}_{\mathrm{intervention}}$ on variants whose primary primitive is that primitive, over base tasks with both conditions; bootstrap $95\%$ confidence intervals resample paired (model, task) units. The row maximum is highlighted. \textbf{(b)} Top-5 intervention families ranked by the number of failed runs ($n \times$ failure rate), pooled across all ten evaluated models (six text-based, three GUI-only, and Qwen3-VL-235B) on intervention buckets with $n \geq 50$; bar color encodes the injection layer. The full ranking is in \cref{app:catalog}. \textbf{(c)} Intervention pass rate per environment and model; \cref{app:env-model} gives the numbers.}
\label{fig:results-overview}
\vspace{-6pt}
\end{figure}

\noindent\textbf{Interventions reduce pass rate across agents.}
\Cref{tab:headline} reports each agent's intervention pass rate per primitive, and \cref{fig:results-overview}(a) breaks the paired drop down by primitive. Of the tasks that an agent solves in the clean condition, it fails on $39.6$--$68.0\%$ under intervention ($49.4\%$ on average; $41.7\%$ for the strongest agent, Gemini-3.1-Pro, and $68.0\%$ for GPT-5.4-mini), so the interventions cause new failures for strong and weak agents alike.

The drops differ between primitives. For five of the six text-based agents, the largest drop in mean score is on backtracking or verification ($0.22$--$0.36$ and $0.22$--$0.30$, respectively), and the smallest drops are on exploration and patience. The ranking of agents is broadly similar across primitives: Gemini-3.1-Pro has the highest clean pass rate on five of the seven primitives (one of them tied) but the highest intervention pass rate on only three, and GPT-5.4-mini has its largest drops on backtracking ($36.6$ points) and verification ($28.2$ points). Below the strongest agent, however, the size of a drop does not follow the clean pass rate; for example, Opus-4.7 ranks fourth in clean pass rate but has the largest planning drop in mean score ($0.27$). Exploration shows a small average drop ($\Delta\approx0.02$ in mean score) because its clean mean score is the lowest of all primitives ($0.38$--$0.67$ per agent and $0.53$ on average, against $0.60$--$0.78$ for the other primitives). Among the tasks that an agent solves in the clean condition, exploration interventions still cause $18$--$62\%$ to fail ($33.7\%$ on average). As \cref{sec:analysis} shows, most failures of the text-based agents are \emph{belief failures}, in which the agent declares the run finished although a required change is missing.

The GUI-only agents start from much lower clean pass rates, $21$--$51$ points below the text-based agents that use the same model. Their drops are therefore smaller in absolute terms (at most $33.8$ points), but they are measured from a low baseline. \Cref{app:headline-full} gives the full breakdown, including the drops in mean score, and the results are stable when the catalog is reweighted or subsampled (\cref{app:env-primitive}).

\noindent\textbf{A single intervention family accounts for many failures.}
The family with the largest effect is fabricated success (\texttt{network/fabricated\_success}). It accounts for $504$ of the $3{,}114$ intervention runs of the six text-based agents ($16.2\%$), and $70\%$ of these runs fail, more than in any other family of similar size (\cref{fig:results-overview}(b)). Variants in this family return an apparently successful HTTP response but leave the server state unchanged, so the correct response is to re-read the relevant state before declaring the run finished. The family mainly tests whether an agent verifies the website state after acting, and its failure rate depends strongly on the model: it ranges from $62\%$ to $94\%$, with GPT-5.4-mini the highest. \Cref{sec:analysis} gives the corresponding failure-mode evidence and a representative trace. Intervention pass rates per environment are summarized in \cref{fig:results-overview}(c) and tabulated in \cref{app:env-model}.

Under intervention, the pass rates of the open-weight models Kimi-K2.5 and Qwen3-VL-235B drop by $14.3$ and $12.1$ points; because their clean pass rates are lower, this is a qualitative check of generalization rather than a comparison of capability (\cref{app:additional-models}). \Cref{app:controlled-studies} reports controlled checks that vary the strength of an intervention, combine two mechanisms on the same tasks, and replicate the drop across seeds.

\section{Analysis}
\label{sec:analysis}

Of the $3{,}469$ failed intervention runs of the nine agents, we classify $3{,}037$ into six mutually exclusive failure modes (\cref{tab:modes}) and use this classification to compare how text-based and GUI-only agents fail. We group the six modes into three classes. A \emph{belief failure} (\texttt{misleading\_success\_taken} or \texttt{premature\_done}) ends with the agent's completion action, by which it declares the run finished, although at least one required change is missing; the agent need not claim success. An \emph{action failure} (\texttt{retry\_loop}, \texttt{plan\_collapse}, or \texttt{selector\_hallucination}) ends without a completion action, for example after repeated actions. In \emph{overreach} (\texttt{silent\_overreach}), no positive check fails but a negative check does, which normally means that the agent modified data that should have remained unchanged.

\begin{figure}[t]
\centering
\includegraphics[width=\linewidth]{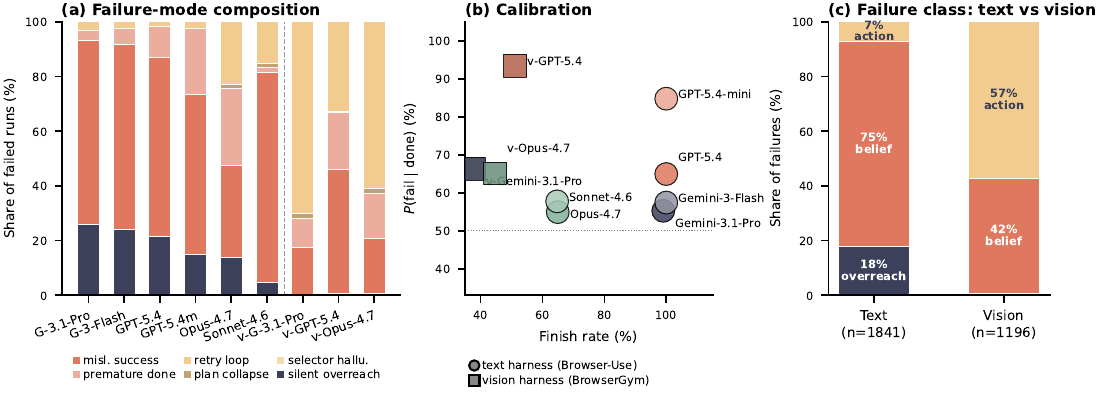}
\caption{Failure modes of the nine agents (six text-based agents left of the dashed divider, three GUI-only agents right). \textbf{(a)} Composition of failure modes; coral shades = belief, gold shades = action, navy = \texttt{silent\_overreach}. \textbf{(b)} Calibration: finish rate (share of runs that end with a completion report) vs.\ $P(\text{failed}\mid\text{declared done})$; circles = text-based, squares = GUI-only. \textbf{(c)} Failure classes aggregated by harness.}
\label{fig:failure-landscape}
\vspace{-11pt}
\end{figure}

\noindent\textbf{Failure-mode taxonomy.}
The classifier assigns each failed run to one mode using features of its trajectory: the final action, the longest run of identical consecutive actions, the number of action errors, the failed positive checks, the failed invariant checks, and a keyword scan of the agent's final reasoning. The rules are applied in the order shown in \cref{tab:modes}; the order matters because \texttt{misleading\_success\_taken} is a special case of \texttt{premature\_done} and must therefore be checked first. The other $432$ failed runs, which mostly end without a completion action after more than five steps (about a third reach the $40$-step limit), receive the fallback label \texttt{abandoned\_run} and are excluded from this analysis (\cref{app:modes}).

\begin{table}[t]
\centering
\caption{Failure-mode taxonomy over the $3{,}037$ failed intervention trajectories of the nine agents (six text-based and three GUI-only; the $432$ runs labeled \texttt{abandoned\_run} are excluded).}
\label{tab:modes}
\footnotesize
\begin{tabular}{p{0.24\linewidth}rp{0.65\linewidth}}
\toprule
Mode & Count & Identification rule \\
\midrule
\texttt{misleading success taken} & 1{,}454 & Final action is a done verb; a positive check failed; final thought contains \emph{saved/success/submitted/confirmed/added/starred/completed/done/sent} \\
\texttt{premature done}            &   433 & Final action is a done verb; a positive check failed \\
\texttt{retry loop}                &   790 & At least $5$ identical consecutive actions \\
\texttt{plan collapse}             &    24 & No done verb; run length $\leq 5$ \\
\texttt{silent overreach}          &   336 & Final action is a done verb; no positive check failed; a negative check failed \\
\texttt{selector hallucination}    &   0 & $\geq 3$ action errors; all such runs match an earlier rule in this evaluation \\
\bottomrule
\end{tabular}
\end{table}

\begin{wraptable}{r}{0.50\linewidth}
\centering
\caption{Shares of belief failures, action failures, and overreach among failed intervention runs, per model. \emph{FPR} is the false-positive rate: the share of runs ending with a completion report in which the task actually failed.
}
\label{tab:belief-vs-action}
\footnotesize
\setlength{\tabcolsep}{3.5pt}
\renewcommand{\arraystretch}{0.95}
\begin{tabular}{lrrrrr}
\toprule
Model & Fail $n$ & Belief & Action & Over.\ & FPR \\
\midrule
Gemini-3.1-Pro     & 288 & 70.8\% &  3.1\% & 26.0\% & 54\% \\
Gemini-3-Flash     & 298 & 73.5\% &  2.3\% & 24.2\% & 57\% \\
GPT-5.4            & 337 & 76.9\% &  1.8\% & 21.4\% & 65\% \\
GPT-5.4-mini       & 440 & 82.7\% &  2.3\% & 15.0\% & 85\% \\
Opus-4.7           & 245 & 61.6\% & 24.5\% & 13.9\% & 55\% \\
Sonnet-4.6         & 233 & 78.5\% & 16.7\% &  4.7\% & 58\% \\
\midrule
v-Gemini-3.1-Pro  & 439 & 28.0\% & 71.8\% &  0.2\% & 66\% \\
v-GPT-5.4         & 361 & 66.2\% & 33.2\% &  0.6\% & 94\% \\
v-Opus-4.7        & 396 & 36.6\% & 62.6\% &  0.8\% & 65\% \\
\bottomrule
\end{tabular}

\end{wraptable}

\Cref{fig:failure-landscape}(a) shows the composition of failures for each of the nine agents (text-based agents left of the dashed divider, GUI-only agents right). For every text-based agent, the most common class is belief failure, and \texttt{misleading\_success\_taken} alone accounts for $34$--$77\%$ of failures. These agents are rarely unable to act on the page. Instead, they declare the run finished while a required change is still missing. Two of the three GUI-only agents fail mainly through action failures, almost all of them \texttt{retry\_loop} ($72\%$ of failures for v-Gemini-3.1-Pro and $63\%$ for v-Opus-4.7): they often cannot operate the interface and repeat the same action. The failures of v-GPT-5.4, in contrast, are mostly belief failures ($66\%$). Aggregated over agents (\cref{fig:failure-landscape}(c)), $75\%$ of the failures of text-based agents are belief failures, $7\%$ are action failures, and $18\%$ are overreach. For GUI-only agents, the corresponding shares are $42\%$, $57\%$, and less than $1\%$. As a share of failures, \texttt{misleading\_success\_taken} is most frequent on backtracking and grounding variants (\cref{app:mode-by-primitive}).

\noindent\textbf{Belief failures dominate.}
Across the $3{,}037$ failed intervention runs (\cref{tab:belief-vs-action}), the split between the three classes depends on the harness. For the text-based agents, belief failures account for $62$--$83\%$ of failures, and action failures reach at most $25\%$ (Opus-4.7). For the GUI-only agents, action failures account for $33$--$72\%$ of failures and overreach for less than $1\%$, because these agents rarely reach the point where all positive checks pass. Text-based agents thus usually stop although the task is not done, whereas GUI-only agents often never reach a completion action. These results suggest different remedies. A longer step budget is unlikely to help text-based agents, which already end their runs themselves; re-reading the relevant website state before declaring the run finished is more promising. GUI-only agents first need better discovery of the interface elements they can act on; only then does verification become the bottleneck.

\noindent\textbf{Run length and calibration.}
On backtracking variants, failed intervention runs are longer than passed ones for five of the six text-based agents (by $2.4$ steps on average), which suggests that agents attempt to recover but do not succeed (\cref{app:step-economy-primitive}). If a deployed system treats an agent's completion action as evidence of success, it should know how often that signal is wrong. Calibration of self-reported correctness has been studied in the single-turn setting~\citep{kadavath2022know,tian2023justask}; \cref{fig:failure-landscape}(b) and the FPR column of \cref{tab:belief-vs-action} measure its counterpart for agents. Four of the six text-based agents end $99$--$100\%$ of intervention runs with the completion action, and $54$--$85\%$ of those runs fail. Opus-4.7 and Sonnet-4.6 end only about $65\%$ of runs this way, but they are not more accurate when they do: $55\%$ and $58\%$ of those runs fail. The three GUI-only agents end only $38$--$51\%$ of intervention runs with the completion action, and $65$--$94\%$ of those runs fail. GUI-only agents therefore rarely reach the completion action, and when they do, they are usually wrong.

\noindent\textbf{A representative trace.} \texttt{amazon\_return\_item} shows a typical fabricated-success failure. Gemini-3.1-Pro submits the return, receives a forged success response (HTTP $200$) from the network layer, and declares the run finished at step~$5$ without checking the order status. The matched clean run passes with score $1.0$ within the same step budget, so the paired drop is $1.0$. The variant is labeled backtracking, but the failure is a missing re-read after a forged success. \Cref{app:cases} presents further cases with the same pattern.


\section{Human Evaluation}
\label{sec:human}

Human trajectories serve three roles in \benchmark{}. First, they check that competent users can solve the selected tasks in both the clean and the intervention condition. Second, they estimate the human cost of an intervention, which separates an intervention that only agents find difficult from one that is intrinsically hard. Third, warm human attempts (defined below) provide an efficiency reference for successful agent runs. Because recording humans on all $519$ base tasks is impractical, we use \emph{Human-140}, a subset of $140$ base tasks with four tasks for each combination of environment and difficulty level. Every task is recorded in both the clean and the intervention condition. In each condition, the annotator assigned to the task attempts it twice: a \emph{cold} attempt with only the user-facing instruction, and a \emph{warm} attempt that immediately repeats the same task and condition after the website is reset, whether or not the cold attempt succeeded. \Cref{app:human-protocol} documents the assignment of annotators, the trace-cleaning pipeline, the post-task rating rubric, and a duplicate audit.

\begin{figure}[t]
\centering
\begin{subfigure}[t]{0.46\linewidth}
\centering
\includegraphics[width=\linewidth]{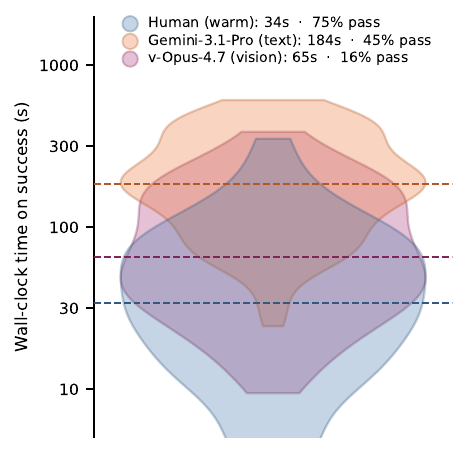}
\end{subfigure}\hfill
\begin{subfigure}[t]{0.52\linewidth}
\centering
\includegraphics[width=\linewidth]{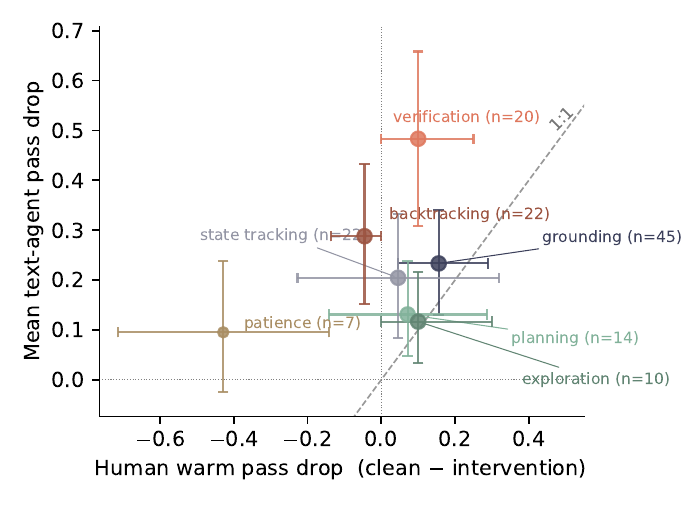}
\end{subfigure}
\caption{Human-vs-agent comparison on Human-140 intervention runs. \textbf{(a)}~Time distribution of successful intervention runs (overlapping violins; dashed lines mark medians) and intervention pass rate (legend). \textbf{(b)}~Per-primitive human warm pass drop versus mean agent pass drop, with bootstrap $95\%$ confidence intervals over Human-140 base tasks; the dashed line marks the $1{:}1$ diagonal.}
\label{fig:human-row}
\vspace{-14pt}
\end{figure}

\begin{wraptable}{r}{0.4\linewidth}
\vspace{-1.0em}
\centering
\caption{Human-140 panel summary. \emph{Time} is median wall-clock seconds; \emph{Events} is median raw browser events.}
\label{tab:human-headline}
\small
\setlength{\tabcolsep}{4pt}
\begin{tabular}{llrrr}
\toprule
Cond.\ & Attempt & Pass & Time & Events\ \\
\midrule
clean   & cold & 76.4\% & 75.7 & 37 \\
clean   & warm & 80.7\% & 32.3 & 26.5 \\
interv.\ & cold & 66.4\% & 82.0 & 47 \\
interv.\ & warm & 75.0\% & 39.7 & 32 \\
\bottomrule
\end{tabular}
\vspace{-0.5em}
\end{wraptable}
\noindent\textbf{Interventions impose a modest cost on humans.} The cold pass rate drops from $76.4\%$ on clean tasks to $66.4\%$ under intervention ($-10.0$ points), and the warm pass rate drops from $80.7\%$ to $75.0\%$ ($-5.7$ points). For comparison, the same interventions lower the pass rates of the six text-based agents by $17.9$--$27.6$ points (\cref{tab:headline}), a drop three to five times larger than that of warm humans. Median wall-clock time grows by $8\%$ (cold) or $23\%$ (warm), and the median number of raw browser events by $27\%$ (cold) or $21\%$ (warm). For humans, the cost of an intervention therefore appears mainly as extra actions and time rather than as failures.
In the warm intervention condition, failed attempts contain about $25$ more raw browser events than successful ones, which suggests that humans, like agents, try to recover rather than give up.
Humans need about half as much time on the warm attempt in both conditions, and their pass rate gains more from the second attempt under intervention ($+8.6$ points) than in the clean condition ($+4.3$ points). This suggests that recovery strategy is what humans improve most clearly on the second attempt; agents in the current harness have no comparable mechanism.

\noindent\textbf{Interventions affect humans much less than agents.} \Cref{fig:human-row}(a) compares warm humans with the strongest agent, Gemini-3.1-Pro ($44.5\%$ intervention pass rate). Even on clean tasks, warm humans outperform Gemini-3.1-Pro ($80.7\%$ vs.\ $72.1\%$); under intervention, the gap grows to $30.5$ points ($75.0\%$ vs.\ $44.5\%$), and on runs that both complete, humans are about five times faster (median $34$\,s vs.\ $184$\,s). The other text-based agents pass fewer intervention runs ($15$--$43\%$), and Opus-4.7 and Sonnet-4.6 are also slower (median $418$\,s and $500$\,s). Grouping the $140$ paired tasks by the primary primitive of the intervention (\cref{fig:human-row}(b)) shows where the gap is largest. All seven primitives lie above the $1{:}1$ diagonal: text-based agents lose more pass rate than warm humans on every primitive. The difference is largest on patience ($+52.4$ points, $n=7$), verification ($+38.3$, $n=20$), and backtracking ($+33.3$, $n=22$), where warm humans even gain $4.5$ points. It is intermediate on state tracking ($+15.9$, $n=22$) and grounding ($+7.8$, $n=45$), and smallest on planning ($+6.0$, $n=14$) and exploration ($+1.7$, $n=10$). Estimates based on few tasks are uncertain; \cref{fig:human-vs-agent-tax-pertask-app} shows the variation across tasks.


\section{Related Work}
\label{sec:related}

\noindent\textbf{Web and computer-use agent benchmarks.} Most benchmarks for web and computer-use agents measure overall task completion and broaden the set of domains, from web micro-tasks in MiniWoB~\citep{DBLP:conf/icml/ShiKFHL17,DBLP:conf/iclr/LiuGPSL18} to full operating systems in OSWorld~\citep{DBLP:conf/nips/XieZCLZCHCSLLXZ24}, with WebShop~\citep{DBLP:conf/nips/Yao0YN22}, WebArena and VisualWebArena~\citep{DBLP:conf/iclr/ZhouX0ZLSCOBF0N24,DBLP:conf/acl/KohLJDLHNZSF24}, Mind2Web~\citep{DBLP:conf/nips/DengGZCSWSS23,xue2025illusion}, and WorkArena~\citep{DBLP:conf/icml/DrouinGCLVM0CL24} in between. BrowserGym~\citep{DBLP:journals/tmlr/ChezellesGLCDBT25} unifies several of them in one harness, and WebVoyager~\citep{DBLP:conf/acl/HeYM0D0L024}, SeeClick~\citep{DBLP:conf/acl/ChengSCX0Z024}, and VisualAgentBench~\citep{DBLP:conf/iclr/LiuZ0ISXZLSYYQY25} target multimodal grounding. \benchmark{} instead keeps the base task and its success criterion fixed, makes the task harder by intervening on the environment, and scores the outcome on the backend state.

\noindent\textbf{Capability decomposition and stress-testing benchmarks.} \citet{shlomov2024grounding} score planning and grounding separately from labeled Mind2Web traces, and Web-CogReasoner~\citep{guo2025webcogreasoner} uses capability decomposition as a training curriculum. AgentBench~\citep{DBLP:conf/iclr/0036YZXLL0DMYZ024}, AgentBoard~\citep{DBLP:conf/nips/MaZZYYJLKH24}, OSWorld's failure analysis~\citep{DBLP:conf/nips/XieZCLZCHCSLLXZ24}, and \citet{riddell2026stalled} derive scores for individual capabilities by analyzing trajectories after they have been collected. ReliabilityBench~\citep{gupta2026reliabilitybench} studies repeated-run consistency, instruction perturbations, and tool faults such as timeouts and rate limits. ST-WebAgentBench~\citep{levy2025stwebagentbench} evaluates task completion and policy adherence in enterprise settings, including error handling and prompt injection from the environment. Indirect prompt injection~\citep{greshake2023indirect}, tested by AgentDojo~\citep{DBLP:conf/nips/DebenedettiZBB024}, InjecAgent~\citep{DBLP:conf/acl/ZhanLYK24}, and WIPI~\citep{wu2024wipi}, measures resistance to untrusted instructions embedded in external content; AgentDojo also compares task utility with and without attacks using checks of the environment state. Unlike these benchmarks, \benchmark{} keeps the instruction and the backend success criterion fixed and changes only the website, at one of four layers (seed, server, network, client). Every change is recoverable, labeled with one of the seven primitives, and compared with the same task on the unmodified website. Prompt injection is one mechanism in our catalog and follows the same design rules.

\noindent\textbf{Agent harnesses.} We evaluate vision-language models~\citep{anthropic2026opus47,anthropic2026sonnet46,openai2026gpt54,openai2026gpt54minino,google2025gemini3,google2025gemini3flash,deepmind2026gemini31pro} with a text-based harness that serializes the DOM (the \texttt{browser-use} library~\citep{browser_use2024}) and with a GUI-only harness that shows rendered screenshots (BrowserGym~\citep{DBLP:journals/tmlr/ChezellesGLCDBT25}). Open agentic models follow the same design~\citep{kimiteam2025k2,kimiteam2026k25,qwenteam2025qwen3vl}, and \citet{xi2023agentsurvey} survey the broader design space. \benchmark{} evaluates each combination of model and harness as a complete system and reports each paired drop under the primary primitive of the variant.

\section{Conclusion}
\label{sec:conclusion}

\benchmark{} constructs harder tasks through controlled, recoverable interventions that keep the task and its success criterion fixed. Instead of a single end-to-end score, it reports which interventions make agents fail on tasks they solve in the clean condition. Agents with the two harnesses fail in different ways, which points to two bottlenecks. Text-based agents need to verify the \emph{external} state after acting and before declaring the run finished: of the intervention runs that the six text-based agents declare finished, $54$--$85\%$ have in fact failed. GUI-only agents first need to find the elements of the rendered page that they can act on; only then does verification become relevant. With either harness, current agents are much less robust to these interventions than humans are.
Labeling interventions by primitive and by layer suggests two directions for future work. First, the same catalog can serve as a training curriculum with interventions that mirror those used at test time, so that agents learn during training to verify their actions and to backtrack. Second, exploration should be retested on easier base tasks, where the clean baseline is high enough for a drop to be visible. Both directions can reuse the existing intervention families without changes to the environments.

\noindent\textbf{Limitations.} Our evaluation uses one text-based harness (\texttt{browser-use}) and one GUI-only harness (BrowserGym) on seven English consumer websites, and the two open-weight models are evaluated only with the text-based harness. Extending the catalog to enterprise and non-English settings is left to future work.

\bibliography{custom}
\bibliographystyle{plainnat}

\newpage
\appendix
\section*{Appendix}
\section{Environment Details}
\label{app:env}

\subsection{Environment Infrastructure}
\label{app:env-infra}

\benchmark{} is a single self-hosted FastAPI application that serves all seven environments behind one process. Each environment is a triple of a React single-page application, a typed Pydantic state model, and a router under \texttt{/api/env/<env\_id>} that mounts the read and write endpoints the SPA consumes. The environments cover email (Gmail), finance (Robinhood), e-commerce (Amazon), social (Reddit), healthcare (a patient portal), education (an LMS), and travel (Booking).

A run is keyed by an opaque \texttt{session\_id} created with an explicit task and integer seed. The session manager constructs a fresh state object, runs the task's seed builder (\cref{app:task-pipeline}), and stores both the live state and an immutable initial snapshot for evaluation; the same \texttt{(task\_id, seed)} pair always yields byte-identical initial state.

Each environment exposes two disjoint endpoint classes. Public endpoints under \texttt{/api/env/<env\_id>} are the only surface the agent and the SPA ever touch: \texttt{GET} routes return Pydantic models serialized to JSON, and \texttt{POST}/\texttt{PATCH}/\texttt{DELETE} routes mutate state through typed handlers. Controller endpoints under \texttt{/control/<env\_id>/<session\_id>} are protected by a per-process secret and reserved for the harness: applying interventions, dumping the canonical diff, resetting the session, and reading audit logs. The agent never holds the controller secret, so it can neither bypass the public API nor read the latent target.

The diff used by the canonical-diff evaluator (\cref{sec:eval-canonical}) is computed on the initial and final snapshots: it returns \textsc{Create}, \textsc{Delete}, and \textsc{Update} records with per-field before/after pairs, sorted by entity ID for deterministic comparison. Audit logs are kept only for debugging and for the trajectory features in \cref{sec:analysis}; they do not enter scoring. Each React SPA is served from \texttt{/env/<env\_id>} on the same code path an end user would exercise; no test IDs or agent-facing DOM hooks are added, and client-layer interventions (\cref{app:catalog}) are applied by a React component compiled into each SPA at build time, which stays inert unless the session registers a dispatch list and exposes no evaluator state or latent target to the agent.

\subsection{Observation Space}
\label{app:obs-space}

\benchmark{} supports two harnesses with different observation contracts: a text-based harness built on the \texttt{browser-use} library~\citep{browser_use2024} and a GUI-only harness implemented in BrowserGym~\citep{DBLP:journals/tmlr/ChezellesGLCDBT25}. The two share the same backend, task and variant catalog, and evaluator; only the observations that the agent receives differ.

\subsubsection{Text-Based Harness}
\label{app:obs-text}

Each step the text-based agent receives the user-facing instruction, the last action and any error string, the current URL, and a flattened accessibility tree of the page. The tree is filtered to the visible, addressable, clickable subset (\cref{tab:axtree-filter}); every interactive node carries a short identifier (\emph{bid}, e.g.\ \texttt{a51}). The agent must address each action by \texttt{bid} rather than by coordinates or CSS selectors, which fixes the action grammar (\cref{app:action-space}) and rules out a class of selector-hallucination failures by construction. The observation budget is capped at $26{,}000$ input tokens; when the conversation would exceed that budget, the harness drops the oldest turns and replaces them with a short fact summary so the system prompt and the latest observation always remain in scope.

\Cref{tab:obs-text-example} reproduces a single text-based harness observation step on the Amazon environment, showing how the four header lines compose with a flattened accessibility-tree fragment.

\begin{table}[ht]
\centering
\caption{Example text-based harness observation on \texttt{amazon\_browse\_category}, step~$3$. The header carries goal, last action, last action error, and URL; the body is the flattened accessibility tree filtered by the rules in \cref{tab:axtree-filter}. Indentation marks DOM nesting; bracketed ids (e.g.\ \texttt{[a51]}) are the addressable \texttt{bid} strings.}
\label{tab:obs-text-example}
\small
\begin{tabular}{p{0.96\linewidth}}
\toprule
\verb|## goal: Browse the Electronics category, find the cheapest item available, and | \\
\verb|   add it to your cart.| \\
\verb|## last_action: click('a14')|\\
\verb|## last_action_error:|\\
\verb|## url: http://localhost:8000/env/amazon/category/electronics?sort=price-asc|\\
\midrule
\verb|RootWebArea "Amazon -- Electronics"|\\
\verb|  [a3] navigation "Primary"|\\
\verb|    [a5] link "Home"|\\
\verb|    [a6] link "Cart (0)"|\\
\verb|  [a10] heading "Electronics"|\\
\verb|  [a11] combobox "Sort by" value="Price: low to high"|\\
\verb|  [a20] list|\\
\verb|    [a21] listitem|\\
\verb|      [a22] link "USB-C Charging Cable 6ft"|\\
\verb|      [a23] StaticText "$7.99"|\\
\verb|      [a24] StaticText "4.3 (820 reviews)"|\\
\verb|      [a25] button "Add to cart" clickable|\\
\verb|    [a31] listitem|\\
\verb|      [a32] link "Wireless Bluetooth Speaker"|\\
\verb|      [a33] StaticText "$24.99"|\\
\verb|      [a35] button "Add to cart" clickable|\\
\verb|    ...|\\
\bottomrule
\end{tabular}
\end{table}

\subsubsection{GUI-Only Harness}
\label{app:obs-pixel}

The GUI-only agent receives the user-facing instruction and, at each step, the rendered viewport screenshot as a base64-encoded PNG. No DOM, accessibility tree, or element list is provided. The last action, action error, and URL fields exposed to the text-based agent are written to the trajectory log for evaluation but not shown to the model.

Vision-language models are trained on images of particular sizes, and their grounding accuracy can degrade at very different resolutions. The harness therefore uses the viewport that each provider recommends for its models (\cref{tab:viewport}).

\begin{table}[ht]
\centering
\caption{Per-model viewport for the GUI-only harness. Each viewport follows the provider's documented recommendation. Image scale (low, medium, or high) refers to the quality settings of each API.}
\label{tab:viewport}
\small
\begin{tabular}{llrr}
\toprule
Model family & Provider docs cited & Viewport (w$\times$h) & Aspect \\
\midrule
Gemini-3.x, Qwen3-VL  & default                   & $1280 \times 720$ & 16:9 \\
Claude (Opus, Sonnet) & Anthropic computer-use   & $1024 \times 768$ & 4:3  \\
GPT-5.x, GPT-4o       & OpenAI CUA               & $1600 \times 900$ & 16:9 \\
\bottomrule
\end{tabular}
\end{table}

Anthropic and OpenAI VLMs are prompted with raw pixel coordinates; Gemini and Qwen are prompted with a $0$--$1000$ normalized grid that the harness rescales to pixels at action time. The latter convention follows the coordinate format that these providers document for their models.

\subsection{Action Space}
\label{app:action-space}

\benchmark{} reuses the BrowserGym action grammars, with one Python-call action per step. Two grammars are used: a BID grammar for the text-based harness, where the agent addresses elements by their accessibility-tree \texttt{bid} string (\cref{tab:action-text}), and a coordinate grammar for the GUI-only harness, where the agent emits $(x, y)$ targets (\cref{tab:action-pixel}). Both share the terminal verbs \texttt{send\_msg\_to\_user} and \texttt{report\_infeasible}; \texttt{noop} only waits.

\begin{table}[ht]
\centering
\caption{BID action grammar used by the text-based harness. Every \texttt{bid} argument is a quoted string that must come from the current observation; the harness rejects calls that pass a numeric \texttt{bid} or an out-of-scope identifier.}
\label{tab:action-text}
\small
\begin{tabular}{p{0.31\linewidth}p{0.62\linewidth}}
\toprule
Action call & Effect \\
\midrule
\texttt{click('a51')}                          & Click the element with bid \texttt{a51}. \\
\texttt{dblclick('12')}                        & Double-click the element. \\
\texttt{fill('b22', 'hello')}                  & Fill a text field with the given string. \\
\texttt{clear('b22')}                          & Clear a text field. \\
\texttt{focus('b22')}                          & Focus an element without clicking. \\
\texttt{hover('d7')}                           & Hover over an element. \\
\texttt{select\_option('c3', 'California')}    & Pick an option in a \texttt{<select>} element. \\
\texttt{press('48', 'Enter')}                  & Press a key while the named element is focused. \\
\texttt{scroll(0, 300)}                        & Scroll by $(\Delta x, \Delta y)$ pixels (positive = down/right). \\
\texttt{drag\_and\_drop('a1', 'b2')}           & Drag bid \texttt{a1} to bid \texttt{b2}. \\
\midrule
\texttt{send\_msg\_to\_user('done')}           & Report a final answer or declare completion (terminates the episode). \\
\texttt{report\_infeasible('reason')}          & Declare the task is impossible (terminates the episode). \\
\texttt{noop(1000)}                            & Wait for $1000$\,ms before the next observation. \\
\bottomrule
\end{tabular}
\end{table}

\begin{table}[ht]
\centering
\caption{Coord action grammar used by the GUI-only harness. Coordinates are pixels for Anthropic and OpenAI models and a $0$--$1000$ normalized grid for Gemini and Qwen models (\cref{app:obs-pixel}).}
\label{tab:action-pixel}
\small
\begin{tabular}{p{0.42\linewidth}p{0.51\linewidth}}
\toprule
Action call & Effect \\
\midrule
\texttt{mouse\_click(x, y)}                          & Click at $(x, y)$. \\
\texttt{mouse\_dblclick(x, y)}                       & Double-click at $(x, y)$. \\
\texttt{mouse\_move(x, y)}                           & Move the pointer without clicking. \\
\texttt{mouse\_drag\_and\_drop(x1, y1, x2, y2)}      & Drag from $(x_1, y_1)$ to $(x_2, y_2)$. \\
\texttt{keyboard\_type('text')}                      & Type the given string into the focused element. \\
\texttt{keyboard\_press('Enter')}                    & Press a single key. \\
\texttt{scroll(dx, dy)}                              & Scroll by pixels (positive $dy$ = down). \\
\texttt{go\_back} / \texttt{go\_forward}             & Browser back/forward. \\
\midrule
\texttt{send\_msg\_to\_user('done')}                 & Report a final answer or declare completion (terminates the episode). \\
\texttt{report\_infeasible('reason')}                & Declare the task is impossible (terminates the episode). \\
\texttt{noop(1000)}                                  & Wait for $1000$\,ms before the next observation. \\
\bottomrule
\end{tabular}
\end{table}

\noindent\textbf{Episode termination.}
An episode ends when the agent issues a terminal verb, when the action or wall-clock budget is exhausted, or when the harness raises an unrecoverable error. Every run in our evaluation uses a fixed 40-action cap and a model-specific wall-clock cap between 600 and 1,200\,s; both caps are identical for the clean and intervention conditions of a base task. The task YAML's \texttt{expected\_steps} and \texttt{time\_limit\_seconds} fields describe task-design expectations and difficulty rather than setting the submitted-sweep caps. Each LLM call has an additional $120$\,s timeout. Budget exhaustion is not automatically scored as failure: the canonical-diff evaluator scores the final backend state regardless of how the trajectory terminated, so an agent that completed the task before running out of steps still earns full credit.

\section{Benchmark Details}
\label{app:bench}
\label{sec:building}

\subsection{Tasks and Canonical-Diff Scoring}
\label{app:eval-canonical}
\label{sec:eval-canonical}
\label{app:tasks}

Each task is generated from an environment-specific seed program. For environment $e$, a seed $z$ induces
\[
G_e(z) \rightarrow (s_0, \tau, x),
\]
where $s_0$ is the initial backend state, $\tau$ is a hidden target structure used only for evaluation, and $x$ is the user-facing instruction. The agent interacts with the rendered UI $R_e(s_t)$ and never observes $\tau$; the evaluator uses $\tau$ to score the change between the initial and final backend states. Tasks are specified by their intended semantic effect, not by a reference UI trajectory: a Gmail triage task may be solved by search, by labels, or by per-thread navigation; a portfolio task may place the required orders in any order. Each seed builder records the latent target alongside the visible state, so the agent must infer $\tau$ through the UI while the evaluator scores against $\tau$ exactly.

We grade a trajectory by comparing the structural diff $\Delta(s_0, s_T) \in \{\textsc{Create}, \textsc{Update}, \textsc{Delete}\}^{*}$ produced by the trajectory against a canonical diff $D=(P,N)$ declared by the task. $P$ is a set of positive obligations (required creates, updates, or deletes specified as predicates over entity fields); $N$ is a set of invariants on protected state. The final score separates accomplishment from safety:
\[
\mathrm{score}
=
\mathrm{clip}_{[0,1]}
\left(
\frac{\sum_{p\in P} w_p \rho_p}{\sum_{p\in P} w_p}
-
\sum_{n\in N:\neg n} \lambda_n
\right),
\]
where $\rho_p\in[0,1]$ is the coverage of obligation $p$ and $\lambda_n$ the severity-weighted penalty for an invariant violation. Invariants are penalty-only: an idle agent earns no credit for avoiding collateral damage.

A singleton obligation is satisfied if some diff entry on the right collection meets the field predicates. Set-valued obligations require one correct operation per member of a hidden target set (mark every unread email, place one order per passing stock); they must be permutation-invariant but cardinality-sensitive. We score them by maximum bipartite matching between target slots $V=\{v_1,\ldots,v_m\}$ and candidate diff entries $C=\{c_1,\ldots,c_n\}$, with an edge $(v_i, c_j)$ whenever $c_j$ satisfies the clause predicates with the loop variable bound to $v_i$. The clause passes iff $|M^\star|=m$, with partial credit $\rho_p = |M^\star|/m$. The bipartite formulation is invariant to UI-generated IDs and operation order while rejecting missing entries, duplicated targets, and numerically incorrect fields. A weaker aggregate check such as ``there are $k$ new entries'' would accept duplicates or off-target entries.

\subsubsection{Worked Example}
\label{app:canonical-example}
\label{app:matching}

Consider an unread-message task with target IDs $\{e_1,\ldots,e_5\}$. If the agent marks $e_1, e_2, e_3, e_4$ as read, the matching has size four: partial credit, but the clause fails. If the agent additionally toggles an already-read distractor, the positive matching still only covers the true slots, and an invariant on non-target messages penalizes the collateral mutation. If the agent marks all five target messages in any order, the task passes because the matching is defined over semantic diff entries rather than action order.

\subsection{Per-Environment Specifications}
\label{app:per-env}
\label{app:per-env-eval}

Every \benchmark{} environment is a self-hosted clone of a real production platform, paired with a typed Pydantic state model and the read/write endpoints the SPA consumes. \Cref{tab:per-env-spec} reports the per-environment task and endpoint counts and the entity collections that intervention variants and canonical-diff clauses most often touch. Because every collection is a list of typed Pydantic models, the same predicate grammar (\cref{app:predicate-grammar}) applies across environments without per-environment customization.

\begin{table}[t]
\centering
\caption{Per-environment summary. \emph{Tasks} is the count of base tasks (each paired one-to-one with a variant). \emph{Endpoints} counts the public read/write routes the SPA consumes. \emph{Graded collections} lists the entity collections most often referenced by canonical-diff clauses; entity types are the corresponding Pydantic classes.}
\label{tab:per-env-spec}
\label{tab:eval-strategy}
\small
\setlength{\tabcolsep}{4pt}
\begin{tabular}{llrrp{0.46\linewidth}}
\toprule
Environment & Domain & Tasks & Endpts.\ & Graded collections (entity type) \\
\midrule
Gmail          & Email      & 84 & 33 & \texttt{emails} (\texttt{Email}), \texttt{drafts}, \texttt{sent}, \texttt{labels}, \texttt{filters}, \texttt{contacts} \\
Amazon         & E-commerce & 70 & 56 & \texttt{cart\_items}, \texttt{orders} (\texttt{Order}), \texttt{returns}, \texttt{addresses}, \texttt{payment\_methods}, \texttt{reviews}, \texttt{wishlist} \\
Reddit         & Social     & 81 & 45 & \texttt{posts} (\texttt{Post}), \texttt{comments}, \texttt{messages}, \texttt{notifications}, \texttt{subscriptions}, \texttt{saved\_post\_ids} \\
Robinhood      & Finance    & 71 & 57 & \texttt{orders}, \texttt{options\_orders}, \texttt{positions}, \texttt{watchlists}, \texttt{transfers}, \texttt{recurring\_investments} \\
Booking        & Travel     & 78 & 53 & \texttt{reservations} (\texttt{Reservation}), \texttt{reviews}, \texttt{saved\_lists}, \texttt{messages}, \texttt{transactions} \\
LMS            & Education  & 65 & 46 & \texttt{assignments}, \texttt{grades}, \texttt{discussions}, \texttt{discussion\_posts}, \texttt{peer\_reviews}, \texttt{enrollments} \\
Patient Portal & Healthcare & 70 & 45 & \texttt{appointments}, \texttt{prescriptions}, \texttt{lab\_results}, \texttt{messages}, \texttt{referrals}, \texttt{claims} \\
\midrule
\textbf{Total} & 7 domains  & \textbf{519} & \textbf{335} & \textbf{130 collections} \\
\bottomrule
\end{tabular}
\end{table}

In total, the seven environments expose $335$ public endpoints and $130$ state collections. Each environment's API is realistic in scale: Amazon spans product browse, cart, checkout, orders, returns, addresses, payment methods, wishlist, reviews, questions, and gift cards; Robinhood spans equity orders, options orders, watchlists, transfers, recurring investments, tax documents, and price alerts.

\subsection{Difficulty Taxonomy}
\label{app:difficulty}

Every base task declares a difficulty level on a five-point scale recorded as the \texttt{difficulty} field of its YAML. The taxonomy is informed by the expected action and time fields the YAML carries, by the number of state collections the canonical diff touches, and by whether the task carries any set-valued obligation (\cref{sec:eval-canonical}); these fields are design metadata rather than the fixed caps used in our evaluation.

\begin{table}[h]
\centering
\caption{Difficulty taxonomy. \emph{Median expected steps} and \emph{median time limit} are computed over all tasks at that level across the seven environments.}
\label{tab:difficulty}
\small
\begin{tabular}{lrrrp{0.42\linewidth}}
\toprule
Level & Tasks & Med.\ steps & Med.\ time (s) & Operational criterion \\
\midrule
easy     &  79 & 10 & 180 & Single-action goal on one collection (e.g.\ mark one email as read). \\
medium   & 118 & 15 & 240 & Sequential edit on one or two collections; no set-valued obligation. \\
hard     & 132 & 22 & 360 & Multi-collection goal with at least one set-valued obligation. \\
expert   &  98 & 30 & 420 & Multi-collection goal that requires reading derived state (filtered list, computed total, ranking). \\
frontier &  92 & 45 & 540 & Cross-collection long-horizon goal with $\ge$3 obligations and at least one negative invariant on adjacent state. \\
\bottomrule
\end{tabular}
\end{table}

All seven environments contain tasks at every difficulty level: every environment contributes between $7$ and $19$ \texttt{easy} tasks, $14$ and $22$ \texttt{medium}, $15$ and $33$ \texttt{hard}, $12$ and $15$ \texttt{expert}, and $10$ and $15$ \texttt{frontier}. The Human-140 panel (\cref{sec:human-panel}) is balanced over (environment, difficulty) cells with exactly four base tasks per cell, so per-difficulty and per-environment human comparisons are well-defined. \texttt{expected\_steps} is a design-time estimate used to assign tiers; it does not set the run budget. Every run uses the same 40-action cap in both conditions, and the cap terminates fewer than $1\%$ of runs (27 of 3,114 per condition), so the frontier tier's median estimate of 45 steps does not translate into cap-truncated runs; the budget audit and an enlarged-budget rerun are reported in \cref{app:baselines}.

\subsection{Task Generation Pipeline}
\label{app:task-pipeline}

Each task is a YAML file with a fixed schema. A task carries seven top-level fields (\texttt{task\_id}, \texttt{env\_id}, \texttt{title}, \texttt{instruction\_template}, \texttt{difficulty}, \texttt{time\_limit\_seconds}, \texttt{expected\_steps}), a \texttt{primary\_primitives} list, a \texttt{seed} block, and a \texttt{canonical\_diff} block. The seed pipeline is a three-stage deterministic procedure executed when a session is created: a fresh state object is constructed with per-environment defaults (currency, time zone, owner profile); the YAML's \texttt{seed.steps} list is interpreted in order, with each step calling a registered builder (\texttt{featured\_product}, \texttt{three\_party\_thread}) that mutates state and records named outputs into a shared context; finally, \texttt{seed.targets} binds latent target variables to expressions over those outputs (e.g., \texttt{cheapest\_id: "\{output.product\_id\}"}), which canonical-diff predicates then read as \texttt{target['key']}. All randomness comes from a Python pseudo-random generator initialized from the integer seed.

\Cref{lst:worked-task} reproduces \texttt{amazon\_browse\_category} verbatim. Because \texttt{cheapest\_id} is bound to the featured product's runtime ID rather than to a literal string, the task remains correct under any reseeding that changes which product receives that ID.

\begin{figure}[t]
\centering
\begin{minipage}{0.96\linewidth}
\hrule\vspace{0.4em}\small
\begin{verbatim}
task_id: amazon_browse_category          # unique within the env
env_id:  amazon                          # registers the task with the Amazon env
title:   Browse Category and Add Cheapest Item
instruction_template: >                  # user-facing prompt
  Browse the Electronics category, find the cheapest item available,
  and add it to your cart.
difficulty: easy                         # see App.~B.2
time_limit_seconds: 150                  # task-design time estimate
expected_steps:    10                    # task-design action estimate
primary_primitives: [grounding]          # one or two from the seven (App.~B.7)
start_path: /                            # initial URL relative to /env/amazon

seed:                                    # deterministic init program
  distractors: 8                         # number of unrelated products
  actors: {}                             # named persons (none for this task)
  steps:
    - use: featured_product              # builder #1: one cheap target
      params: {name: USB-C Charging Cable 6ft, brand: CableTech,
               category: Electronics, price: 7.99, rating: 4.3,
               features: [Fast charging support, Braided nylon cable,
                          6 foot length], variants: [], in_stock: true}
      outputs: [product_id, product_name, product_price]
    - use: product_catalog               # builder #2: six distractors
      params: {category: Electronics, count: 6,
               price_range: [15.0, 200.0]}
      outputs: [product_ids]
  targets:                               # latent ground truth
    cheapest_id:   "{output.product_id}"
    cheapest_name: "{output.product_name}"

canonical_diff:                          # graded against backend state
  create:                                # one positive obligation
    - entity: CartItem
      desc:   Cheapest Electronics product added to cart
      properties:                        # field-level predicates
        product_id:   {expr: "x == target['cheapest_id']"}
        quantity:     {eq: 1}
        product_name: {expr: "x == target['cheapest_name']"}
        unit_price:   {any: true}        # auto-populated from product
        variant_selections: {any: true}
        added_at:     {any: true}        # server-set timestamp
  invariant:                             # protected collateral state
    - {collection: state.cart_items,
       filter: "a.product_id != target['cheapest_id']", preserve: ALL}
    - {collection: state.products,        preserve: ALL}
    - {collection: state.addresses,       preserve: ALL}
    - {collection: state.payment_methods, preserve: ALL}
    - {collection: state.orders,          preserve: ALL}
    - {collection: state.returns,         preserve: ALL}
\end{verbatim}
\vspace{-0.3em}\hrule
\end{minipage}
\caption{The \texttt{amazon\_browse\_category} task YAML, reproduced verbatim. Field-level comments are this paper's; the file itself is unannotated.}
\label{lst:worked-task}
\end{figure}

\subsection{Predicate Grammar}
\label{app:predicate-grammar}

A canonical-diff clause is a tree of predicates. Every leaf is a single-key mapping; the key picks one of $19$ admitted verbs and the value supplies its argument. Predicates that take an inner predicate (\texttt{fields}, \texttt{length}, \texttt{not}, \texttt{all\_of}, \texttt{any\_of}) recurse with a shifted scope, so the grammar is fully nestable. \Cref{tab:predicate-grammar} lists every key with its semantics and a typical use site. The \texttt{expr} key is the only predicate admitting Python-like syntax; it evaluates under an AST allowlist that forbids dunder attribute access, imports, and most builtins, exposing only \texttt{x} (the bound entity), \texttt{target} (latent ground truth), \texttt{initial} and \texttt{state} (initial and final state), \texttt{v} (the bipartite-matching loop variable), and \texttt{session\_start}. All other predicates are pure data and require no sandbox.

\begin{table}[t]
\centering
\caption{Canonical-diff predicate grammar. The four scalar predicates and four collection predicates compose with the four text predicates and the five logical/structural combinators to grade any field on any Pydantic state model. The \texttt{matches\_semantic} predicate uses a $0.8$-threshold sequence-matcher ratio; \texttt{eq} uses a fuzzy equality that snaps numeric strings.}
\label{tab:predicate-grammar}
\small
\begin{tabular}{llp{0.34\linewidth}p{0.26\linewidth}}
\toprule
Class & Key & Semantics & Example \\
\midrule
\multirow{5}{*}{Scalar}
 & \texttt{eq}                & exact (or fuzzy numeric) match & \texttt{quantity: \{eq: 1\}} \\
 & \texttt{in}                & membership in a literal list   & \texttt{status: \{in: [paid, pending]\}} \\
 & \texttt{between}           & numeric range, inclusive       & \texttt{rating: \{between: [4, 5]\}} \\
 & \texttt{any}               & always true (placeholder)      & \texttt{added\_at: \{any: true\}} \\
 & \texttt{expr}               & sandboxed Python expression on \texttt{x}/\texttt{target}/\texttt{state} & \texttt{\{expr: "x == target['id']"\}} \\
\midrule
\multirow{5}{*}{Collection}
 & \texttt{set\_eq}           & set equality                    & \texttt{labels: \{set\_eq: [inbox, work]\}} \\
 & \texttt{subset}            & subset of literal set           & \texttt{labels: \{subset: [inbox, sent]\}} \\
 & \texttt{superset}          & superset of literal set         & \texttt{tags: \{superset: [urgent]\}} \\
 & \texttt{contains}          & membership in collection        & \texttt{labels: \{contains: starred\}} \\
 & \texttt{length}            & inner predicate on \texttt{len(x)} & \texttt{items: \{length: \{eq: 3\}\}} \\
\midrule
\multirow{5}{*}{Text}
 & \texttt{substring}         & literal substring               & \texttt{body: \{substring: "Q3"\}} \\
 & \texttt{substring\_all}    & all listed substrings           & \texttt{body: \{substring\_all: [a, b]\}} \\
 & \texttt{substring\_any}    & at least one listed substring   & \texttt{body: \{substring\_any: [yes, no]\}} \\
 & \texttt{regex}             & \texttt{re.search}              & \texttt{subject: \{regex: "RE:.*"\}} \\
 & \texttt{matches\_semantic} & sequence-matcher $\ge$ threshold & \texttt{name: \{matches\_semantic: \{value: J. Doe, threshold: 0.85\}\}} \\
\midrule
Structural
 & \texttt{fields}            & per-field predicate map         & \texttt{\{fields: \{a: \{eq: 1\}\}\}} \\
\midrule
\multirow{3}{*}{Logical}
 & \texttt{not}               & predicate negation              & \texttt{\{not: \{eq: 0\}\}} \\
 & \texttt{all\_of}           & conjunction                      & \texttt{\{all\_of: [\dots]\}} \\
 & \texttt{any\_of}           & disjunction                      & \texttt{\{any\_of: [\dots]\}} \\
\bottomrule
\end{tabular}
\end{table}

\subsection{Task Statistics}
\label{app:bench-stats}

\Cref{tab:bench-stats} reports per-environment instruction-length and task-design statistics. Instruction tokens are computed by whitespace splitting; expected steps and time limits are the YAML metadata used during construction and difficulty calibration, not the fixed termination caps of our evaluation (\cref{app:env}).

\begin{table}[h]
\centering
\caption{Per-environment task statistics. Instruction words are whitespace-split tokens of \texttt{instruction\_template}; expected steps and time limit are task-design estimates stored in the YAML.}
\label{tab:bench-stats}
\small
\begin{tabular}{lrrrrr}
\toprule
Environment & Tasks & Avg.\ inst.\ words & Avg.\ exp.\ steps & Avg.\ time limit (s) \\
\midrule
Gmail          & 84 & 103.4 & 38.9 & 440 \\
Reddit         & 81 &  50.2 & 29.4 & 394 \\
Amazon         & 70 &  46.3 & 28.6 & 358 \\
Booking        & 78 &  95.0 & 24.1 & 346 \\
Patient Portal & 70 &  45.8 & 23.4 & 349 \\
LMS            & 65 &  50.2 & 20.3 & 323 \\
Robinhood      & 71 &  34.2 & 16.3 & 298 \\
\midrule
\textbf{Suite} & \textbf{519} & \textbf{61.0} & \textbf{26.4} & \textbf{358} \\
\bottomrule
\end{tabular}
\end{table}

Gmail and Booking are the most prose-heavy because their instructions describe a multi-stakeholder situation (a forwarded thread, a hotel-comparison rationale), whereas Robinhood and Amazon often issue a one-sentence trade or shopping directive. Design-time step estimates follow the same gradient: Gmail tasks expect $\sim$$39$ actions on average against $\sim$$16$ for Robinhood. The suite mean is $26.4$ expected steps and $358$ expected seconds per task.

\subsection{Quality Control and Primitive Purity}
\label{app:qc}

Every variant declares a single \texttt{target\_primitive} in its YAML manifest and was reviewed by four annotators (at least two per variant) against a primitive-purity rubric: the variant must require the declared primitive and may add at most one primitive to the set that the base task declares (formally, $|T_{\text{task}} \cup \{p_{\text{variant}}\}| \le 2$, where $T_{\text{task}}$ is the base-task primitive set declared in \texttt{primary\_primitives}; in $514$ of the $519$ pairs, the target primitive is already in this set); and it must remain detectable, recoverable, and realistic in the sense of \cref{sec:suite}. Disagreements were sent to a third reviewer; the final \texttt{target\_primitive} is the consensus tag. \Cref{tab:coverage-prim-env} reports the variant-by-environment matrix; most cells contain at least five variants. Four cells are empty (Reddit/planning, Reddit/patience, Reddit/exploration, Patient Portal/patience). These combinations have no variants in the released catalog, so \cref{app:env-cases} reports no results for them.

The construction log records 48 adjudicated flags (9.2\% of variants; 43 on existing items), which measures review coverage rather than per-primitive agreement. On 224 variants without scaffolding, verification and backtracking remain the two largest drops (31.8 and 29.2 points); naturally occurring decoy-only, silent-failure-only, and combined cohorts yield drops of 4.8, 51.5, and 57.1 points, respectively, which we treat as an additivity diagnostic rather than randomized causal evidence.

\begin{table}[h]
\centering
\caption{Variant counts by (environment, target primitive). Empty cells are left blank.}
\label{tab:coverage-prim-env}
\small
\begin{tabular}{lrrrrrrr}
\toprule
Environment & Ground. & Plan. & State & Backtrk. & Patience & Explore & Verify \\
\midrule
Amazon          & 18 &  7 & 13 & 16 &  4 &  3 &  9 \\
Booking         & 35 &  4 &  5 & 10 &  5 &  1 & 18 \\
Gmail           & 25 &  2 & 13 & 16 &  5 &  9 & 14 \\
LMS             & 10 & 17 & 10 &  2 &  2 & 10 & 14 \\
Patient Portal  & 15 & 16 & 11 & 16 &    &  6 &  6 \\
Reddit          & 55 &    & 11 &  9 &    &    &  6 \\
Robinhood       & 11 &  6 & 20 & 11 &  8 &  9 &  6 \\
\midrule
\textbf{Total}  & \textbf{169} & \textbf{52} & \textbf{83} & \textbf{80} & \textbf{24} & \textbf{38} & \textbf{73} \\
\bottomrule
\end{tabular}
\end{table}

\Cref{tab:coverage-layer-prim} reports the layer-by-primitive matrix, with one row per injection layer and one column per target primitive (counts are injection records, summed over multi-layer variants). The matrix is consistent with the per-layer primary-primitive list in \cref{tab:layers}: seed and network are the dominant layers, each with $\ge$$360$ injection records. Seed records concentrate on grounding ($182$) and state tracking ($81$); network records concentrate on backtracking ($89$), verification ($65$), and patience ($38$); server contributes most heavily to planning ($31$) and state tracking ($32$); the client layer is sparse but spread evenly across primitives. No server- or client-layer record targets backtracking: most backtracking variants rely on runtime feedback from the network layer (a $409$ or $5xx$ response on the recovery attempt), but $16$ of them plant seed-layer content (contradictory updates, decoys, or planted wrong answers) that leads the agent down a path it must abandon, and $10$ of these have no network injection.

\begin{table}[h]
\centering
\caption{Injection records by (layer, target primitive). Variants that stack more than one injection layer contribute one record per layer, so the total over all rows exceeds $519$.}
\label{tab:coverage-layer-prim}
\small
\begin{tabular}{lrrrrrrrr}
\toprule
Layer    & Ground. & Plan. & State & Backtrk. & Patience & Explore & Verify & \textbf{Total} \\
\midrule
Seed     & 182 & 18 & 81 & 17 &  1 & 36 & 26 & 361 \\
Server   &  20 & 31 & 32 &  0 &  6 & 11 &  2 & 102 \\
Network  &  68 & 37 & 47 & 89 & 38 & 16 & 65 & 360 \\
Client   &  11 &  4 &  2 &  0 &  6 &  8 &  8 &  39 \\
\bottomrule
\end{tabular}
\end{table}

\section{Intervention Catalog Details}
\label{app:catalog}

\subsection{Design Choices}
\label{app:primitive-choice}
\label{app:invariant-justification}

\noindent\textbf{Why seven primitives.} The set in \cref{sec:primitives} is a coarse set of primitives that still allows each variant to declare a single primary primitive, by which its results are grouped. Coarser sets would merge two distinct deficits (for example, the split between verification and backtracking that separates belief failures of text-based agents from action failures), while finer sets would force many variants to declare two equally important primitives, which breaks the single-primitive rule. The seven primitives are not specific to the web; only the four-layer injection mechanism that implements them is. Transferring the catalog to OS or IDE agents would keep the primitives but require a new set of layers suited to that environment.

\noindent\textbf{Why the five design rules.} Each rule excludes a kind of variant whose paired drop would be hard to interpret. \emph{Determinism} makes the paired drop reproducible: the same seed produces exactly the same intervention in every repetition. \emph{Detectability} gives the agent something to act on: if the changed state were invisible in the DOM, the HTTP status, and form readback, the variant would test luck rather than capability. \emph{Recoverability} separates a harder task from an infeasible one; a variant that blocks the task completely would also block the clean run and would not produce a paired drop. \emph{Primitive purity} keeps the primary-primitive label of a variant meaningful: the admission rule $|T_{\text{task}} \cup \{p_{\text{variant}}\}| \le 2$ limits a base task and its variant to at most two primitives together. \emph{Realism} keeps the catalog close to failures that real users encounter, as opposed to synthetic adversarial constructions.

\subsection{Variant Specification Format}
\label{app:variant-spec}

A variant is a YAML file with a fixed schema: a unique \texttt{variant\_id}, the \texttt{base\_task\_id} it intervenes on, the single \texttt{target\_primitive} (\cref{app:qc}), a free-text \texttt{description} summarizing mechanism and recovery, and an ordered list of \texttt{injections} composing the intervention. Each injection records one layer choice and a \texttt{params} block whose \texttt{action} key picks one of the registered dispatches at that layer; remaining keys are layer- and dispatch-specific (\cref{app:layer-mechanics}). About $60\%$ of the $519$ variants use a single layer and the rest stack two or three; multi-layer composition lets a single variant impose, for instance, a seed-layer decoy plus a network-layer silent failure on writes targeting the decoy.

\Cref{lst:worked-variant} reproduces \texttt{amazon\_browse\_category\_\_cheapest\_decoy} verbatim: three near-identical USB-C cable decoys plant a four-way tie under sort-by-price, and the network layer ignores the first two add-to-cart requests while reporting success, so the agent must check that the cart actually changed. The primary primitive of this variant is grounding, because the main difficulty is the tie; the ignored requests are a secondary verification element.

\begin{figure}[t]
\centering
\begin{minipage}{0.96\linewidth}
\hrule\vspace{0.4em}\small
\begin{verbatim}
variant_id:        amazon_browse_category__cheapest_decoy
base_task_id:      amazon_browse_category
target_primitive:  grounding
description: >
  Three near-duplicate Electronics SKUs share the target's brand and
  thumbnail and tie at exactly $7.99, the same lowest price as the
  canonical target. Sort-by-price-low-to-high produces a four-way tie at
  the top, so the agent must read the canonical product_id (or full
  listing details) rather than just clicking the first cheapest item.
  The first two cart-add calls silently no-op so the agent must verify
  the cart actually changed.

injections:
- layer: seed                              # decoy products at session init
  params:
    action: add_confusing_decoys
    decoys:
    - {type: product, name: USB-C Charging Cable 6ft - Pro Edition,
       brand: CableTech, seller: CableTech, category: Electronics,
       price: 7.99, rating: 4.3, review_count: 820,
       match_image_of: USB-C Charging Cable 6ft,
       description: Pro-edition sibling at the same lowest price.}
    - {type: product, name: USB-C Charging Cable 6ft 2-Pack,
       brand: CableTech, seller: CableTech, category: Electronics,
       price: 7.99, rating: 4.3, review_count: 580,
       match_image_of: USB-C Charging Cable 6ft,
       description: Two-unit bundle at the same lowest price.}
    - {type: product, name: USB-C Charging Cable 6ft Refurbished,
       brand: OEM Reseller, seller: ElectroDeals Marketplace,
       category: Electronics, price: 7.99, rating: 4.3,
       review_count: 410, match_image_of: USB-C Charging Cable 6ft,
       description: Marketplace refurbished listing tied at the same
                    lowest price.}
- layer: network                           # silent no-op on first 2 writes
  params:
    action: silent_fail
    url_pattern: '**/api/env/amazon/cart/add'
    methods: [POST]
    fail_count: 2
    response_body:
      cart_item: {id: cart_fake_retry, product_id: prod_fake_retry,
                  product_name: Pending cart add}
\end{verbatim}
\vspace{-0.3em}\hrule
\end{minipage}
\caption{The \texttt{amazon\_browse\_category\_\_cheapest\_decoy} variant YAML, reproduced verbatim. Field-level comments are this paper's; the file itself is unannotated. Three decoy products and a two-call silent fail compose into one $\Delta$-pair against the matching base task.}
\label{lst:worked-variant}
\end{figure}

\subsection{Layer Mechanics}
\label{app:layer-mechanics}

The four layers act on different state or at different points of the request path; each carries an action key dispatched to a registered handler at run time, and all four share the determinism contract of \cref{app:env-infra}. \Cref{tab:layers} summarizes per-layer family counts and primary primitives.

\begin{table}[t]
\centering
\caption{Intervention families and dispatch branches by injection layer. \emph{Primary primitives targeted} lists the primitives most often targeted by variants at that layer; every layer reaches multiple primitives.}
\label{tab:layers}
\small
\begin{tabular}{lrrp{0.40\linewidth}}
\toprule
Layer  & Families & Dispatches & Primary primitives targeted \\
\midrule
Seed     & 7  & 14 & grounding, state tracking, exploration, backtracking, verification \\
Server   & 5  & 10 & planning, state tracking, grounding, exploration, verification \\
Network  & 7  & 8  & patience, verification, backtracking, state tracking \\
Client   & 10 & 16 & grounding, verification, exploration, backtracking, patience \\
\midrule
\textbf{Total} & \textbf{29} & \textbf{48} & all seven primitives \\
\bottomrule
\end{tabular}
\end{table}

The \emph{seed} layer mutates initial state once at session creation, after the base task's seed program has populated the environment, dispatching on the variant's \texttt{action} key to one of $14$ builders that append or rewrite entities (decoys, adversarial bodies, contradictory updates, hidden targets). Once seeded, its content is passive: the agent must read and disambiguate. The \emph{server} layer applies a single structural mutation after the seed-layer pass; its $10$ dispatches scramble timestamps, shuffle list orders, hide prerequisite entities, inject distractor notifications, or corrupt one field on a target entity. Unlike the seed layer, it edits already-seeded entities rather than appending new ones.

The \emph{network} layer is a Starlette middleware registered globally on the FastAPI application; it intercepts every matching request at runtime. Variants address requests by URL glob and HTTP method, and the middleware tracks per-pattern call counters so a dispatch can fire on the first $N$ calls, on every $k$-th call, on call indices in a correlated window, or with a seeded probability $p$. Eight dispatches are registered: \texttt{delay}, \texttt{error\_then\_success}, \texttt{silent\_fail}, \texttt{misleading\_success}, \texttt{stale\_data}, \texttt{concurrent\_modification}, \texttt{rate\_limit}, and \texttt{session\_expiry}. The \emph{client} layer is a React component injected into every SPA at build time; given a registered dispatch list, it applies the corresponding DOM mutation with direct access to the rendered DOM, the SPA's React state, and \texttt{aria} metadata. Sixteen dispatches span label misalignment, decoy elements, intercepting overlays, click swallowing, save drift, double-submit traps, and stuck loaders.

\subsection{Family Inventory}
\label{app:family-inventory}

The $48$ dispatch branches form $29$ intervention families under two equivalence relations: environment-specific specializations of one mechanism (\texttt{add\_decoy\_notifications} and \texttt{add\_noise\_orders} both inject decoy entities but write to different state collections), and behavior modes inside one dispatch (the \texttt{delay} dispatch exposes six modes that together form the \emph{Latency} family). \Cref{tab:family-inventory} lists every family with its layer, the primitives it most often targets, the representative dispatch, and its injection-record count in the catalog.

\begin{table}[t]
\centering
\caption{The $29$ intervention families of \benchmark{}, grouped by injection layer. \emph{Records} counts the injection records (a multi-layer variant contributes more than one) the family accounts for in the $519$-variant catalog. Representative dispatches name the most-used branch in code; minor specializations (e.g.\ \texttt{add\_noise\_orders} for \texttt{add\_confusing\_decoys}) fold into the same family.}
\label{tab:family-inventory}
\small
\setlength{\tabcolsep}{4pt}
\begin{tabular}{@{}llp{0.16\linewidth}rp{0.27\linewidth}@{}}
\toprule
Layer & Family & Primary primitives & Records & Representative dispatch \\
\midrule
\multirow{7}{*}{Seed}
 & Decoys \& aliases       & ground., state    & 294 & \texttt{add\_confusing\_decoys} \\
 & Adversarial content     & ground., verify   &   7 & \texttt{inject\_adv\_content} \\
 & Split information       & state             &   0 & \texttt{split\_information} \\
 & Contradictory update    & state, plan       &  11 & \texttt{add\_contradictory\_update} \\
 & Content inflation       & ground., explore  &  20 & \texttt{inflate\_target\_content} \\
 & Planted wrong answer    & verify, ground.   &  21 & \texttt{plant\_wrong\_answer} \\
 & Hidden target           & explore           &   6 & \texttt{hide\_in\_non\_obvious\_loc} \\
\midrule
\multirow{5}{*}{Server}
 & Timestamp scramble      & plan, state       &  36 & \texttt{scramble\_timestamps} \\
 & Ordering shuffle        & state, ground.    &  15 & \texttt{shuffle\_positions} \\
 & Distractor injection    & ground., state    &  16 & \texttt{inject\_distractor\_emails} \\
 & Prerequisite hiding     & explore, plan     &  22 & \texttt{add\_correction\_notice} \\
 & Field corruption        & verify, state     &   3 & \texttt{modify\_response} \\
\midrule
\multirow{7}{*}{Network}
 & Latency                 & patience          &  34 & \texttt{delay} (6 modes) \\
 & Transient error         & backtrack, patience & 60 & \texttt{error\_then\_success} \\
 & Fabricated success      & verify            & 210 & \texttt{silent\_fail} \\
 & Stale response          & state, verify     &  62 & \texttt{stale\_data} \\
 & Optimistic conflict     & backtrack         &   3 & \texttt{concurrent\_modification} \\
 & Rate limit              & patience          &   0 & \texttt{rate\_limit} \\
 & Session expiry          & backtrack         &   0 & \texttt{session\_expiry} \\
\midrule
\multirow{10}{*}{Client}
 & Label misbinding        & ground.           &   1 & \texttt{label\_input\_misalignment} \\
 & Decoy element           & ground.           &   1 & \texttt{adjacent\_selection} \\
 & Hidden/restricted affordance & explore      &   5 & \texttt{hide\_affordance} \\
 & Deceptive banner        & ground., verify   &   4 & \texttt{false\_banner} \\
 & Swallowed click         & verify, patience  &   2 & \texttt{click\_swallow} \\
 & Input perturbation      & ground., verify   &   3 & \texttt{input\_corruption} \\
 & Double-fire trap        & verify            &   1 & \texttt{double\_submit\_trap} \\
 & Intercepting overlay    & ground., explore  &   3 & \texttt{intercepting\_overlay} \\
 & Stuck loader            & patience          &   1 & \texttt{skeleton\_never\_resolves} \\
 & Interrupting modal      & ground., explore  &  10 & \texttt{distractor\_modal} \\
\bottomrule
\end{tabular}
\end{table}

The catalog is heavily concentrated: three families (\emph{Decoys \& aliases}, \emph{Fabricated success}, and \emph{Stale response}) account for $566$ injection records, more than $60$\% of the $963$ total. The long tail of single-digit-count families is intentional; rare families exist to cover failure classes that real users encounter (a stuck loader, a confirmation-dialog modal, a swallowed click) even though they are not the dominant agent failure mode. Three families currently have zero variants in the released catalog: \emph{Split information} (seed layer), and \emph{Rate limit} and \emph{Session expiry} (network layer). All three mechanisms are implemented by their respective injection layers and pass the harness's dispatch tests; we list them in \cref{tab:family-inventory} for completeness and as a hook for future catalog additions, because adding variants for an existing mechanism later requires much less work than adding a new mechanism to a layer.

\subsection{Selected Family Details}
\label{app:family-deep-dive}

We document the two largest families: \emph{Decoys \& aliases} (the most frequent overall) and \emph{Fabricated success} (the headline single-family failure source in \cref{sec:benchmarking}). The remaining families follow the same pattern (mechanism, parameters, and an explicit recovery path) and can be reconstructed from the dispatch names in \cref{tab:family-inventory} and the layer descriptions above.

\paragraph{Decoys \& aliases (seed, 294 records).} The family inserts $k$ near-duplicate entities into the seeded state along an axis that ties the canonical target. Specializations write to different collections: \texttt{add\_confusing\_decoys} appends Amazon products, Reddit posts, or LMS announcements; \texttt{add\_noise\_orders} writes Robinhood orders; \texttt{add\_decoy\_notifications} writes notifications; \texttt{alias\_entities} replicates an existing entity with mild lexical perturbations (``Alex Chen (Engineering)'' versus ``Alex Chen (Marketing)''). Decoy fields tie the target on at least one salient attribute (price, rating, brand, sender, or timestamp), so a single sort or filter cannot disambiguate them. The competent recovery is to read each candidate's discriminating attribute (full product detail, full email body, full order line) before committing; the canonical target is always uniquely identifiable from one attribute the decoys do not match.

\paragraph{Fabricated success (network, 210 records).} Two sibling dispatches at the network layer never forward the request to the real handler: \texttt{silent\_fail} returns a synthesized $200$ with a body shaped like a successful resource (an empty cart-item record, a placeholder message ID), and \texttt{misleading\_success} additionally injects a \texttt{toast: "Saved."} field so the SPA renders a green confirmation banner. In both cases server state never mutates. Parameters control which calls are intercepted (\texttt{url\_pattern}, \texttt{methods}, \texttt{fail\_count}) and what shape the synthesized response takes (\texttt{response\_body}, \texttt{toast\_message}); once \texttt{fail\_count} calls have been served the middleware lets requests through unmodified, so the variant is recoverable. The competent recovery is a re-read: after \texttt{POST /cart/add}, fetch \texttt{GET /cart} and observe whether the item is actually there. Across the six text-based agents, this family accounts for $504$ of the $3{,}114$ intervention runs ($16.2\%$; \cref{sec:benchmarking}) and $354$ of the $2{,}091$ failed intervention runs ($16.9\%$), a failure rate of $0.70$.

\section{Agents and Inference Setup}
\label{app:baselines}

\subsection{Models, Snapshots, and Decoding}
\label{app:models-decoding}

\Cref{tab:models-decoding} reports the API model identifier, provider, and decoding configuration for each agent in our evaluation. The six text-based agents share the same harness, action budget, observation contract, and evaluator; only the model identity changes. The three GUI-only agents share the evaluator and task/variant catalog but use a different harness and action grammar (\cref{app:obs-pixel,app:action-space}).

\begin{table}[t]
\centering
\caption{Model snapshots and decoding settings used in the reported \benchmark{} sweep. \emph{Provider} is the route the harness uses; Anthropic models can be served either through the Anthropic API directly or through AWS Bedrock with the same snapshot. All snapshots use \texttt{max\_tokens}=$4096$. Temperature is left at the provider default for Anthropic and Google; the OpenAI GPT-5.x family ignores any non-default temperature, so the harness omits the kwarg entirely.}
\label{tab:models-decoding}
\small
\setlength{\tabcolsep}{4pt}
\begin{tabular}{@{}llp{0.18\linewidth}p{0.24\linewidth}@{}}
\toprule
Model & Snapshot identifier & Provider & Decoding \\
\midrule
Claude-Opus-4.7   & \texttt{claude-opus-4-7}        & Anthropic / Bedrock & default temp.\ \\
Claude-Sonnet-4.6 & \texttt{claude-sonnet-4-6}      & Anthropic / Bedrock & default temp.\ \\
GPT-5.4           & \texttt{gpt-5.4}                & OpenAI              & temp.\ omitted, \texttt{reasoning\_effort=medium} \\
GPT-5.4-mini      & \texttt{gpt-5.4-mini}           & OpenAI              & temp.\ omitted, \texttt{reasoning\_effort=medium} \\
Gemini-3.1-Pro    & \texttt{gemini-3.1-pro}         & Google              & default temp.\ \\
Gemini-3-Flash    & \texttt{gemini-3-flash-preview} & Google              & default temp.\ \\
\bottomrule
\end{tabular}
\end{table}

Our evaluation uses a fixed 40-action cap and model-specific wall-clock caps of 600--1,200\,s, identical between the clean and intervention conditions of a base task; the step and time estimates in the task files do not limit the runs (\cref{app:env}, \emph{Episode termination}). Each run starts from a fresh backend state, browser context, and model conversation; agents carry no memory across tasks or conditions, and conversation trimming follows the $26{,}000$-token contract of \cref{app:obs-text}. The Gemini route rotates across multiple API keys on $429$ and $503$ responses; OpenAI and Anthropic use the SDK's exponential backoff; any LLM call exceeding $120$\,s is aborted and the harness advances to the next step.

\noindent\textbf{Budget audit.} The 40-action cap terminates 27 of 3,114 runs (0.87\%) in each condition, while Opus and Sonnet wall-clock timeout rates are similar across conditions. Of 36 failed Opus cap-hit runs repeated with 60 actions and 2,400\,s, 31 use more than 40 actions but 28 still fail; the eight recoveries split evenly between clean and intervention, leaving the paired gap unchanged. Additional controlled checks are reported in \cref{app:controlled-studies}.

\subsection{Additional Agent Evaluations}
\label{app:additional-models}

\noindent\textbf{Independent rollouts.} Across three rollouts on the same stratified 140-task subset, Opus-4.7 and Sonnet-4.6 have paired pass-rate drops of $27.1\pm1.4$ and $27.6\pm2.5$ points; backtracking and verification are both top-3 in five of six model--rollout cells.

\noindent\textbf{Open-weight agents.}
We additionally evaluated full 519-pair sweeps of Kimi-K2.5 and Qwen3-VL-235B under the same text observation/action contract, evaluator, and canonical-diff scoring. Both use a 40-action cap and a non-binding 2,400-s wall-clock cap so provider latency does not determine termination. \Cref{tab:open-weight} reports paired bootstrap intervals over base tasks.

\begin{table}[h]
\centering
\caption{Open-weight text-harness results on all 519 paired tasks. Differences in pass rate are in percentage points; score differences are in canonical-diff score units.}
\label{tab:open-weight}
\small
\setlength{\tabcolsep}{4pt}
\begin{tabular}{lcccc}
\toprule
Model & Clean Pass / Score & Intervention Pass / Score & $\Delta$Pass [95\% CI] & $\Delta$Score [95\% CI] \\
\midrule
Kimi-K2.5       & 32.4\% / 0.463 & 18.1\% / 0.310 & 14.3 [10.0, 18.5] & 0.153 [0.116, 0.190] \\
Qwen3-VL-235B   & 21.2\% / 0.352 &  9.1\% / 0.211 & 12.1 [8.9, 15.6]  & 0.141 [0.106, 0.176] \\
\bottomrule
\end{tabular}
\end{table}

Both models have double-digit pass-rate drops across score thresholds 0.5--1.0. Their largest drops include backtracking and verification, but their lower clean baselines make this a qualitative generalization test; Qwen is evaluated only through the text harness.

\subsection{System Prompts}
\label{app:prompts}

\Cref{lst:prompt-text} reproduces the text-based harness system prompt verbatim. It fixes the action grammar (\cref{tab:action-text}), names the accessibility-tree convention, requires exactly one Python function call per step, and explicitly asks the agent to verify the task before issuing \texttt{send\_msg\_to\_user}.

\begin{figure}[h]
\centering
\begin{minipage}{0.96\linewidth}
\hrule\vspace{0.4em}\small
\begin{verbatim}
You are a web agent. You interact with web pages to complete tasks.

## Observation
Each step you receive an accessibility tree (AXTree) of the current page.
Elements are marked with string IDs called `bid`:
- `[bid] role "name"` -- interactive or meaningful elements
- Indentation shows nesting
- Attributes: value="...", checked, unchecked, disabled, focused, selected
- `[OVERLAY]` marks dialogs blocking the page -- handle these first

## Actions
Respond with a single Python function call. Available actions:

click('a51')                      -- Click element
fill('b22', 'hello world')        -- Fill text field
select_option('c3', 'California') -- Select dropdown option
hover('d7')                       -- Hover over element
press('48', 'Enter')              -- Press key (e.g. 'Enter', 'Backspace')
scroll(0, 300)                    -- Scroll (pixels, positive=down/right)
dblclick('12')                    -- Double-click element
drag_and_drop('a1', 'b2')         -- Drag and drop
clear('b22')                      -- Clear text field
focus('b22')                      -- Focus element
send_msg_to_user('done')          -- Report your answer/completion
report_infeasible('reason')       -- Report task is impossible
noop(1000)                        -- Wait (default 1000ms)

## Rules
1. Output EXACTLY ONE function call per step. No extra text, no markdown.
2. Use bid values from the CURRENT observation only. Never reuse old bids.
3. Any action argument that refers to a bid must be a quoted string,
   e.g. click('75'), not click(75).
4. Handle overlays/dialogs before interacting with background elements.
5. Before calling send_msg_to_user, verify the task is actually complete.
6. If the task asks for a specific answer, pass it to send_msg_to_user.
\end{verbatim}
\vspace{-0.3em}\hrule
\end{minipage}
\caption{System prompt for the text-based harness, reproduced verbatim.}
\label{lst:prompt-text}
\end{figure}

The GUI-only harness prompt differs in three blocks (\cref{lst:prompt-pixel}): a \texttt{<think>...</think>} reasoning block before the action, a per-model coordinate system filled in from the viewport in \cref{tab:viewport}, and the coordinate action grammar (\cref{tab:action-pixel}) in place of the BID actions. The pixel-coordinate version is sent to Anthropic and OpenAI models; Gemini and Qwen receive the same prompt with the coordinate block swapped for the $0$--$1000$ normalized-grid variant.

\begin{figure}[h]
\centering
\begin{minipage}{0.96\linewidth}
\hrule\vspace{0.4em}\small
\begin{verbatim}
# Response Format (BOTH parts are required, in this order)

1. A `<think>...</think>` block with your reasoning. NEVER skip this.
2. Exactly ONE valid action call on a new line after the `</think>`.

Inside `<think>` answer: (1) what you observe, (2) what your previous
action did, (3) whether the task goal is satisfied, (4) otherwise what
to do next and why.

# Coordinate System (PIXELS, viewport is {w} x {h})

  (0, 0)       = top-left corner
  ({cx}, {cy}) = center of the viewport
  ({w}, {h})   = bottom-right corner

Output coordinates in actual pixel values matching the screenshot
dimensions.
\end{verbatim}
\vspace{-0.3em}\hrule
\end{minipage}
\caption{GUI-only harness system prompt: the blocks that differ from the text-based harness prompt (\cref{lst:prompt-text}). The task-completion preamble matches \cref{lst:prompt-text} verbatim; the action grammar is replaced by \cref{tab:action-pixel}. Width $w$ and height $h$ are filled in from the per-model viewport (\cref{tab:viewport}), with $c_x = w/2$, $c_y = h/2$. Gemini and Qwen receive the same prompt with the coordinate-system block replaced by the $0$--$1000$ normalized-grid variant.}
\label{lst:prompt-pixel}
\end{figure}

\subsection{Observation Post-Processing}
\label{app:obs-filter}

The text observation is not the raw accessibility tree of the rendered page. The harness flattens the tree under four filters that drop nodes the agent cannot meaningfully act on, and prepends a header with the goal, the last action, its error string (if any), and the current URL. \Cref{tab:axtree-filter} states the filter rules, in the spirit of OSWorld's a11y-tree filter table~\citep{DBLP:conf/nips/XieZCLZCHCSLLXZ24}.

\begin{table}[h]
\centering
\caption{Accessibility-tree filtering rules applied to the flattened tree the agent observes each step.}
\label{tab:axtree-filter}
\small
\begin{tabular}{lp{0.65\linewidth}}
\toprule
Filter & Effect \\
\midrule
\texttt{with\_clickable=True}        & Annotates clickable nodes with the \texttt{clickable} role marker so the agent can prefer them over decorative elements. \\
\texttt{with\_visible=True}          & Annotates each node with its \texttt{visible} flag from the BrowserGym extra-properties dict. \\
\texttt{filter\_visible\_only=True}  & Drops nodes whose computed visibility is \texttt{false} (off-screen, \texttt{display:none}, or covered). \\
\texttt{filter\_with\_bid\_only=True}& Drops nodes that lack an addressable \texttt{bid} so the agent cannot emit an action against an unaddressable target. \\
\midrule
\multicolumn{2}{l}{Post-formatting cleanup:} \\
\multicolumn{2}{l}{\quad ASCII-control characters (\texttt{\textbackslash x00}--\texttt{\textbackslash x1f} excluding tab and newline) are stripped before serialization.} \\
\multicolumn{2}{l}{\quad Goal, last action, last action error, and URL are prepended as four labeled lines.} \\
\bottomrule
\end{tabular}
\end{table}

The GUI-only harness applies no post-processing: the agent receives the rendered viewport screenshot encoded as a base64 PNG data URL (\cref{app:obs-pixel}).

\subsection{Compute and Cost}
\label{app:compute-cost}

Our evaluation covers six text-based agents and three GUI-only agents at $1{,}038$ task-conditions each ($519$ clean and $519$ intervention). \Cref{tab:runtime-cost} reports per-model wall-clock derived from the released trajectory bundle.

\begin{table}[h]
\centering
\caption{Per-model runtime summary on the full $1{,}038$ task-condition sweep. Median / mean / p90 are over per-trajectory wall-clock seconds. Total wall-clock is the sum of episode wall-clock across the $1{,}038$ runs; it does not include harness overhead or the time spent waiting on per-call provider rate-limits. Token-level billed counts are stored next to each trajectory but are not aggregated into the table because billed-token totals depend on provider-side accounting that we do not redistribute; we therefore do not report a USD figure here.}
\label{tab:runtime-cost}
\small
\setlength{\tabcolsep}{4pt}
\begin{tabular}{lrrrrrr}
\toprule
Model & Runs & Pass & Median\,s & Mean\,s & p90\,s & Total\,hr \\
\midrule
Gemini-3.1-Pro    & 1038 & 58.3\% & 189 & 240 &  515 &  69.3 \\
Gemini-3-Flash    & 1038 & 52.9\% &  98 & 128 &  260 &  36.8 \\
GPT-5.4           & 1038 & 48.2\% & 129 & 165 &  364 &  47.7 \\
GPT-5.4-mini      & 1038 & 24.2\% &  61 &  78 &  154 &  22.6 \\
Opus-4.7          & 1038 & 43.1\% & 498 & 534 &  999 & 154.0 \\
Sonnet-4.6        & 1038 & 39.2\% & 692 & 783 & 1305 & 225.6 \\
\midrule
v-Gemini-3.1-Pro & 1038 & 18.8\% & 438 & 378 &  590 & 107.0 \\
v-GPT-5.4        & 1038 &  6.3\% & 177 & 167 &  283 &  47.5 \\
v-Opus-4.7       & 1038 & 22.9\% & 214 & 183 &  278 &  51.9 \\
\bottomrule
\end{tabular}
\end{table}

The harness is single-process; each run launches a Playwright-driven headless Chromium container on a CPU-only worker (16 vCPU, 32\,GB RAM), and no GPU is used because every model is consumed through a remote provider API. The sweep was scheduled as independent SLURM jobs with up to $32$ concurrent workers; total compute was dominated by per-call API latency, so the ``Total\,hr'' column of \cref{tab:runtime-cost} is an upper bound on cluster time after start-up. Per-trajectory billed-token totals are stored alongside the raw provider responses but are not aggregated into a USD figure here, since that would require redistributing provider-side billing data; cost can be reconstructed from the released response logs and the corresponding provider list rates. Runs that exhaust the action or wall-clock budget are not excluded: the canonical-diff evaluator scores their final backend state by the rule of \cref{app:eval-canonical}.

\subsection{Released Artifacts}
\label{app:repro}

The public release contains the harness, the seven environment SPAs, the task and intervention catalogs, the canonical-diff evaluator, the trajectory feature extractor, and the scripts that emit every table and figure cited in this paper. The aggregated trajectory bundle holds one row per task-condition (model, task, env, difficulty, condition, score, steps, elapsed time, source variant) with a parallel feature table containing the terminal verb, repeat-action signature, error counts, positive/negative check counts, final-thought keyword scan, and the rule-based failure mode of \cref{app:modes}; the human-trace bundle is parallel, with per-attempt metadata and raw and cleaned event lists. \Cref{tab:asset-card} summarizes the released artifacts, and the code and data are available at \href{https://www.breakingweb.app}{\nolinkurl{www.breakingweb.app}}. Annotator identifiers are coded throughout the bundle, server logs do not retain client IP addresses, and viewport screenshots and free-text rubric comments are withheld pending a personal-information audit.

\begin{table}[h]
\centering
\caption{Asset card for the public release: code, task and variant YAMLs, and aggregated trajectory bundles.}
\label{tab:asset-card}
\small
\setlength{\tabcolsep}{4pt}
\begin{tabular}{p{0.22\linewidth}p{0.70\linewidth}}
\toprule
Asset & Contents \\
\midrule
Code                & Harness, seven environment SPAs, evaluator, intervention dispatcher, and the scripts that regenerate every figure and table from the trajectory bundle. \\
Task and variant YAMLs & $519$ base task YAMLs paired one-to-one with $519$ intervention variants drawn from $29$ intervention families. \\
Trajectory bundles  & Per-trajectory raw artifacts and the aggregated agent and human tables on which the paper's numbers are computed. \\
\bottomrule
\end{tabular}
\end{table}

\section{Results by Primitive}
\label{app:per-primitive}

\subsection{Full Per-Model and Per-Primitive Breakdown}
\label{app:headline-full}

\Cref{tab:headline-full} is the full counterpart of the compact \cref{tab:headline} in the main paper: each cell reports the clean baseline and the matched intervention-condition pass rate (clean$\to$iv), followed by the paired delta in pass rate ($\Delta$p, in points) and the paired delta in mean canonical-diff score ($\Delta$s, in score units) over the same base tasks. We report $\Delta$s separately because the score is continuous: it can decrease even when the pass/fail outcome does not change. The first numeric column aggregates over all $519$ paired base tasks; the seven primitive columns restrict to base tasks whose intervention variant targets that primitive.

\begin{table}[h]
\centering
\caption{Per-(model, primitive) clean-and-intervention breakdown, full version of \cref{tab:headline}. Each cell carries: the clean pass rate, the matched intervention pass rate (separated by $\to$), the paired $\Delta$ in pass rate ($\Delta$p, in points; $\downarrow$ in the main table corresponds to a positive value here), and the paired $\Delta$ in mean canonical-diff score ($\Delta$s, score units). All numbers are computed only over base tasks with both conditions available.}
\label{tab:headline-full}
\scriptsize
\setlength{\tabcolsep}{2.5pt}
\renewcommand{\arraystretch}{0.95}
\resizebox{\linewidth}{!}{%
\begin{tabular}{lcccccccc}
\toprule
 & & \multicolumn{7}{c}{Per primitive} \\
\cmidrule(lr){3-9}
Model & Total & Grnd & Plan & State & Back & Patnce & Expl & Verif \\
\midrule
Gemini-3.1-Pro     & \makecell[r]{72.1$\to$44.5\\{\scriptsize $\Delta$p\,+27.6\,/\,$\Delta$s\,+0.18}} & \makecell[r]{76.8$\to$42.3\\{\scriptsize $\Delta$p\,+34.5\,/\,$\Delta$s\,+0.20}} & \makecell[r]{69.2$\to$59.6\\{\scriptsize $\Delta$p\,+9.6\,/\,$\Delta$s\,+0.08}} & \makecell[r]{61.9$\to$48.8\\{\scriptsize $\Delta$p\,+13.1\,/\,$\Delta$s\,+0.04}} & \makecell[r]{78.0$\to$40.2\\{\scriptsize $\Delta$p\,+37.8\,/\,$\Delta$s\,+0.32}} & \makecell[r]{72.0$\to$48.0\\{\scriptsize $\Delta$p\,+24.0\,/\,$\Delta$s\,+0.17}} & \makecell[r]{48.6$\to$37.8\\{\scriptsize $\Delta$p\,+10.8\,/\,$\Delta$s\,$-$0.02}} & \makecell[r]{80.3$\to$40.8\\{\scriptsize $\Delta$p\,+39.4\,/\,$\Delta$s\,+0.29}} \\
Gemini-3-Flash     & \makecell[r]{63.2$\to$42.6\\{\scriptsize $\Delta$p\,+20.6\,/\,$\Delta$s\,+0.14}} & \makecell[r]{56.5$\to$34.5\\{\scriptsize $\Delta$p\,+22.0\,/\,$\Delta$s\,+0.15}} & \makecell[r]{65.4$\to$61.5\\{\scriptsize $\Delta$p\,+3.8\,/\,$\Delta$s\,+0.03}} & \makecell[r]{60.7$\to$47.6\\{\scriptsize $\Delta$p\,+13.1\,/\,$\Delta$s\,+0.09}} & \makecell[r]{78.0$\to$42.7\\{\scriptsize $\Delta$p\,+35.4\,/\,$\Delta$s\,+0.28}} & \makecell[r]{52.0$\to$48.0\\{\scriptsize $\Delta$p\,+4.0\,/\,$\Delta$s\,+0.01}} & \makecell[r]{51.4$\to$43.2\\{\scriptsize $\Delta$p\,+8.1\,/\,$\Delta$s\,$-$0.01}} & \makecell[r]{73.2$\to$39.4\\{\scriptsize $\Delta$p\,+33.8\,/\,$\Delta$s\,+0.24}} \\
GPT-5.4            & \makecell[r]{61.3$\to$35.1\\{\scriptsize $\Delta$p\,+26.2\,/\,$\Delta$s\,+0.20}} & \makecell[r]{55.4$\to$26.8\\{\scriptsize $\Delta$p\,+28.6\,/\,$\Delta$s\,+0.22}} & \makecell[r]{73.1$\to$44.2\\{\scriptsize $\Delta$p\,+28.8\,/\,$\Delta$s\,+0.23}} & \makecell[r]{51.2$\to$36.9\\{\scriptsize $\Delta$p\,+14.3\,/\,$\Delta$s\,+0.09}} & \makecell[r]{76.8$\to$41.5\\{\scriptsize $\Delta$p\,+35.4\,/\,$\Delta$s\,+0.29}} & \makecell[r]{56.0$\to$52.0\\{\scriptsize $\Delta$p\,+4.0\,/\,$\Delta$s\,+0.05}} & \makecell[r]{43.2$\to$29.7\\{\scriptsize $\Delta$p\,+13.5\,/\,$\Delta$s\,+0.08}} & \makecell[r]{71.8$\to$35.2\\{\scriptsize $\Delta$p\,+36.6\,/\,$\Delta$s\,+0.30}} \\
GPT-5.4-mini       & \makecell[r]{33.1$\to$15.2\\{\scriptsize $\Delta$p\,+17.9\,/\,$\Delta$s\,+0.18}} & \makecell[r]{29.8$\to$15.5\\{\scriptsize $\Delta$p\,+14.3\,/\,$\Delta$s\,+0.15}} & \makecell[r]{26.9$\to$15.4\\{\scriptsize $\Delta$p\,+11.5\,/\,$\Delta$s\,+0.14}} & \makecell[r]{26.2$\to$17.9\\{\scriptsize $\Delta$p\,+8.3\,/\,$\Delta$s\,+0.11}} & \makecell[r]{45.1$\to$8.5\\{\scriptsize $\Delta$p\,+36.6\,/\,$\Delta$s\,+0.36}} & \makecell[r]{28.0$\to$8.0\\{\scriptsize $\Delta$p\,+20.0\,/\,$\Delta$s\,+0.19}} & \makecell[r]{21.6$\to$18.9\\{\scriptsize $\Delta$p\,+2.7\,/\,$\Delta$s\,+0.03}} & \makecell[r]{47.9$\to$19.7\\{\scriptsize $\Delta$p\,+28.2\,/\,$\Delta$s\,+0.22}} \\
Opus-4.7           & \makecell[r]{54.5$\to$31.6\\{\scriptsize $\Delta$p\,+22.9\,/\,$\Delta$s\,+0.19}} & \makecell[r]{53.6$\to$29.2\\{\scriptsize $\Delta$p\,+24.4\,/\,$\Delta$s\,+0.20}} & \makecell[r]{44.2$\to$23.1\\{\scriptsize $\Delta$p\,+21.2\,/\,$\Delta$s\,+0.27}} & \makecell[r]{50.0$\to$31.0\\{\scriptsize $\Delta$p\,+19.0\,/\,$\Delta$s\,+0.15}} & \makecell[r]{67.1$\to$37.8\\{\scriptsize $\Delta$p\,+29.3\,/\,$\Delta$s\,+0.24}} & \makecell[r]{40.0$\to$32.0\\{\scriptsize $\Delta$p\,+8.0\,/\,$\Delta$s\,$-$0.01}} & \makecell[r]{32.4$\to$27.0\\{\scriptsize $\Delta$p\,+5.4\,/\,$\Delta$s\,+0.01}} & \makecell[r]{71.8$\to$39.4\\{\scriptsize $\Delta$p\,+32.4\,/\,$\Delta$s\,+0.25}} \\
Sonnet-4.6         & \makecell[r]{50.3$\to$28.1\\{\scriptsize $\Delta$p\,+22.2\,/\,$\Delta$s\,+0.20}} & \makecell[r]{54.2$\to$24.4\\{\scriptsize $\Delta$p\,+29.8\,/\,$\Delta$s\,+0.23}} & \makecell[r]{40.4$\to$21.2\\{\scriptsize $\Delta$p\,+19.2\,/\,$\Delta$s\,+0.20}} & \makecell[r]{44.0$\to$26.2\\{\scriptsize $\Delta$p\,+17.9\,/\,$\Delta$s\,+0.16}} & \makecell[r]{59.8$\to$34.1\\{\scriptsize $\Delta$p\,+25.6\,/\,$\Delta$s\,+0.22}} & \makecell[r]{28.0$\to$28.0\\{\scriptsize $\Delta$p\,+0.0\,/\,$\Delta$s\,+0.09}} & \makecell[r]{29.7$\to$27.0\\{\scriptsize $\Delta$p\,+2.7\,/\,$\Delta$s\,+0.02}} & \makecell[r]{63.4$\to$38.0\\{\scriptsize $\Delta$p\,+25.4\,/\,$\Delta$s\,+0.27}} \\
\midrule
v-Gemini-3.1-Pro  & \makecell[r]{28.1$\to$13.1\\{\scriptsize $\Delta$p\,+15.0\,/\,$\Delta$s\,+0.11}} & \makecell[r]{26.2$\to$11.9\\{\scriptsize $\Delta$p\,+14.3\,/\,$\Delta$s\,+0.09}} & \makecell[r]{19.2$\to$11.5\\{\scriptsize $\Delta$p\,+7.7\,/\,$\Delta$s\,+0.09}} & \makecell[r]{31.0$\to$15.5\\{\scriptsize $\Delta$p\,+15.5\,/\,$\Delta$s\,+0.08}} & \makecell[r]{30.5$\to$12.2\\{\scriptsize $\Delta$p\,+18.3\,/\,$\Delta$s\,+0.15}} & \makecell[r]{8.0$\to$8.0\\{\scriptsize $\Delta$p\,+0.0\,/\,$\Delta$s\,$-$0.01}} & \makecell[r]{21.6$\to$16.2\\{\scriptsize $\Delta$p\,+5.4\,/\,$\Delta$s\,+0.01}} & \makecell[r]{43.7$\to$15.5\\{\scriptsize $\Delta$p\,+28.2\,/\,$\Delta$s\,+0.26}} \\
v-GPT-5.4         & \makecell[r]{10.2$\to$5.6\\{\scriptsize $\Delta$p\,+4.6\,/\,$\Delta$s\,+0.07}} & \makecell[r]{9.5$\to$7.1\\{\scriptsize $\Delta$p\,+2.4\,/\,$\Delta$s\,+0.04}} & \makecell[r]{7.7$\to$3.8\\{\scriptsize $\Delta$p\,+3.8\,/\,$\Delta$s\,+0.04}} & \makecell[r]{10.7$\to$4.8\\{\scriptsize $\Delta$p\,+6.0\,/\,$\Delta$s\,+0.10}} & \makecell[r]{11.0$\to$4.9\\{\scriptsize $\Delta$p\,+6.1\,/\,$\Delta$s\,+0.11}} & \makecell[r]{0.0$\to$4.0\\{\scriptsize $\Delta$p\,$-$4.0\,/\,$\Delta$s\,+0.09}} & \makecell[r]{2.7$\to$5.4\\{\scriptsize $\Delta$p\,$-$2.7\,/\,$\Delta$s\,$-$0.00}} & \makecell[r]{19.7$\to$5.6\\{\scriptsize $\Delta$p\,+14.1\,/\,$\Delta$s\,+0.13}} \\
v-Opus-4.7        & \makecell[r]{33.7$\to$15.8\\{\scriptsize $\Delta$p\,+17.9\,/\,$\Delta$s\,+0.15}} & \makecell[r]{33.3$\to$16.1\\{\scriptsize $\Delta$p\,+17.3\,/\,$\Delta$s\,+0.13}} & \makecell[r]{25.0$\to$17.3\\{\scriptsize $\Delta$p\,+7.7\,/\,$\Delta$s\,+0.08}} & \makecell[r]{28.6$\to$16.7\\{\scriptsize $\Delta$p\,+11.9\,/\,$\Delta$s\,+0.11}} & \makecell[r]{40.2$\to$14.6\\{\scriptsize $\Delta$p\,+25.6\,/\,$\Delta$s\,+0.23}} & \makecell[r]{20.0$\to$16.0\\{\scriptsize $\Delta$p\,+4.0\,/\,$\Delta$s\,$-$0.03}} & \makecell[r]{24.3$\to$13.5\\{\scriptsize $\Delta$p\,+10.8\,/\,$\Delta$s\,+0.08}} & \makecell[r]{49.3$\to$15.5\\{\scriptsize $\Delta$p\,+33.8\,/\,$\Delta$s\,+0.31}} \\
\bottomrule
\end{tabular}%
}
\end{table}

\subsection{Failure Modes by Primitive}
\label{app:mode-by-primitive}

\Cref{tab:mode-by-primitive} reports failure-mode counts for each target primitive over the $1{,}841$ failed intervention runs from the six text-based agents. The two belief-failure modes (\texttt{misleading\_success\_taken}, \texttt{premature\_done}) dominate on grounding and backtracking variants: \texttt{misleading\_success\_taken} alone accounts for $443$ of $636$ grounding failures and $216$ of $289$ backtracking failures. The aggregated text-based vs GUI-only view of the same data appears in \cref{fig:failure-landscape}(c).

\begin{table}[h]
\centering
\caption{Failure-mode counts by target primitive ($1{,}841$ failed intervention runs across the six text-based agents; the additional $1{,}196$ GUI-only failures are absorbed into \cref{fig:failure-landscape}(c)).}
\label{tab:mode-by-primitive}
\small
\begin{tabular}{lrrrrrrr}
\toprule
Mode & Ground. & Plan. & State & Backtrk. & Patience & Explore & Verify \\
\midrule
\texttt{misleading\_success\_taken} & 443 & 91 & 132 & 216 & 37 & 66 & 149 \\
\texttt{premature\_done}            &  72 & 28 &  42 &  37 & 10 & 21 &  36 \\
\texttt{retry\_loop}                &  36 & 14 &  20 &   9 & 21 & 13 &  10 \\
\texttt{plan\_collapse}             &   0 &  0 &   3 &   2 &  2 &  0 &   1 \\
\texttt{silent\_overreach}          &  85 & 28 &  87 &  25 & 14 & 34 &  57 \\
\bottomrule
\end{tabular}
\end{table}

The mode profile differs by primitive in two reproducible ways. First, \texttt{retry\_loop} accounts for a larger share of failures on patience variants ($21$ of $84$, or $25\%$) than on any other primitive. This is consistent with the latency interventions (\cref{app:family-deep-dive}, \emph{Latency}): an agent that takes a slow response for a failure tends to retry. Second, \texttt{silent\_overreach} accounts for the largest shares of failures on state-tracking ($87$ of $284$, or $31\%$), exploration ($25\%$), and verification ($23\%$) variants, which is consistent with collateral damage to non-target state being hardest to detect from the rendered DOM on these variants.

\subsection{Step Economy on Backtracking Variants}
\label{app:step-economy-primitive}

A per-primitive view of the step-economy data (\cref{sec:analysis}) shows that, on backtracking variants, failed runs are longer than passed runs for five of the six text-based agents (by $2.4$ steps on average). Agents therefore usually attempt to recover but do not succeed.

\begin{table}[h]
\centering
\caption{Mean steps on backtracking-targeted intervention runs, partitioned by outcome. The fail-minus-pass column is positive for every model except GPT-5.4 ($-0.1$).}
\label{tab:step-economy-primitive}
\small
\begin{tabular}{lrrr}
\toprule
Model & Mean steps (fail) & Mean steps (pass) & Fail $-$ Pass \\
\midrule
Gemini-3.1-Pro     & 15.2 & 13.4 & $+1.8$ \\
Gemini-3-Flash     & 16.9 & 11.7 & $+5.2$ \\
GPT-5.4            & 14.6 & 14.6 & $-0.1$ \\
GPT-5.4-mini       & 10.1 &  9.4 & $+0.7$ \\
Opus-4.7           & 13.0 & 10.3 & $+2.7$ \\
Sonnet-4.6         & 13.8 &  9.7 & $+4.1$ \\
\bottomrule
\end{tabular}
\end{table}

\subsection{Per-Task Human Tax Versus Agent Drop}
\label{app:human-vs-agent-tax-pertask}

\begin{figure}[h]
\centering
\includegraphics[width=0.65\linewidth]{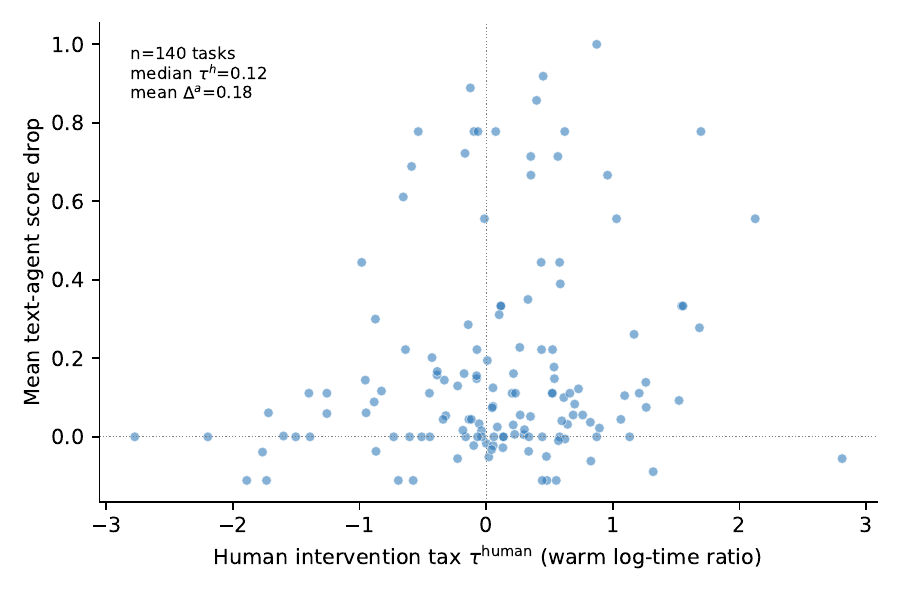}
\caption{Per-task human intervention tax $\tau^{\mathrm{human}}_t$ (warm log-time ratio) versus mean agent drop $\Delta^{\mathrm{agent}}_t$ over the $140$ paired tasks of Human-140, averaged over the six text-based models. The aggregate per-primitive view appears in \cref{fig:human-row}(b).}
\label{fig:human-vs-agent-tax-pertask-app}
\end{figure}

\Cref{fig:human-vs-agent-tax-pertask-app} disaggregates \cref{fig:human-row}(b) from $7$ primitive means to all $140$ paired Human-140 tasks. The horizontal axis is the warm log-time ratio $\tau^{\mathrm{human}}_t$ (the log ratio of the median warm wall-clock time under intervention to that in the clean condition); the vertical axis is the matched mean agent score drop $\Delta^{\mathrm{agent}}_t$ averaged across the six text-based models. The vertical band of points at $\tau^{\mathrm{human}} \approx 0$ (the warm intervention attempt is no slower than the warm clean attempt) contains tasks on which the intervention costs warm humans almost no extra time but causes the agents to fail completely ($\Delta = 1.0$).

\subsection{Robustness to Coarser Primitive Groupings}
\label{app:primitive-merge}

To test whether seven primitives are too fine-grained, we repeat the primitive analysis after merging the seven primitives into five broader groups and report the result here as a robustness check; the main paper continues to report seven primitives.

\begin{figure}[h]
\centering
\includegraphics[width=\linewidth]{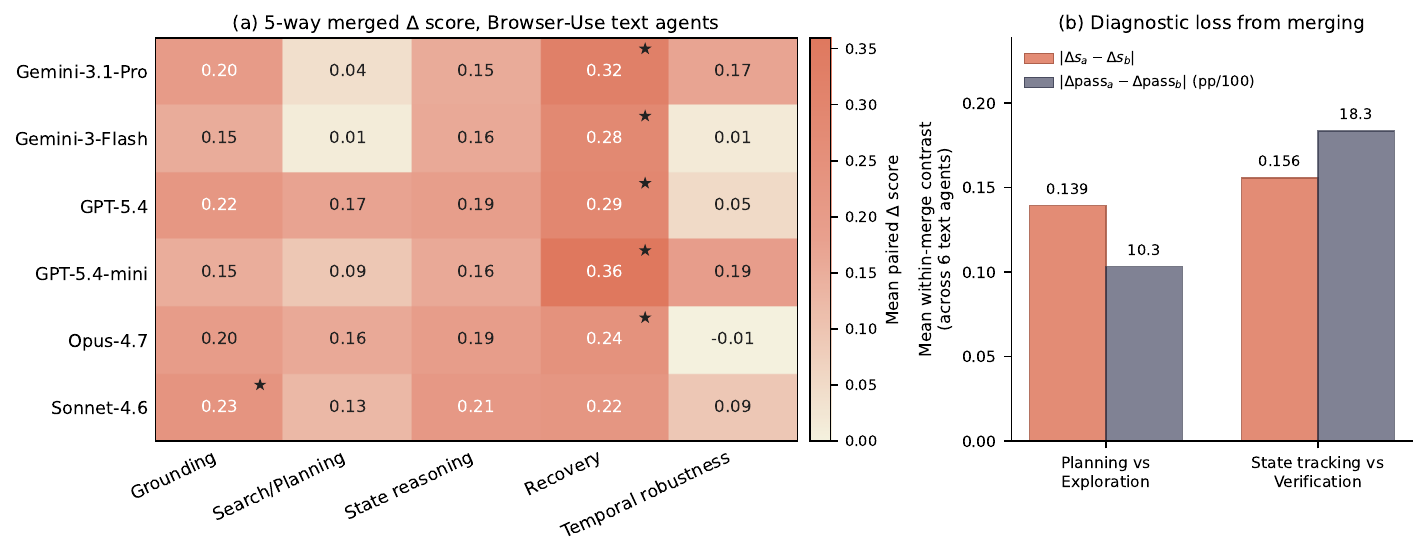}
\caption{Coarsening the seven \benchmark{} primitives to five groups preserves the broad vulnerability pattern, but merging planning with exploration and state tracking with verification hides nontrivial within-group contrasts. \textbf{(a)} Five-way merged paired $\Delta$ score heatmap on the six text-based agents; row-maximum cells are starred. \textbf{(b)} Within-merge diagnostic loss for the two real merges, averaged across the six text agents.}
\label{fig:primitive-merge}
\end{figure}

\Cref{tab:primitive-merge-map} states the mapping. \Cref{fig:primitive-merge}(a) shows the five-way merged paired score-drop heatmap on the six text-based agents, and \cref{fig:primitive-merge}(b) the within-merge diagnostic loss for the two non-trivial merges.

\begin{table}[h]
\centering
\caption{$7 \rightarrow 5$ primitive merge mapping used in the robustness check.}
\label{tab:primitive-merge-map}
\small
\begin{tabular}{lll}
\toprule
Merged group & Original primitives & Rationale \\
\midrule
Grounding             & grounding                            & observation / disambiguation primitive, kept distinct \\
Search/Planning       & planning, exploration                & both involve finding or sequencing paths \\
State reasoning       & state\_tracking, verification        & both involve reasoning about state \\
Recovery              & backtracking                         & recovery after a wrong or blocked path is its own behavior \\
Temporal robustness   & patience                             & waiting / retrying under latency is distinct from recovery \\
\bottomrule
\end{tabular}
\end{table}

\noindent\textbf{The coarser grouping preserves the main pattern.} Mean paired $\Delta$ score on the six text agents, ranked by merged group: Recovery~0.286, Grounding~0.191, State reasoning~0.178, Search/Planning~0.100, Temporal robustness~0.085. Recovery (= backtracking) is the largest merged group by mean $\Delta$ score, and state reasoning (= verification + state tracking) is third, behind grounding. The seven-way phrasing ``verification and backtracking are the most consistent weak primitives'' therefore survives the merge: backtracking maps directly onto the largest merged group, and verification's contribution remains visible after averaging into state reasoning. The main result therefore does not depend on the choice of seven primitives.

\noindent\textbf{Why the main paper still reports seven.} The merge loses diagnostic resolution. State tracking and verification differ by 0.156 score-drop units and 18.3 points on average across the six text agents; planning and exploration differ by 0.139 score-drop units and 10.3 points. These contrasts correspond to different repair strategies: verifying committed backend state is not the same as maintaining latent state, and searching for a hidden affordance is not the same as executing a multi-step plan. We therefore report the seven-way taxonomy in the main paper because it preserves diagnostic resolution, and use the five-way view as a robustness check.

\section{Results by Environment}
\label{app:env-cases}

\subsection{Per-Environment Intervention Pass Rate}
\label{app:env-model}

\begin{figure}[h]
\centering
\includegraphics[width=0.7\linewidth]{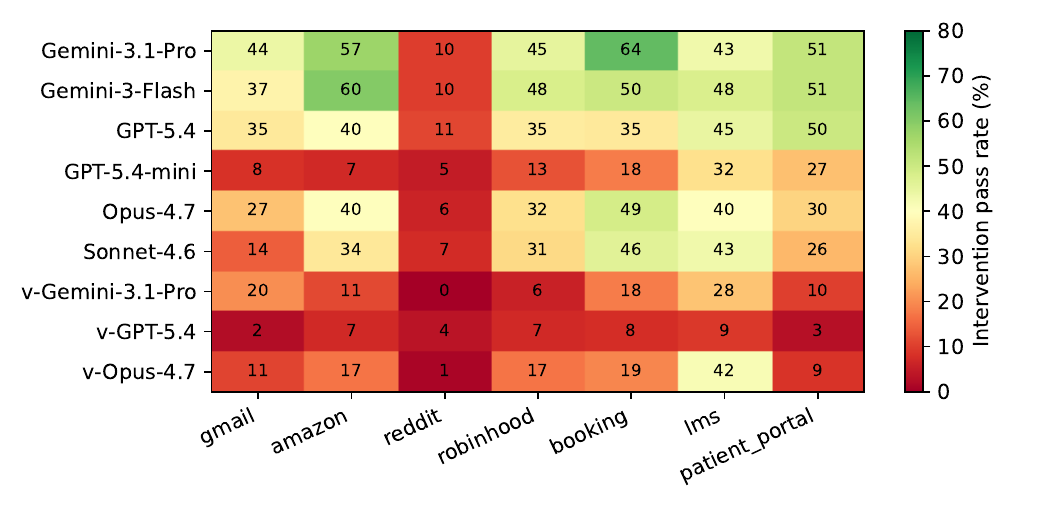}
\caption{Intervention pass rate (\%) by environment and model. Cells report the fraction of intervention runs that fully passed. ``--'' marks (model, environment) pairs with no runs in the intervention sweep.}
\label{fig:env-model}
\end{figure}

\Cref{fig:env-model} shows intervention pass rate by (environment, model). Reddit is the hardest environment for every agent (intervention pass rates of $2$--$11\%$), probably because most long-form decoy and adversarial-content interventions are in this environment; LMS, Booking, and Patient Portal are easier, with intervention pass rates of $30$--$64$\% for the strong models. The cross-environment ranking of models is preserved, but the overall level depends on the environment.

\subsection{Per-Environment Drop by Primitive}
\label{app:env-primitive}

\Cref{tab:env-primitive} reports the mean paired score drop $\Delta$ for each (environment, target primitive) cell, averaged over the six text-based agents. Empty cells indicate environments without any variant targeting that primitive in the released catalog (\cref{tab:coverage-prim-env}).

\begin{table}[h]
\centering
\caption{Mean paired score drop $\Delta = \mathrm{score}_{\mathrm{clean}} - \mathrm{score}_{\mathrm{intervention}}$ by (environment, target primitive), averaged across the six text-based agents. Cells with no targeting variants in the catalog are left blank; cells with fewer than five base tasks per model in the sweep are italicized.}
\label{tab:env-primitive}
\small
\begin{tabular}{lrrrrrrr}
\toprule
Environment & Ground. & Plan. & State & Backtrk. & Patience & Explore & Verify \\
\midrule
Amazon          & 0.23 & 0.39 & 0.17 & 0.45 & \emph{-0.01} & \emph{0.01} & 0.47 \\
Booking         & 0.07 & \emph{0.02} & 0.09 & 0.12 & -0.06 & \emph{0.10} & 0.23 \\
Gmail           & 0.08 & \emph{-0.04} & 0.03 & -0.00 & 0.09 & -0.01 & 0.27 \\
LMS             & 0.08 & 0.12 & 0.07 & \emph{0.21} & \emph{0.00} & -0.00 & 0.12 \\
Patient Portal  & 0.18 & 0.20 & 0.12 & 0.34 &       & 0.02 & 0.26 \\
Reddit          & 0.35 &       & 0.25 & 0.45 &       &       & 0.42 \\
Robinhood       & 0.05 & 0.05 & 0.06 & 0.45 & 0.25 & 0.05 & 0.20 \\
\bottomrule
\end{tabular}
\end{table}

The table shows two patterns. (i) Verification and backtracking deficits are not Reddit-specific: they show up in every environment that ships variants targeting them, so these weaknesses are not specific to one environment. (ii) Reddit's hardness in \cref{fig:env-model} is concentrated on grounding (long-form decoys and adversarial content); environments with broader primitive coverage (Gmail, LMS, Patient Portal) spread the deficit across primitives.

\begin{figure}[h]
\centering
\includegraphics[width=\linewidth]{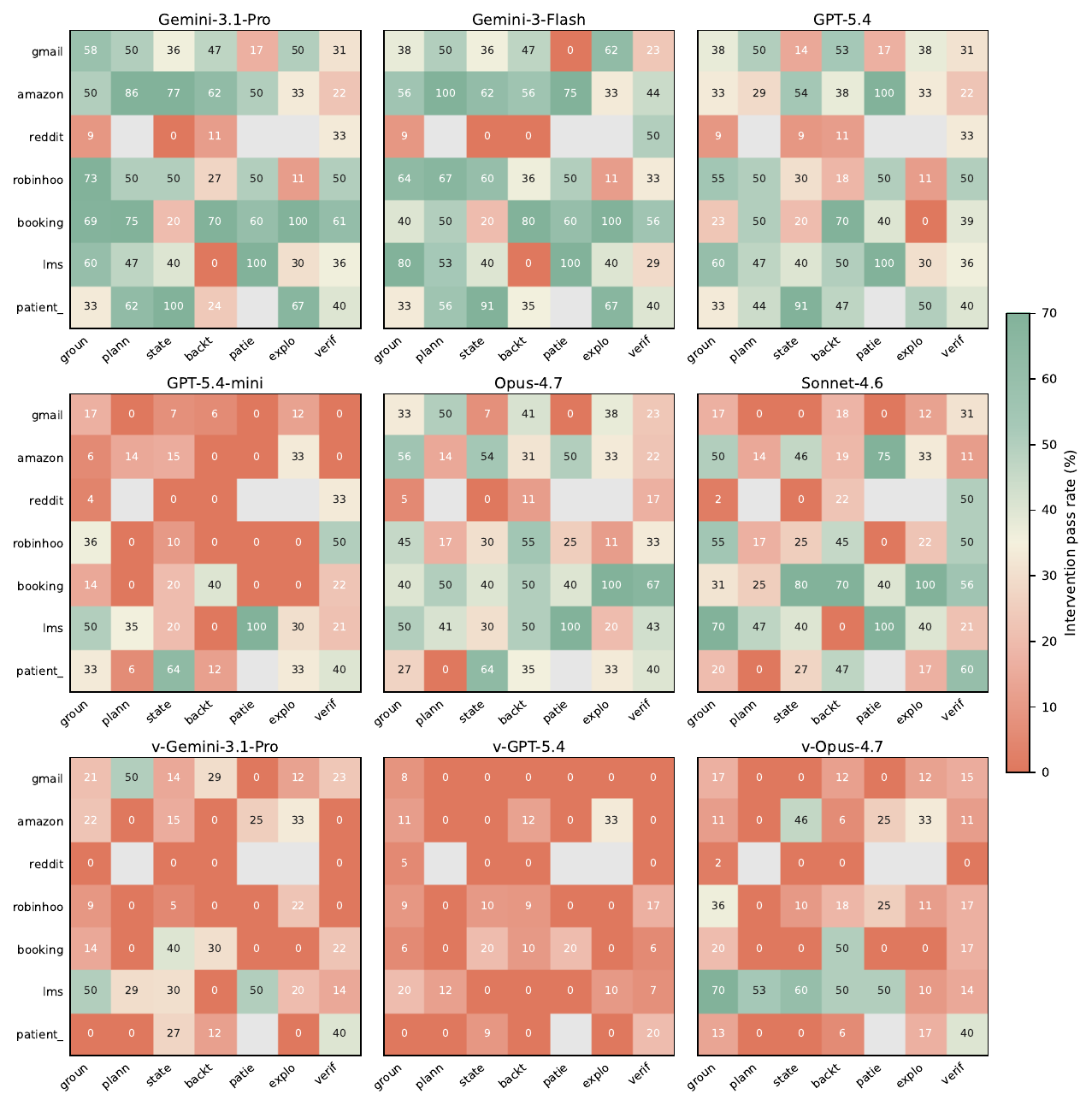}
\caption{Per-(environment, primitive) intervention pass rate, broken down by agent. Each subplot is one model; rows are the seven environments and columns the seven primitives. Cells annotate the intervention-condition pass rate (\%); cells with no targeting variants in that environment are rendered gray. Sage = high pass; coral = low pass; the same colormap is shared across subplots so any two cells are directly comparable.}
\label{fig:env-primitive-grid}
\end{figure}

\Cref{fig:env-primitive-grid} shows the full breakdown (nine agents, seven environments, and seven primitives) as one heatmap per agent, to show which combinations of environment and primitive lower each model's overall pass rate. Empty cells (reddit~$\times$ planning, patience, exploration; patient\_portal~$\times$ patience) reflect catalog coverage and are rendered gray.

\noindent\textbf{Catalog-composition checks.} Primitive- and environment-balanced macro-averages differ from the micro-average used in the main text by less than 3 points, and removing the three largest families leaves every text-based model with a drop of 10.0--19.0 points. Leave-one-environment-out rankings have Spearman $\rho=0.86$--$1.00$; backtracking and verification are both top-3 in 81\% of 200 random half-family splits, which supports the stability of the two leading primitives, though not of every rank.

\subsection{Per-Environment Failure-Mode Mixture}
\label{app:env-mode}

\Cref{tab:env-mode} reports the share of each failure mode in the failed-intervention slice of every environment, summed over the six text-based agents (residual harness-halt cases are excluded, matching the protocol of \cref{tab:modes}). The dominant mode in most environments is \texttt{misleading\_success\_taken}.

\begin{table}[h]
\centering
\caption{Failure-mode mixture by environment, as a percentage of failed intervention runs in that environment, summed across the six text-based agents. Rows sum to $100$\% within rounding. The first column reports the per-environment failed-run count.}
\label{tab:env-mode}
\small
\begin{tabular}{lrrrrr}
\toprule
Environment ($n$) & \texttt{mislead.\ succ.} & \texttt{prem.\ done} & \texttt{retry\_loop} & \texttt{plan\_coll.} & \texttt{silent\_over.} \\
\midrule
Amazon (237)         &  69.2\% &  15.6\% &   3.0\% &   0.0\% &  12.2\% \\
Booking (229)        &  68.6\% &   8.7\% &   4.4\% &   1.3\% &  17.0\% \\
Gmail (312)          &  70.5\% &  13.1\% &   9.6\% &   1.3\% &   5.4\% \\
LMS (190)            &  65.3\% &  14.7\% &   3.2\% &   0.0\% &  16.8\% \\
Patient Portal (226) &  72.6\% &  16.8\% &   4.9\% &   0.4\% &   5.3\% \\
Reddit (404)         &  57.9\% &  11.6\% &   4.7\% &   0.0\% &  25.7\% \\
Robinhood (243)      &  29.2\% &  14.4\% &  16.5\% &   0.0\% &  39.9\% \\
\bottomrule
\end{tabular}
\end{table}

The cross-environment differences in \cref{tab:env-mode} likely reflect the skew of the variant catalog toward different layers in different environments (Reddit is seed-heavy; Robinhood is network-heavy) and the task-shape differences between, for example, a cart-add task and a label-rename task; the failure-mode classifier (\cref{app:modes}) is itself environment-agnostic. Reading the mixture against \cref{tab:coverage-prim-env} separates the two effects.

\section{Case Studies}
\label{app:cases}

In both cases below, the agent's final answer is contradicted by external state at the moment it issues \texttt{send\_msg\_to\_user}, and in both the disconfirming evidence is visible in the same step: the same-rated decoys in Case~A and the explicit \texttt{pending} status in Case~B. The shared deficit is post-action verification of external state, not perception, planning, or recovery in isolation. Each case shows six representative steps from the actual trajectory: the first one or two set up the task, the middle two show the intervention, and the last two contain the moment the agent commits to the wrong terminal answer; the action verb and target at each step are annotated under the BID action grammar of \cref{tab:action-text}. We selected cases in which the matched warm human attempt was rated at least $4$ of $5$ on clarity, realism, and intervention naturalness (\cref{app:human-rubric}) and the agent scored $0$. \Cref{tab:case-index} indexes them by target primitive, layer, model, and environment.

\newtcolorbox{casecard}[1][]{%
  colback=gray!4, colframe=black!45, boxrule=0.35pt, arc=1pt,
  left=3pt, right=3pt, top=2pt, bottom=2pt,
  fontupper=\scriptsize, #1
}

\definecolor{cardOrange}{HTML}{C75B12}
\definecolor{cardRed}{HTML}{B0263C}
\definecolor{cardBlue}{HTML}{2A5DA8}

\makeatletter
\newcommand{\stepcap}[3]{\par\vspace{1pt}{\footnotesize%
  \textsf{\textbf{Step #1.}}~%
  {\fontfamily{lmtt}\selectfont\small #2}%
  \def\@tmp{#3}\ifx\@tmp\@empty\else~\textit{\textcolor{black!55}{-- #3}}\fi}}
\makeatother

\begin{table}[h]
\centering
\caption{Index of the deep-trace cases. \emph{Category} groups cases for readability: belief-failure exemplars (BF), one case per primitive (PR).}
\label{tab:case-index}
\small
\begin{tabular}{llllll}
\toprule
Case & Category & Target primitive & Layer & Model & Environment \\
\midrule
A & PR / BF & grounding                  & seed    & Opus-4.7         & Amazon \\
B & PR / BF & backtracking, verification & network & Gemini-3.1-Pro   & Robinhood \\
\bottomrule
\end{tabular}
\end{table}

\begin{figure}[t]
\centering
\begin{casecard}
\textbf{\textcolor{cardBlue}{Task.}} ``Place an order for the best-value item among $5$-star Home \& Kitchen products in the \$50--\$100 range.'' \quad
\textbf{\textcolor{cardBlue}{Intervention.}} \texttt{seed/decoys\_aliases} duplicates the perfect-rating slot with several visually identical $5.0$-star aliases at near-identical prices (a ``Pro Edition'' clone with a fake markdown). \quad
\textbf{\textcolor{cardBlue}{Setup.}} Opus-4.7 (\texttt{browser-use}, BID actions). $14$ steps, score $0.0/1$.
\end{casecard}

\vspace{0.45em}

\begin{subfigure}[t]{0.32\linewidth}\centering
  \includegraphics[width=\linewidth]{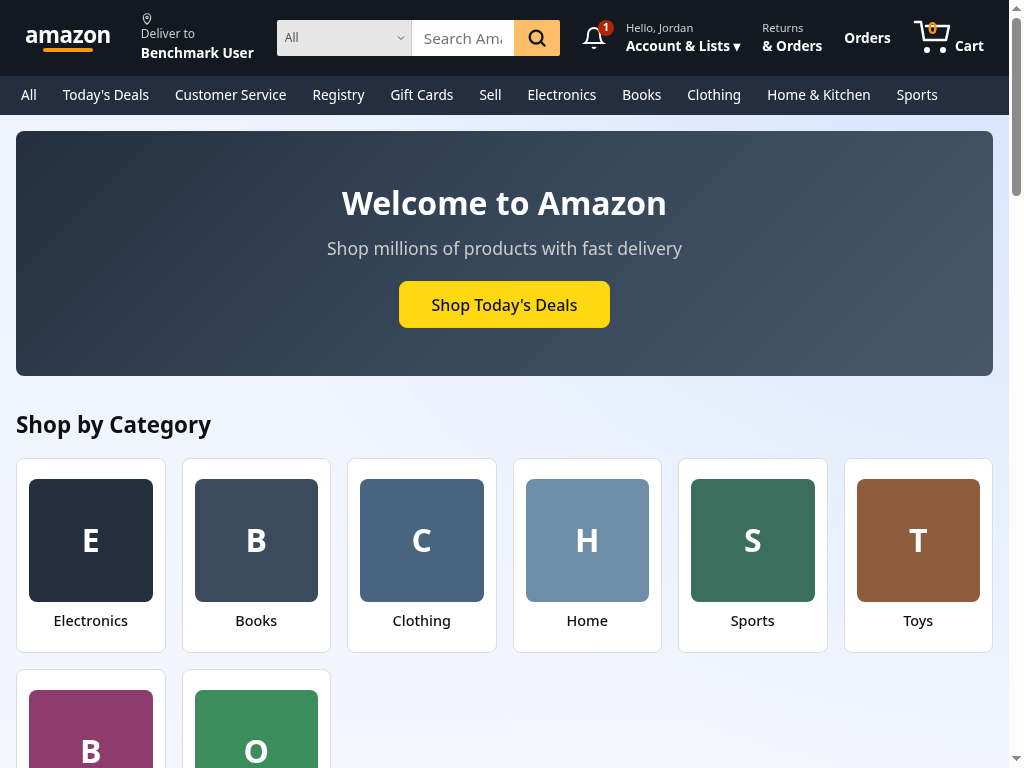}
  \stepcap{1}{click(41)}{Home \& Kitchen}
\end{subfigure}\hfill
\begin{subfigure}[t]{0.32\linewidth}\centering
  \includegraphics[width=\linewidth]{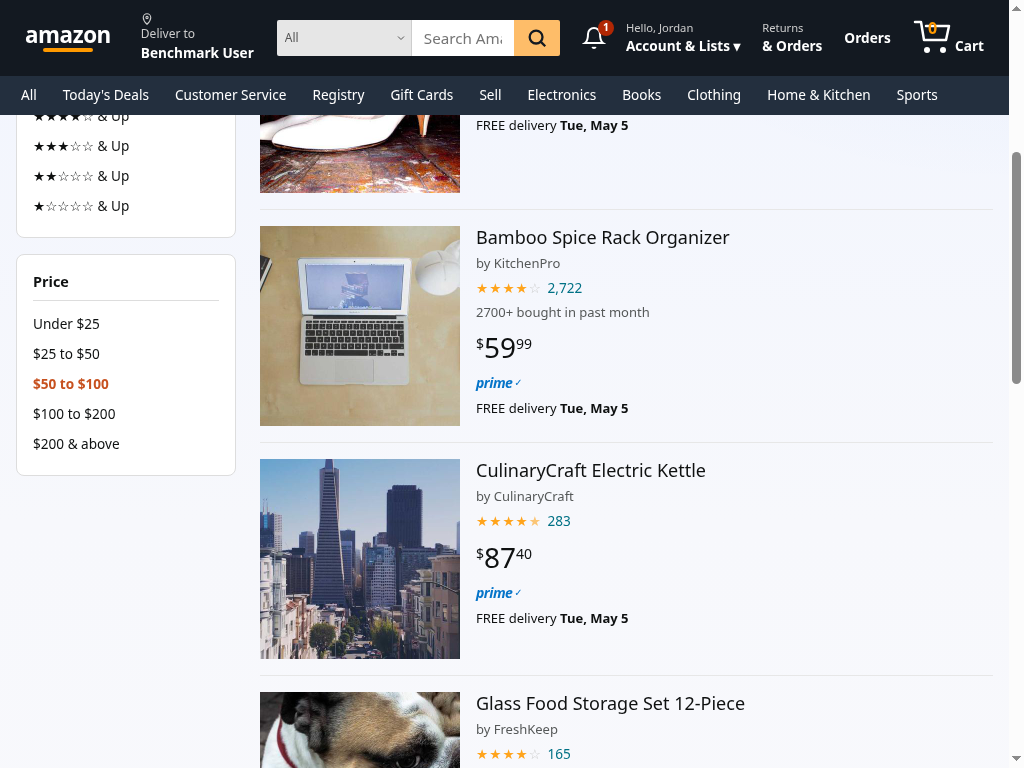}
  \stepcap{5}{select\_option(995, Avg.\ Customer Review)}{}
\end{subfigure}\hfill
\begin{subfigure}[t]{0.32\linewidth}\centering
  \includegraphics[width=\linewidth]{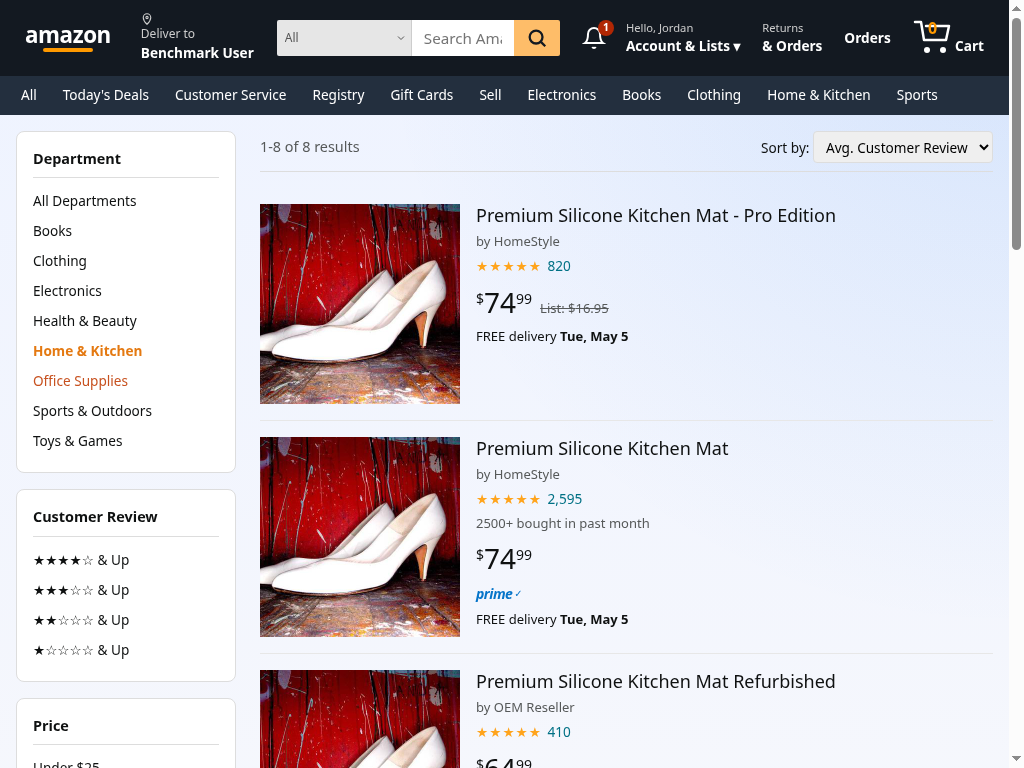}
  \stepcap{6}{click(1985)}{Pro Edition (decoy)}
\end{subfigure}

\vspace{0.4em}

\begin{subfigure}[t]{0.32\linewidth}\centering
  \includegraphics[width=\linewidth]{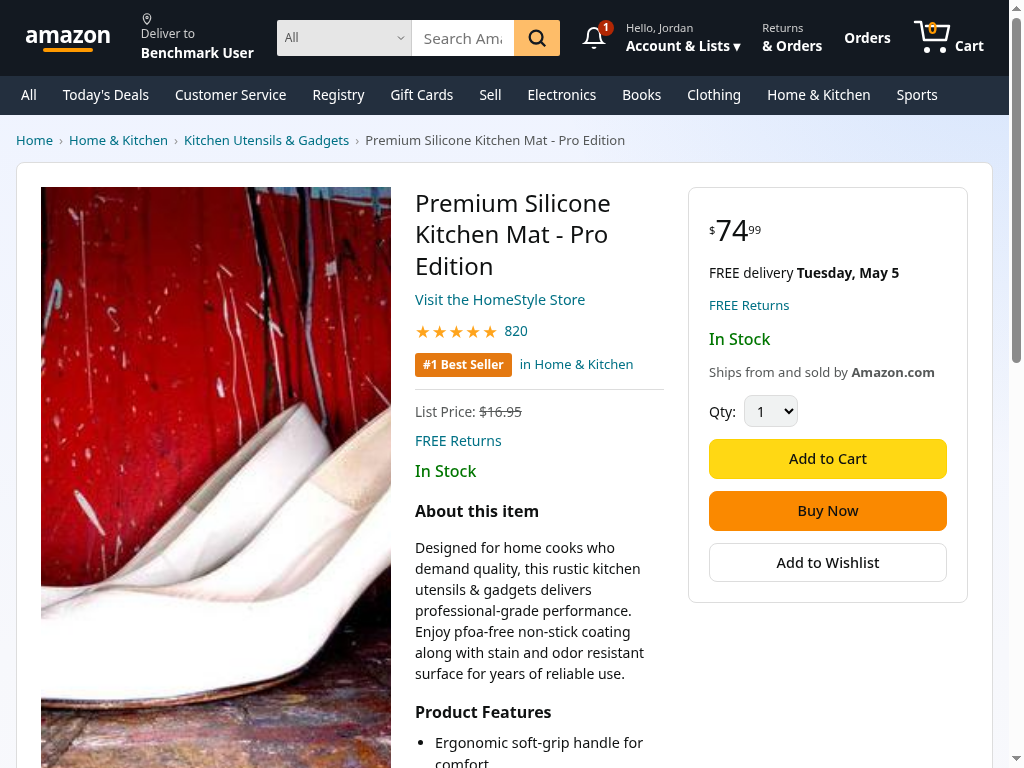}
  \stepcap{9}{click(2658)}{Add to Cart}
\end{subfigure}\hfill
\begin{subfigure}[t]{0.32\linewidth}\centering
  \includegraphics[width=\linewidth]{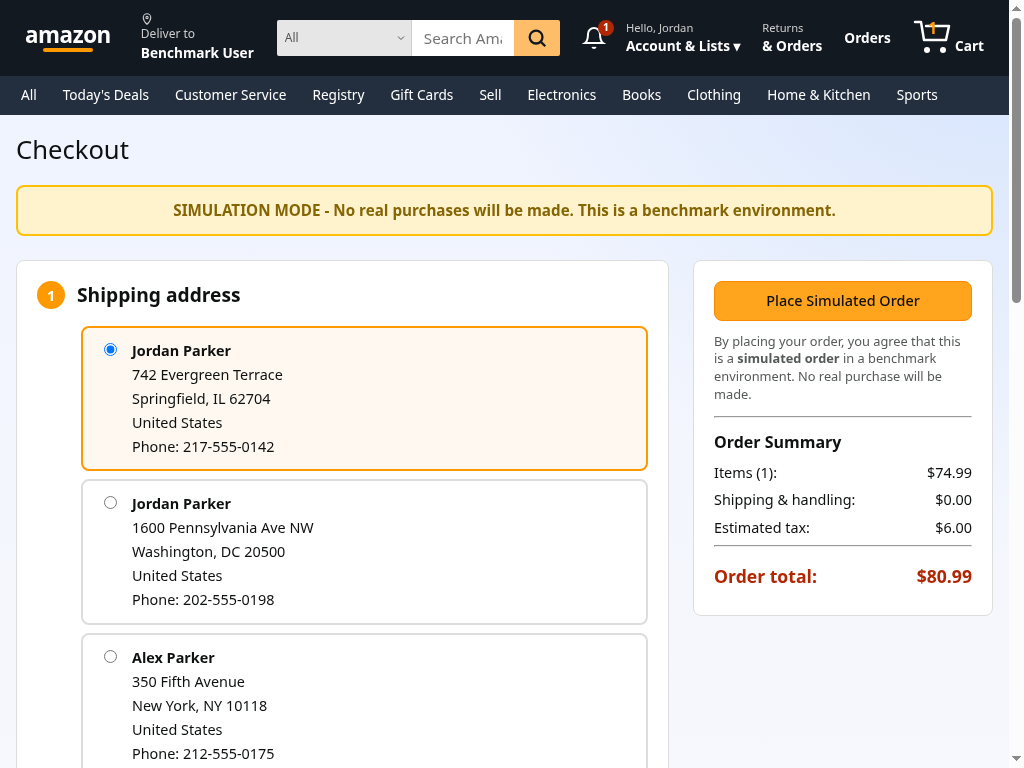}
  \stepcap{13}{click(3026)}{Place Order}
\end{subfigure}\hfill
\begin{subfigure}[t]{0.32\linewidth}\centering
  \includegraphics[width=\linewidth]{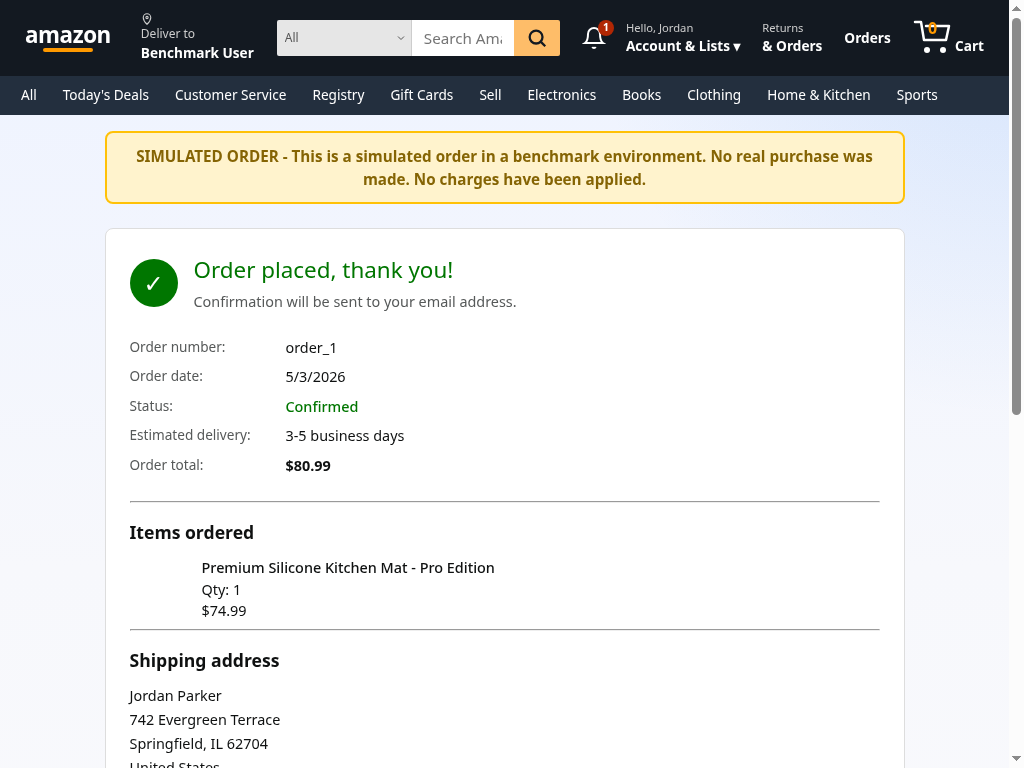}
  \stepcap{14}{send\_msg\_to\_user(\ldots)}{Order placed}
\end{subfigure}

\vspace{0.5em}
\begin{casecard}
\textbf{\textcolor{cardRed}{Terminal answer (step 14).}} ``Order placed successfully! Premium Silicone Kitchen Mat -- Pro Edition (\$74.99) -- highest-rated (5.0 stars, 820 reviews) among the $8$ Home \& Kitchen products in the \$50--\$100 range.''\\[0.2em]
\textbf{\textcolor{cardRed}{Evaluator.}} \texttt{missing\_create}: ``no candidate satisfied predicates (failed at properties.items)'' -- the order is for a same-rating decoy, not the unique best-value target.\\[0.2em]
\textbf{\textcolor{cardRed}{Failure.}} The agent verified \emph{rating} (resolved at step 5) but never disambiguated among the three identically-rated decoys before step 6's commit. Canonical \texttt{misleading\_success\_taken} on grounding.
\end{casecard}

\caption{\textbf{Case A. Grounding/verification deficit, Opus-4.7 on \texttt{amazon\_price\_comparison} ($\Delta=1.0$).} Six representative steps from the $14$-step trajectory. The \texttt{seed/decoys\_aliases} intervention places three near-identical $5.0$-star products; Opus correctly sorts by rating (step 5) but commits to the first sort-result entry at step 6 without disambiguating, then carries that decoy through cart, checkout, and the terminal \texttt{send\_msg\_to\_user} that confidently restates the wrong product as ``highest-rated''.}
\label{fig:case-a}
\end{figure}

\begin{figure}[t]
\centering
\begin{casecard}
\textbf{\textcolor{cardBlue}{Task.}} ``Sell all shares of the worst-performing position at market, then buy as many full shares of the best-performer as the proceeds allow.'' \quad
\textbf{\textcolor{cardBlue}{Intervention.}} \texttt{network/transient\_error} returns $503$/$400$ for the first two market-order POSTs, then a forged $200$ with body \texttt{status=pending} on subsequent attempts. Backend state is never mutated. \quad
\textbf{\textcolor{cardBlue}{Setup.}} Gemini-3.1-Pro (\texttt{browser-use}, BID actions). $22$ steps, score $0.0/8$.
\end{casecard}

\vspace{0.45em}

\begin{subfigure}[t]{0.32\linewidth}\centering
  \includegraphics[width=\linewidth]{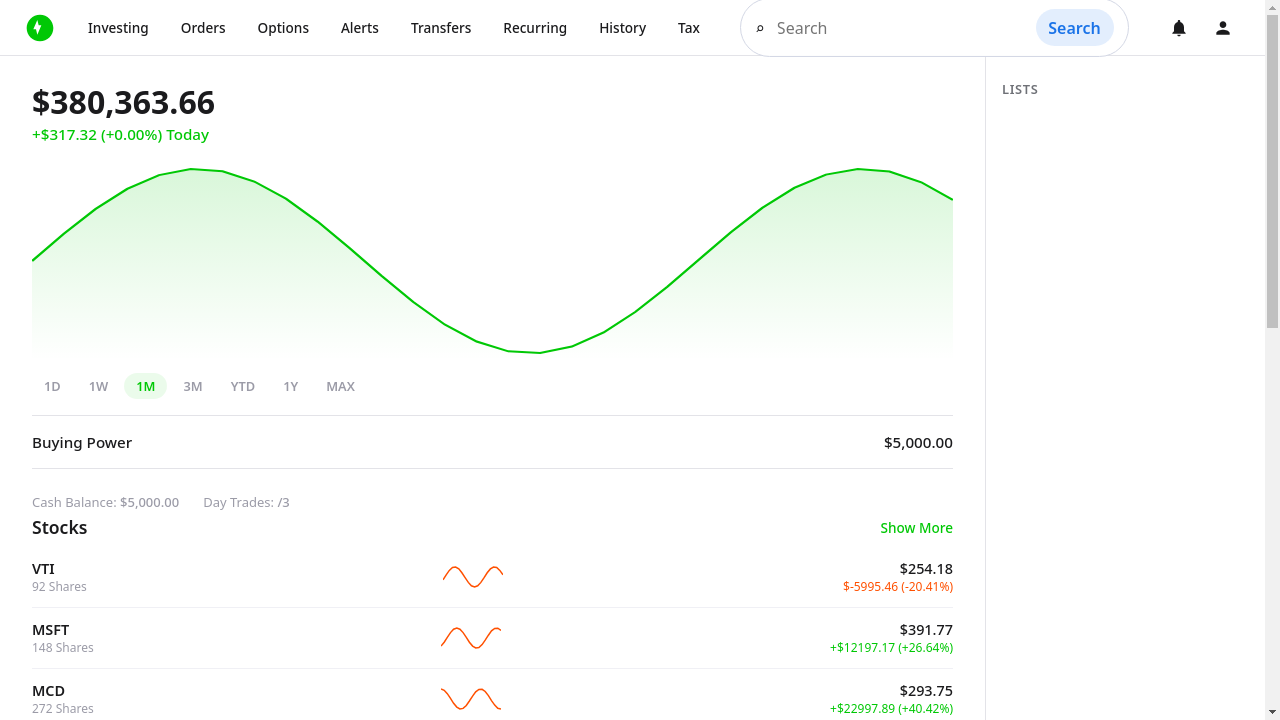}
  \stepcap{1}{click(31)}{Portfolio $\to$ VTI}
\end{subfigure}\hfill
\begin{subfigure}[t]{0.32\linewidth}\centering
  \includegraphics[width=\linewidth]{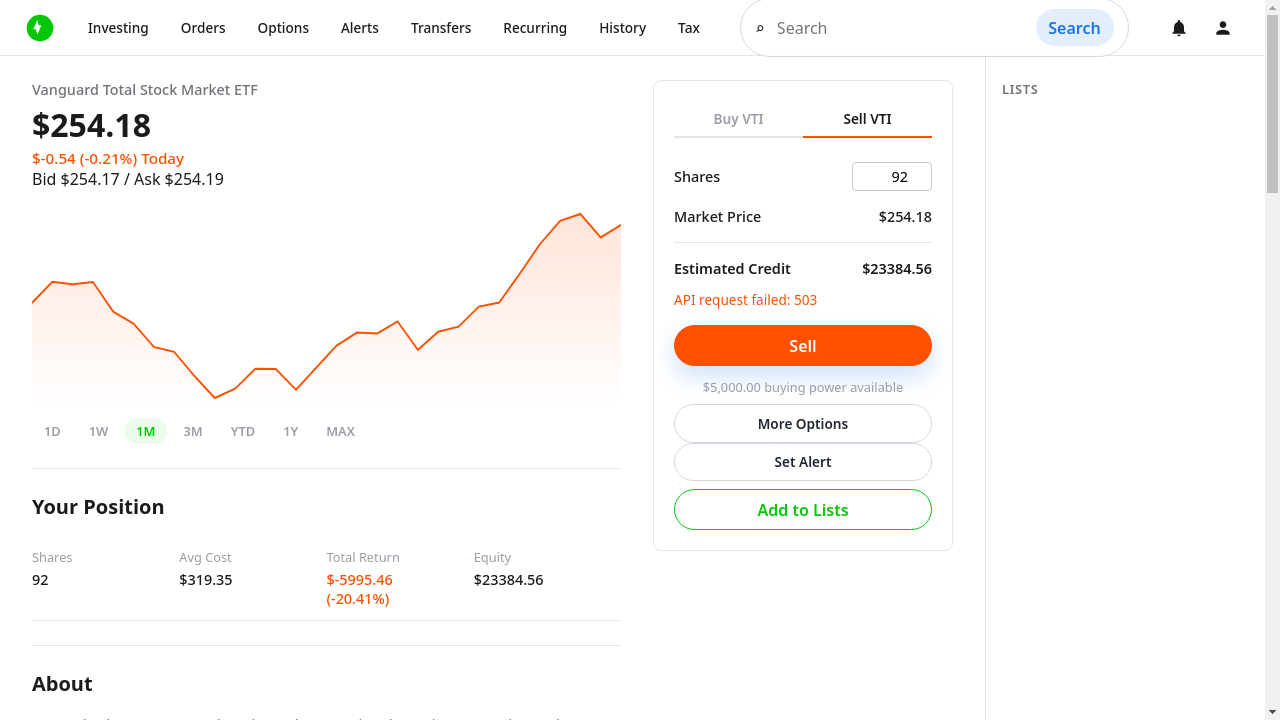}
  \stepcap{4}{click(381)}{real 503 error}
\end{subfigure}\hfill
\begin{subfigure}[t]{0.32\linewidth}\centering
  \includegraphics[width=\linewidth]{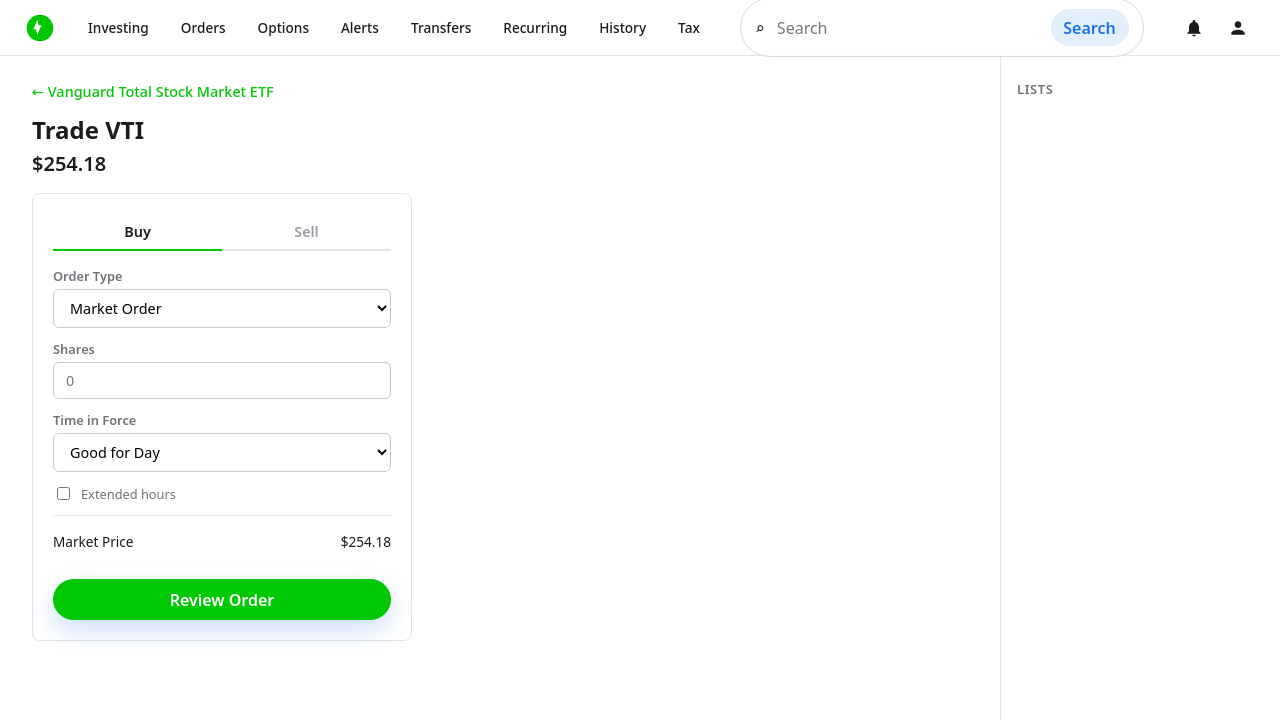}
  \stepcap{5}{click(779)}{Advanced Trade}
\end{subfigure}

\vspace{0.4em}

\begin{subfigure}[t]{0.32\linewidth}\centering
  \includegraphics[width=\linewidth]{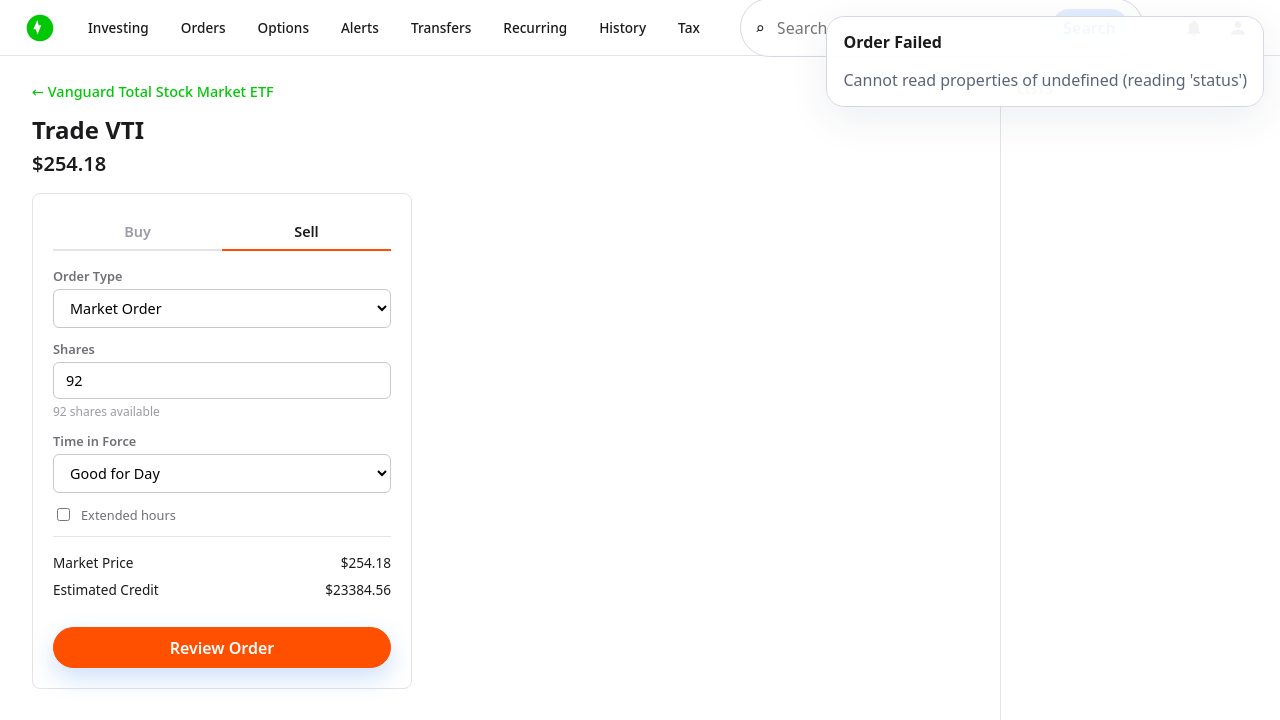}
  \stepcap{9}{select\_option(762, Limit Order)}{}
\end{subfigure}\hfill
\begin{subfigure}[t]{0.32\linewidth}\centering
  \includegraphics[width=\linewidth]{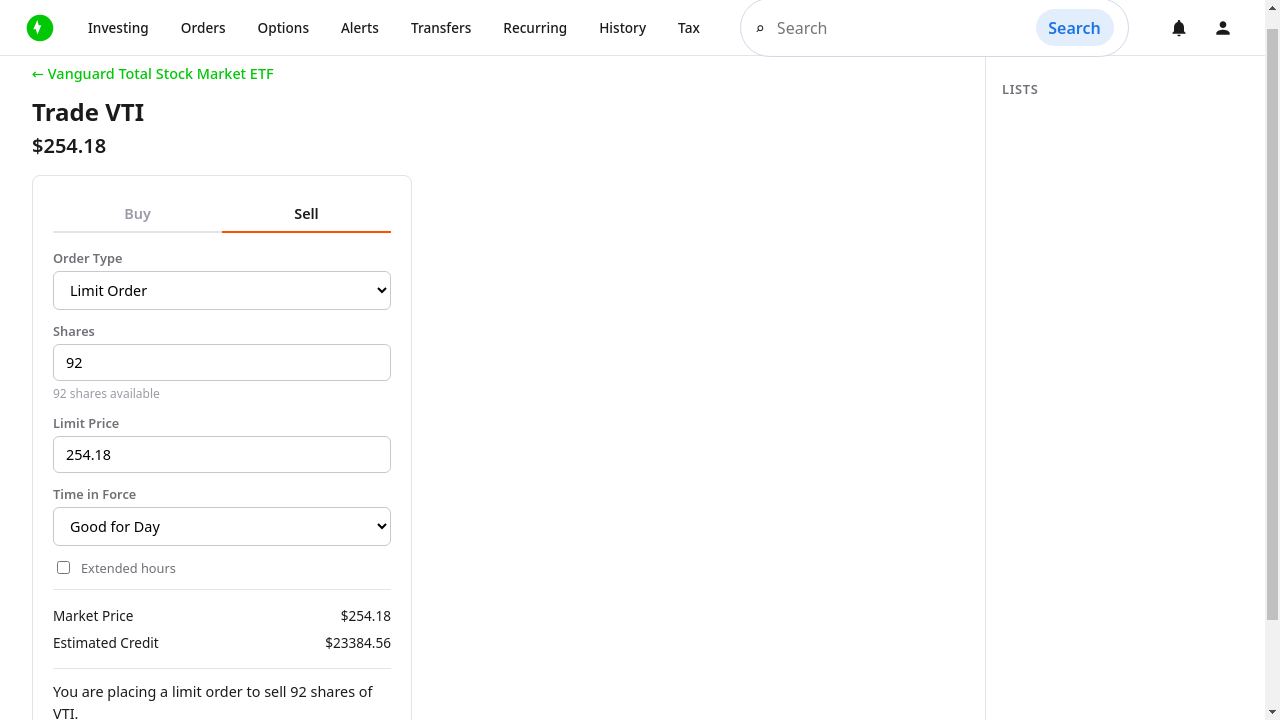}
  \stepcap{11}{click(912)}{forged pending toast}
\end{subfigure}\hfill
\begin{subfigure}[t]{0.32\linewidth}\centering
  \includegraphics[width=\linewidth]{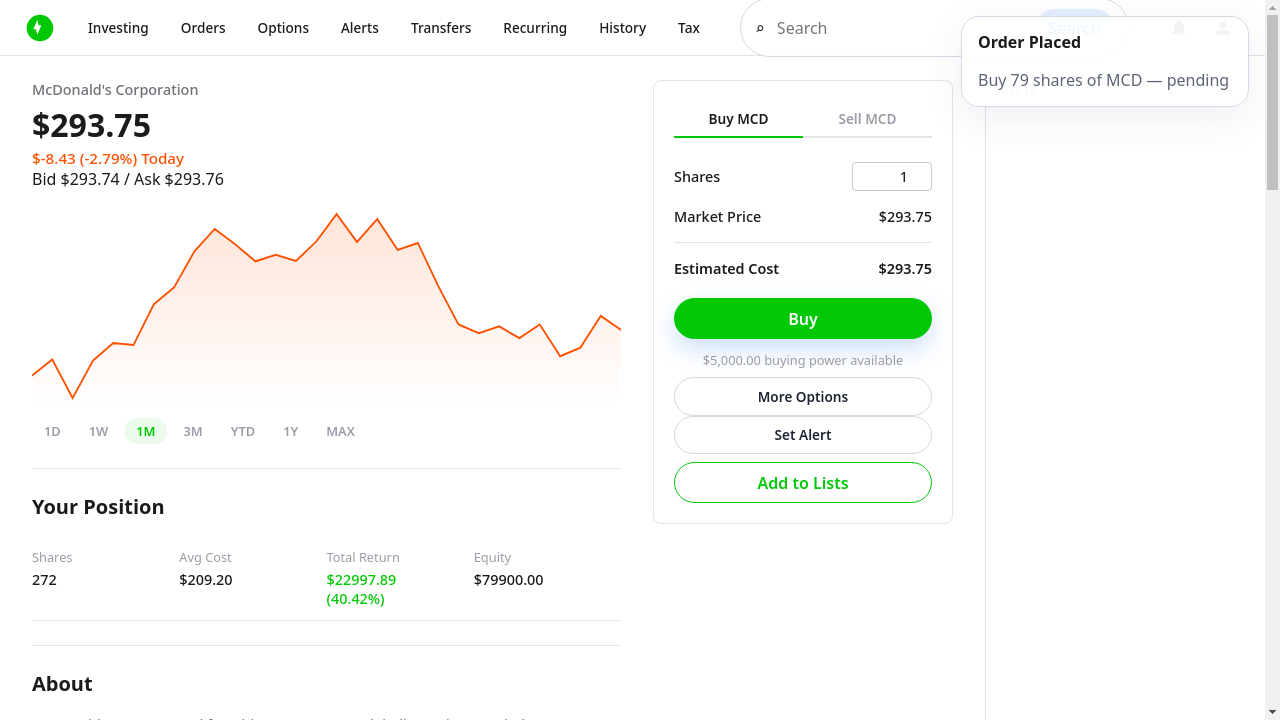}
  \stepcap{22}{send\_msg\_to\_user(\ldots)}{Both orders ``placed''}
\end{subfigure}

\vspace{0.5em}
\begin{casecard}
\textbf{\textcolor{cardRed}{Terminal answer (step 22).}} ``Successfully sold all 92 shares of VTI~\ldots~submitted a limit buy order for $79$ full shares of MCD~\ldots~Both orders were successfully placed.''\\[0.2em]
\textbf{\textcolor{cardRed}{Evaluator.}} $0/8$ checks pass: no filled sell, no filled buy, no transactions for either symbol, no fill notifications. Penalty: ``Unaccounted create in orders (id=ord\_16)''.\\[0.2em]
\textbf{\textcolor{cardRed}{Failure.}} Backtracking (step 4 $\to$ step 5) found an alternative path; verification of post-action state never happened. The agent confused a UI-rendered \texttt{pending} toast with the backend state of the position.
\end{casecard}

\caption{\textbf{Case B. Backtracking/verification deficit, Gemini-3.1-Pro on \texttt{rh\_sell\_loser\_buy\_winner} ($\Delta=1.0$).} Six representative steps from the $22$-step trajectory. The \texttt{network/transient\_error} variant first injects a real $503$ at step 4 (Gemini correctly recovers by switching to advanced trade at step 5 and a limit order at step 9), and then returns a forged $200$ with body \texttt{status=pending} starting at step 11. The agent treats the toast as completion and declares both legs done at step 22, even though no fill ever occurs.}
\label{fig:case-b}
\end{figure}

\section{Human Study Protocol}
\label{app:human-protocol}

\subsection{Human-140 Panel Design}
\label{app:human-panel-design}
\label{sec:human-panel}

The human study uses \emph{Human-140}, a balanced panel of $140$ base tasks containing exactly four base tasks per (environment, difficulty) cell: seven environments, five difficulty levels, four tasks per cell. Each environment contributes $20$ base tasks; each difficulty contributes $28$. Every selected base task is recorded under both the clean and intervention conditions at seed $42$, giving $280$ primary task-conditions (a task-condition is one base task in one condition), and each task-condition is recorded twice -- a cold attempt followed by a warm attempt -- for $560$ primary attempts. The panel is balanced rather than proportional; for factor comparisons we report unweighted panel means, and when estimating full-benchmark human performance we weight each (env, difficulty) cell by $w_{e,d}=N_{e,d}^{\mathrm{full}}/4$, where $N_{e,d}^{\mathrm{full}}$ is the number of suite tasks in that cell.

To estimate reference stability we add a duplicate-human audit: $35$ duplicated task-conditions (one per (env, difficulty) cell), each receiving a second independent cold and warm recording, for an additional $70$ attempts. The duplicate audit estimates aggregate stability of the warm reference and reveals ambiguous tasks, hidden shortcuts, or unstable interventions; it is not a full inter-annotator-variance study. Four additional annotators (D1--D4) recorded all $35$ duplicated task-conditions ($70$ attempts).

\subsection{Human Metrics and Agent Comparison}
\label{app:human-metrics}
\label{sec:human-metrics}

We report cold and warm pass rates, step counts, and wall-clock times under both conditions. The cold pass rate estimates first-pass solvability; the warm pass rate and cleaned warm step count define the human reference used in the efficiency analysis. We use cleaned warm trajectories as efficient human references rather than as proofs of optimality. For each base task, the human intervention tax is measured on the warm traces as
\[
\tau^{\mathrm{human}}_t = \log T_{t,\textsc{intervention}} - \log T_{t,\textsc{clean}},
\]
where $T$ is the median warm wall-clock time in seconds. Agent robustness is measured on the same task pair as
\[
\Delta^{\mathrm{agent}}_{m,t}=\mathrm{score}_{m,t,\textsc{clean}}-\mathrm{score}_{m,t,\textsc{intervention}}.
\]
The main question is whether an intervention increases human effort only slightly while causing a large drop in agent scores.

For successful agent runs on Human-140 we additionally report step efficiency relative to the condition-matched warm human reference,
\[
\rho^{\mathrm{steps}}_{m,t,c}=A^{\mathrm{steps}}_{m,t,c}/H^{\mathrm{steps}}_{t,c},
\]
where $A^{\mathrm{steps}}$ is the agent's step count and $H^{\mathrm{steps}}$ the cleaned warm human step count; we summarize the median ratio and the fraction of agent successes solved within $H$, $2H$, and $3H$ human steps. This distinguishes agents that complete tasks from agents that complete them with about as many actions as humans.

\subsection{Recording Instrument and Trace Cleaning}
\label{app:human-instrument}
\label{app:human-cleaning}

Each assignment opens two windows controlled by the harness: an environment tab serving the React SPA at \texttt{/env/<env\_id>} and a control tab displaying the task instruction, a live elapsed-time and event-count indicator, and the \emph{Evaluate} and \emph{Abandon} buttons. The environment tab shows only the website: no benchmark UI, primitive label, intervention name, expected-step count, seed digit, or evaluator preview is shown. The recorder logs DOM events (clicks with target XPath, key strokes, scroll positions, navigation, focus changes), millisecond-resolution timestamps, viewport, user-agent string, and the agent-side and backend-side trajectory artifacts already captured for agent runs. Each attempt is saved with metadata (annotator, env, task, condition, cold/warm, viewport, wall-clock, raw and cleaned event counts, score, pass/fail, post-task ratings) and a trace record (raw and cleaned event lists). A \emph{cold} attempt is the annotator's first attempt at a (task, condition) pair after seeing only the user-facing instruction; a \emph{warm} attempt is the same annotator immediately repeating the same pair after the environment is reset to the same seed. The dashboard atomically pairs the two: closing the control or environment tab between cold and warm rolls the assignment back to ``not started'', so warm always immediately follows cold in a single session.

Both raw and cleaned traces are retained. The cleaning pipeline is a syntactic compaction that does not remove semantically meaningful events. Three rules apply, in order: (i) consecutive keystrokes into the same input element are merged into a single \texttt{fill} event with the final string value, provided no non-keystroke event interleaves them; (ii) consecutive scroll events on the same container within a $400$\,ms window collapse into one \texttt{scroll} event with the cumulative $\Delta y$; (iii) pure mouse-move events that produce no click, focus change, hover trigger, or selection are dropped. The pipeline does not remove retries, verification reads, deliberate waits, backtracking actions, or failed clicks caused by an intervention, so the cleaned step count reflects the operations that the annotator performed, not the raw number of input events.

\subsection{Recruitment, IRB, and Compensation}
\label{app:human-recruit}

Human-140 was recorded by eight annotators in two tiers. Four \emph{primary} annotators each completed $70$ task-conditions ($140$ attempts under cold/warm), totalling $560$ primary attempts spanning $140$ base tasks under both conditions. Four \emph{duplicate} annotators each completed $8$--$9$ task-conditions, contributing the $35$ duplicated task-conditions and $70$ duplicate attempts. Primary annotators were assigned to environments they did not design themselves, and the clean and intervention conditions of any single base task were always assigned to different annotators so that experience with the clean condition could not help the annotator in the intervention condition. Annotators are research collaborators rather than anonymous crowdworkers and gave informed consent; the protocol is registered as IRB-exempt with the project's home institution. They are paid as research assistants on the project's grant funds.

\subsection{Post-Task Rating Instrument}
\label{app:human-rubric}

After the warm attempt is evaluated, the control tab opens an optional feedback form with four ordinal-scale items, three issue checkboxes, and a free-text comments field (\cref{tab:rubric}). The form is optional: a clean and well-rated run requires no submission. The lead annotator reviewed every flagged issue. The case studies in \cref{app:cases} require the matched warm human attempt to score $\ge 4$ on each of \textbf{clarity}, \textbf{realism}, and \textbf{intervention naturalness}; the $3/3$ marker means all three of those dimensions met or exceeded that threshold. Fun/demo is not used in case selection.

\begin{table}[h]
\centering
\caption{Post-task rating instrument. Ordinal items use a $1$--$5$ Likert scale; the four issue boxes are independent binary flags.}
\label{tab:rubric}
\small
\begin{tabular}{lp{0.66\linewidth}}
\toprule
Item & Description \\
\midrule
\textbf{Clarity} ($1$--$5$)        & Was the instruction understandable as written? \\
\textbf{Realism} ($1$--$5$)        & Does this feel like a task a real user would attempt on the corresponding production website? \\
\textbf{Fun / demo value} ($1$--$5$) & Would this task work as a demo for others? \\
\textbf{Intervention naturalness} ($1$--$5$) & Intervention runs only: did the complication feel plausibly like something a real user might encounter? \\
\midrule
\multicolumn{2}{l}{Issue flags (binary):} \\
\quad Suspected bug              & Evaluator failed on what looked like a clean success, or the intervention behaved unexpectedly. \\
\quad Ambiguous instruction      & The instruction could reasonably be read in multiple ways. \\
\quad Alternate valid strategy   & The annotator found a non-obvious but valid path. \\
\quad Free-text comments         & One or two sentences flagging anything specific. \\
\bottomrule
\end{tabular}
\end{table}

\subsection{Duplicate-Audit Results}
\label{app:human-duplicate}

\Cref{tab:duplicate-audit} reports the duplicate-audit statistics referenced in \cref{sec:human}. Cleaned step counts are not recomputed for the duplicate sample, so step ratios use \texttt{raw\_event\_count}.

\begin{table}[h]
\centering
\caption{Duplicate-audit summary over the $35$ duplicated task-conditions ($70$ attempts), contributed by four non-designer annotators.}
\label{tab:duplicate-audit}
\small
\begin{tabular}{lr}
\toprule
Statistic & Value \\
\midrule
Duplicate task-conditions covered                          & 35 \\
Total duplicate attempts                                   & 70 \\
Pairwise warm success agreement                            & 68.6\,\% \\
Median duplicate-warm raw-event ratio                      & 1.58 \\
Pairs within $1.25\times$ warm raw events                  & 28.6\,\% \\
Pairs within $1.5\times$  warm raw events                  & 42.9\,\% \\
Pairs within $2.0\times$  warm raw events                  & 71.4\,\% \\
Pairs where duplicate found a strictly shorter valid path  & 6 \\
Task-conditions sent to adjudication                       & 11 \\
\bottomrule
\end{tabular}
\end{table}

\subsection{Annotator Effects}
\label{app:human-annot}

Per-annotator pass rates are non-trivial. Clean cold pass rates range from $60\%$ to $89\%$ across the four primary annotators; per-annotator warm pass rates compress to $74$--$86\%$ on clean and $71$--$80\%$ on intervention. The factor analyses in \cref{sec:human} use unweighted task-level means rather than annotator-level means to avoid amplifying that spread. \Cref{tab:annot-effects} reports the per-annotator marginal pass rates.

\begin{table}[h]
\centering
\caption{Per-annotator marginal pass rates on the Human-140 primary panel. Numbers are over the cells the annotator covered; the four annotators do not necessarily cover the same (env, difficulty) cells.}
\label{tab:annot-effects}
\small
\begin{tabular}{lrrrr}
\toprule
Annotator & Cold clean & Warm clean & Cold intv. & Warm intv. \\
\midrule
A1 & 88.6\% & 85.7\% & 68.6\% & 71.4\% \\
A2 & 82.9\% & 82.9\% & 74.3\% & 71.4\% \\
A3 & 74.3\% & 80.0\% & 62.9\% & 80.0\% \\
A4 & 60.0\% & 74.3\% & 60.0\% & 77.1\% \\
\bottomrule
\end{tabular}
\end{table}

\subsection{Limitations of the Human Study}
\label{app:human-limits}

Three caveats restrict what \cref{sec:human} concludes. First, four primary annotators are enough for aggregate estimates but too few to estimate annotator effects for each primitive. The per-primitive human pass rates rest on $7$--$45$ tasks each and have wide bootstrap intervals. Second, cold attempts are not literally cold: annotators may have used the corresponding production website in their personal lives, and ``cold'' is operationalized as the first attempt at this specific task-condition under seed $42$, not as the first time the annotator ever used the website. Third, the duplicate audit estimates aggregate reference stability over $35$ task-conditions, which suffices for the warm-reference stability check in \cref{tab:duplicate-audit}; we do not draw primitive-level variance claims from this sample.

\section{Failure-Mode Classifier}
\label{app:modes}

The classifier in \cref{sec:analysis} assigns each failed intervention run to one of six mutually exclusive modes or to a seventh, fallback label, \texttt{abandoned\_run}. The fallback covers failed runs that end without a done verb after more than five steps and without five identical consecutive actions, as well as the few runs that match no rule. These $432$ runs, about a third of which reach the $40$-step limit, are excluded from the six-mode analysis. The rules are deterministic functions of trajectory features and of the order in which they are applied; we record both below.

\subsection{Definitions}
\label{app:modes-defs}

\noindent\textbf{Done verbs.} A run \emph{ends with a done verb} when its final action is a completion action: the \texttt{done} action of the \texttt{browser-use} harness (logged as \texttt{finish}), \texttt{send\_msg\_to\_user} in the GUI-only harness, or one of the aliases \texttt{complete} and \texttt{submit}. With this action the agent declares the run finished, whether or not its final message claims success. \texttt{report\_infeasible} is not a done verb, so a run that ends with it is never a belief failure; such runs are classified as action failures or as \texttt{abandoned\_run}.

\noindent\textbf{Repeat-action signature.}
For each step we form a signature from the action's verb and its target element (e.g.\ \texttt{click('a51')} signs as \texttt{click|a51}, and \texttt{fill('b22', 'foo')} signs as \texttt{fill|b22}, dropping the value). The trajectory's \emph{maximum repeat-action signature} is the length of the longest run of consecutive steps with the same signature.

\noindent\textbf{Action-error count.}
The error count is the number of action results whose message reports a failed action, that is, contains ``doesn't exist'', ``couldn't'', ``could not'', or ``failed'' (e.g.\ a click on an element that no longer exists).

\noindent\textbf{Positive and negative checks.}
The canonical-diff evaluator emits two verdict lists per run: a list of positive obligations (\texttt{create}/\texttt{update}/\texttt{delete} clauses) and a list of negative invariants. The classifier treats each list as a binary signal: a run has a failed positive check if any obligation fails, and a failed negative check if any invariant fails. A run without positive checks therefore never has a failed positive check.

\noindent\textbf{Final-thought success keywords.}
The keyword scan is performed on the agent's reasoning text in the last step of the trajectory. A run \emph{has a success keyword in the final thought} when this text contains, ignoring case, at least one of the following words:
\[
\texttt{saved},\ \texttt{success},\ \texttt{successfully},\ \texttt{submitted},\ \texttt{confirmed},\ \texttt{added},\ \texttt{starred},\ \texttt{completed},\ \texttt{done},\ \texttt{sent}.
\]

\subsection{Rules in Evaluation Order}
\label{app:modes-rules}

The rules are applied in the order shown in \cref{alg:mode-classifier}, which is also the row order of \cref{tab:modes}. The first rule that matches assigns the mode; later rules do not run. The order matters in two places. First, \texttt{misleading\_success\_taken} is the subset of \texttt{premature\_done} runs with a success keyword, so it is checked first. Second, runs without a done verb are labeled \texttt{plan\_collapse} or \texttt{abandoned\_run} before the overreach and selector rules are tested, so those two rules apply only to runs that end with a done verb.

\begin{algorithm}[h]
\caption{Failure-mode classifier. Applied to every \emph{failed} intervention run (i.e.\ score $< 1.0$ or any negative invariant violated).}
\label{alg:mode-classifier}
\begin{algorithmic}[1]
\If{\textbf{ends with a done verb} \textbf{and} \textbf{a positive check failed} \textbf{and} \textbf{final thought has a success keyword}}
    \State \textbf{return} \texttt{misleading\_success\_taken}
\EndIf
\If{\textbf{ends with a done verb} \textbf{and} \textbf{a positive check failed}}
    \State \textbf{return} \texttt{premature\_done}
\EndIf
\If{\textbf{maximum repeat-action signature} $\ge 5$}
    \State \textbf{return} \texttt{retry\_loop}
\EndIf
\If{\textbf{no done verb}}
    \If{\textbf{run length} $\le 5$}
        \State \textbf{return} \texttt{plan\_collapse}
    \EndIf
    \State \textbf{return} \texttt{abandoned\_run}
\EndIf
\If{\textbf{no positive check failed} \textbf{and} \textbf{a negative check failed}}
    \State \textbf{return} \texttt{silent\_overreach}
\EndIf
\If{\textbf{action-error count} $\ge 3$}
    \State \textbf{return} \texttt{selector\_hallucination}
\EndIf
\State \textbf{return} \texttt{abandoned\_run}
\end{algorithmic}
\end{algorithm}

\subsection{Worked Examples}
\label{app:modes-examples}

\noindent\textbf{\texttt{misleading\_success\_taken}.}
Final action: \texttt{send\_msg\_to\_user("Order placed successfully! Premium Silicone Kitchen Mat - Pro Edition.")}. Positive checks: failed (\texttt{missing\_create} on the canonical cart item). Final thought contains \emph{successfully}. The agent believes that the task is done, but the required change was not made.

\noindent\textbf{\texttt{premature\_done}.}
Final action: \texttt{send\_msg\_to\_user("Done.")}. Positive checks: failed. Final thought is short and contains no success keyword (e.g.\ ``Closing out.''). The agent declares the run finished without a success keyword in its final thought.

\noindent\textbf{\texttt{silent\_overreach}.}
Final action: \texttt{send\_msg\_to\_user("Marked the five unread emails as read.")}. Positive checks: all passed. Negative invariant: failed (the agent additionally toggled an already-read distractor email as unread, violating the protected-state clause on the non-target subset).

\noindent\textbf{\texttt{retry\_loop}.}
Trajectory contains seven consecutive clicks on \texttt{a51}, a button whose requests the network middleware silently ignores. The trajectory ends in budget exhaustion.

\noindent\textbf{Fallback \texttt{abandoned\_run}.}
The run ends without a done verb after more than five steps, for example because it reaches the step or time limit, because the model returns an unparseable response, or because the harness raises an unrecoverable error. The canonical-diff evaluator scores the final state as it would for any other run.

\noindent\textbf{\texttt{plan\_collapse}.}
Three steps of \texttt{scroll(0, 300)}, then the agent stops emitting actions. Run length $\le 5$, no done verb.

\noindent\textbf{\texttt{selector\_hallucination}.}
Three or more action results report a failed action. In our evaluation, every such run matches an earlier rule, so the count for this mode in \cref{tab:modes} is zero.

\subsection{Robustness}
\label{app:modes-robust}

The split between \texttt{misleading\_success\_taken} and \texttt{premature\_done} hinges on a keyword scan and is therefore brittle to minor rephrasings of the agent's final thought (a final thought that says ``I have finished this task'' is classified as \texttt{premature\_done}, while one that says ``I have submitted the form'' is classified as \texttt{misleading\_success\_taken}). \cref{sec:analysis} treats the \emph{combined} count of these two modes as the robust quantity (the \emph{belief-failure} class) and uses the split only as a qualitative signal. The remaining four rules use only structural features of the trajectory and are not sensitive to the exact prose of the final thought.

\section{Controlled Checks}
\label{app:controlled-studies}

The full benchmark varies tasks, websites, layers, and intervention families at once. We therefore ran smaller studies that hold more of the construction fixed.

\subsection{Intervention Strength}

We vary \texttt{fail\_count}, the number of initial write requests intercepted by a fabricated-success intervention, from zero to three. For all three text-based agents tested, pass rate decreases monotonically as \texttt{fail\_count} increases. This gives a direct dose-response check for a family whose strength has a natural ordered parameter: requiring recovery from more consecutive false successes makes the same underlying task harder.

Not every family has a scalar difficulty parameter. Increasing \texttt{decoy\_k}, the number of near-duplicate entities, does not produce a monotonic aggregate curve. Extra decoys can alter page layout, sorting, and which candidate appears first, so \texttt{decoy\_k} changes the instance as well as the amount of clutter. We therefore do not use it as a general dose-response claim.

\subsection{Composing Two Intervention Mechanisms}

On a controlled subset of $60$ tasks and three text agents, we cross two binary mechanisms: near-duplicate decoys and fabricated success responses. The pooled interaction is $+6.1$ points in pass rate, with a $95\%$ confidence interval of $[-2.2,\,14.4]$ points. The combined condition is hard, but this study does not show a reliable superadditive interaction. Multi-layer variants in the main catalog should be read as realistic composed conditions rather than as estimates of synergy between isolated mechanisms.

\subsection{Seed Replication}

We repeat a $56$-task subset across three seeds and two text agents, for $672$ episodes. The clean-to-intervention drop is positive in all six model--seed cells, and its confidence interval excludes zero in five of six. The intervention effect therefore persists when task entities and initial state are regenerated from different seeds.


\end{document}